\pdfoutput=1
\RequirePackage{fix-cm} 
\documentclass{article}

\usepackage[preprint]{neurips_2024}

\usepackage{iftex}
\ifPDFTeX
  \usepackage[utf8]{inputenc}
\fi
\usepackage[T1]{fontenc}
\usepackage{microtype}
\usepackage{inconsolata}
\usepackage{graphicx}
\usepackage[font=small,labelfont=bf]{caption}
\usepackage[skins]{tcolorbox}
\usepackage{array}

\usepackage{amsmath}
\usepackage{amssymb}
\usepackage{booktabs}
\usepackage{enumitem}
\usepackage{float}
\usepackage{placeins} 
\usepackage{afterpage} 
\usepackage{needspace} 
\usepackage{multirow}
\usepackage{makecell}
\usepackage{xcolor}
\usepackage{colortbl}
\usepackage{tabularx}

\newcolumntype{Y}{>{\centering\arraybackslash}X}
\usepackage{tikz}
\usetikzlibrary{positioning,arrows.meta,calc,decorations.pathreplacing}
\usepackage{pgfplots}
\pgfplotsset{compat=1.17}
\usepgfplotslibrary{groupplots}
\definecolor{accOnce}{gray}{0.80}
\definecolor{accWin}{HTML}{B8D2E6}
\definecolor{accLoop}{HTML}{DEBE59}
\definecolor{accLoopDark}{HTML}{8B7429}
\definecolor{accHead}{HTML}{292F33}
\definecolor{accAccent}{HTML}{2B6A9B} 
\colorlet{accTitle}{accAccent}
\definecolor{accRunningTitle}{HTML}{667C8D} 
\definecolor{accTint}{HTML}{B8D2E6}
\definecolor{accCite}{HTML}{294F6D}
\colorlet{accRule}{accHead!20}
\colorlet{accAbsBack}{accTint!9}
\colorlet{accAbsFrame}{accTint!28}
\colorlet{accMeta}{accHead!75}

\tikzset{
  pics/accordion/.style args={#1/#2}{code={%
    \pgfmathsetmacro{\accw}{#1}%
    \pgfmathtruncatemacro{\accn}{#2}%
    \pgfmathsetmacro{\accoff}{-#1/2}%
    \begin{scope}[xshift=\accoff cm]
      \foreach \i in {1,...,\accn}{%
        \pgfmathsetmacro{\xa}{(\i-1)*\accw/\accn}%
        \pgfmathsetmacro{\xb}{\i*\accw/\accn}%
        \pgfmathsetmacro{\ya}{mod(\i-1,2)==0 ? 0.25 : 0.2}%
        \pgfmathsetmacro{\yb}{mod(\i,2)==0 ? 0.25 : 0.2}%
        \pgfmathtruncatemacro{\shd}{mod(\i,2)==0 ? 100 : 55}%
        \fill[accLoop!\shd] (\xa,\ya) -- (\xb,\yb) -- (\xb,-\yb) -- (\xa,-\ya) -- cycle;
      }
      \fill[gray!60, rounded corners=0.6pt] (-0.17,-0.31) rectangle (0,0.31);
      \fill[white] (-0.14,-0.26) rectangle (-0.06,0.26);
      \foreach \j in {-0.18,-0.09,0,0.09,0.18}{\draw[gray!70, line width=0.2pt] (-0.14,\j) -- (-0.06,\j);}
      \fill[gray!60, rounded corners=0.6pt] (\accw,-0.31) rectangle ({\accw+0.17},0.31);
      \foreach \j in {-0.18,-0.06,0.06,0.18}{\fill[white] ({\accw+0.085},\j) circle (0.025);}
    \end{scope}
  }},
}

\ifXeTeX
  \usepackage[no-math]{fontspec}
  \usepackage{xeCJK}
\else
  \usepackage{CJKutf8}
  
\fi

\newcommand{\method}{Accordion}
\newcommand{\mainmodel}{Accordion-6B-A0.5B}

\usepackage[colorlinks=true,linkcolor=accHead,citecolor=accCite,urlcolor=accHead]{hyperref}
\setcitestyle{round,aysep={,},yysep={,}}
\usepackage{fancyhdr}

\makeatletter
\let\accNeuripsNormalsize\normalsize
\let\accNeuripsSmall\small
\let\accNeuripsFootnotesize\footnotesize
\let\accNeuripsLarge\large
\let\accNeuripsCapitalLarge\Large
\let\accNeuripsSubsubsection\subsubsection
\let\accNeuripsParagraph\paragraph
\newlength{\accNeuripsParskip}
\newlength{\accNeuripsFloatSep}
\newcommand{\accTableFont}[2]{\fontsize{9}{10}\selectfont}
\newcommand{\accTableStretch}[1]{\renewcommand{\arraystretch}{1}}
\newcommand{\accLocalParskip}[1]{} 

\newcommand{\accUseCompactAppendix}{%
  \renewcommand{\accTableStretch}[1]{\renewcommand{\arraystretch}{1}}%
  \captionsetup[table]{skip=5pt}
  \setlength{\floatsep}{8pt plus 2pt minus 2pt}%
  \setlength{\textfloatsep}{10pt plus 2pt minus 2pt}%
  \setlength{\intextsep}{8pt plus 2pt minus 2pt}%
}

\newcommand{\accHeadingSpace}[1]{%
  \if@nobreak\else\Needspace*{#1\baselineskip}\fi
}

\newcommand{\accUseNeuripsBody}{%
  \renewcommand{\normalsize}{%
    \accNeuripsNormalsize
    \@setfontsize\normalsize{10.3}{13}%
  }%
  \let\small\accNeuripsSmall
  \let\footnotesize\accNeuripsFootnotesize
  \let\large\accNeuripsLarge
  \let\Large\accNeuripsCapitalLarge
  \renewcommand{\subsubsection}{\accHeadingSpace{3}\accNeuripsSubsubsection}%
  \renewcommand{\paragraph}{\accHeadingSpace{2}\accNeuripsParagraph}%
  \renewcommand{\section}{%
    \accHeadingSpace{6}%
    \@startsection{section}{1}{\z@}%
      {-2.0ex \@plus -0.5ex \@minus -0.2ex}%
      {1.5ex \@plus 0.3ex \@minus 0.2ex}%
      {\large\bfseries\raggedright\color{accAccent}}%
  }%
  \renewcommand{\subsection}{%
    \accHeadingSpace{4}%
    \@startsection{subsection}{2}{\z@}%
      {-1.8ex \@plus -0.5ex \@minus -0.2ex}%
      {0.8ex \@plus 0.2ex}%
      {\normalsize\bfseries\raggedright\color{accAccent}}%
  }%
  \normalsize
  \flushbottom 
  \setlength{\parskip}{4.5pt plus 1pt minus 0.5pt}%
  \setlength{\@fpsep}{\accNeuripsFloatSep}
  \captionsetup{font=normalsize,skip=7pt}
  \captionsetup[table]{position=top,skip=7pt}
}

\renewcommand{\@noticestring}{}
\renewcommand{\@notice}{}

\renewcommand{\normalsize}{%
  \accNeuripsNormalsize
  \@setfontsize\normalsize{10.8}{15}%
}
\renewcommand{\small}{\@setfontsize\small{9.5}{11.5}}
\renewcommand{\footnotesize}{\@setfontsize\footnotesize{9}{11}}
\let\large\accNeuripsLarge
\let\Large\accNeuripsCapitalLarge
\normalsize
\let\accOriginalFloatBarrier\FloatBarrier
\newcommand{\accEvidenceBarrier}{%
  \begingroup\let\FloatBarrier\accOriginalFloatBarrier\FloatBarrier\endgroup}
\renewcommand{\FloatBarrier}{\par}
\renewcommand{\@maketitle}{%
  \vbox{%
    \hsize\textwidth
    \linewidth\hsize
    \centering
    \setlength{\parskip}{0pt}
    {\@title\par}
    \vskip 10pt
    {\fontsize{10.7}{14}\selectfont\color{black}\@author\par}%
    \vskip 3pt
    {\normalsize\color{black}
      \textsuperscript{1}\,Tsinghua University\qquad
      \textsuperscript{2}\,Baidu Inc.\par}%
    \vskip 3pt
    {\fontsize{9.5}{12}\selectfont\color{black}%
      \href{mailto:btzhangyudong@gmail.com}{\texttt{btzhangyudong@gmail.com}}\quad
      \href{mailto:teresading999@gmail.com}{\texttt{teresading999@gmail.com}}\quad
      \href{mailto:zhengkl18@163.com}{\texttt{zhengkl18@163.com}}\par
      \href{mailto:dingyuchen@baidu.com}{\texttt{dingyuchen@baidu.com}}\quad
      \href{mailto:zhangzhenyu07@baidu.com}{\texttt{zhangzhenyu07@baidu.com}}\par}%
    \vskip 2pt
    {\color{accTint}\rule{\textwidth}{0.6pt}}%
    \vskip 2pt
  }%
}

\newtcolorbox{accabstractbox}{enhanced, arc=5pt, boxrule=0.5pt,
  colback=accAbsBack, colframe=accAbsFrame,
  left=12pt, right=12pt, top=12pt, bottom=12pt, before skip=4pt, after skip=16pt}
\newcommand{\hfbase}{https://huggingface.co/btzyd/Accordion-6B-A0.5B-Base-8K}
\newcommand{\hfbasethirtytwok}{https://huggingface.co/btzyd/Accordion-6B-A0.5B-Base-32K}
\newcommand{\hfinstruct}{https://huggingface.co/btzyd/Accordion-6B-A0.5B-SFT}
\newcommand{\hfrl}{https://huggingface.co/btzyd/Accordion-6B-A0.5B-RL}
\newcommand*{\projectwebsite}{https://btzyd.github.io/Accordion/}
\newcommand{\reportmeta}{%
  \par\vspace{4pt}%
  {\small
    \hypersetup{urlcolor=accAccent}%
    {\color{accMeta}\bfseries Models:}\enspace
    \href{\hfbase}{\textbf{Accordion-Base}}\quad
    \href{\hfbasethirtytwok}{\textbf{Accordion-Base-32K}}\quad
    \href{\hfinstruct}{\textbf{Accordion-Instruct}}\quad
    \href{\hfrl}{\textbf{Accordion-RL}}\par
    {\color{accMeta}\bfseries Website:}\enspace
    \ifx\projectwebsite\@empty
      {\color{accMeta}\textit{URL to be added}}%
    \else
      \href{\projectwebsite}{\textbf{Project website}}%
    \fi\par}%
}
\renewenvironment{abstract}%
  {\begin{accabstractbox}%
   {\large\bfseries\color{accAccent}Abstract}\par\vspace{4pt}%
   \setlength{\parskip}{5pt plus 1pt minus 0.5pt}}%
  {\reportmeta\end{accabstractbox}}

\renewcommand{\section}{%
  \@startsection{section}{1}{\z@}%
    {-8pt \@plus -1pt \@minus -1pt}%
    {4pt \@plus 1pt \@minus 1pt}%
    {\Large\bfseries\raggedright\color{accAccent}}%
}
\renewcommand{\subsection}{%
  \@startsection{subsection}{2}{\z@}%
    {-6pt \@plus -1pt \@minus -1pt}%
    {3pt \@plus 1pt \@minus 1pt}%
    {\large\bfseries\raggedright\color{accAccent}}%
}
\renewcommand{\subsubsection}{%
  \@startsection{subsubsection}{3}{\z@}%
    {-4pt \@plus -1pt \@minus -1pt}%
    {2pt \@plus 1pt}%
    {\normalsize\bfseries\raggedright}%
}
\renewcommand{\paragraph}{%
  \@startsection{paragraph}{4}{\z@}%
    {4pt \@plus 1pt \@minus 1pt}%
    {-1em}%
    {\normalsize\bfseries}%
}
\makeatother
\AddToHook{begindocument/end}{%
  \newgeometry{textheight=9in, textwidth=6.5in, top=1in, headheight=14pt, headsep=25pt, footskip=30pt}%
  \setlength{\headwidth}{\textwidth}}
\fancypagestyle{plain}{\fancyhf{}\fancyfoot[C]{\small\thepage}}

\title{%
  {\bfseries\fontsize{24}{26}\selectfont\color{accTitle}%
    \raisebox{-0.14em}{\includegraphics[height=0.95em]{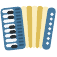}}\hspace{0.32em}%
    Accordion\par}%
  \vskip 5pt
  {\bfseries\fontsize{18}{22}\selectfont\color{accHead}%
    \mbox{Elastic-Depth Inference via Fixed-Prefix Layer Looping}\par}%
}

\author{%
  Yudong Zhang\textsuperscript{1}\quad
  Wei Ding\textsuperscript{1}\quad
  Kaili Zheng\textsuperscript{1}\quad
  Yuchen Ding\textsuperscript{2}\quad
  Zhenyu Zhang\textsuperscript{2,*}%
}
\date{}

\hypersetup{pdftitle={Accordion: Elastic-Depth Inference via Fixed-Prefix Layer Looping},
  pdfauthor={Yudong Zhang, Wei Ding, Kaili Zheng, Yuchen Ding, Zhenyu Zhang}}

\DeclareCaptionFont{accCoverCaption}{\fontsize{10.3}{13}\selectfont}

\begin{document}
\maketitle
\thispagestyle{plain}

\begin{abstract}
Large language models serve applications with diverse inference budgets, but training and maintaining separate models for each budget incurs substantial costs. 
Recent Looped Transformers enable flexible inference computation through parameter reuse, but maintaining strong performance across different inference budgets remains an open challenge. 
We introduce Accordion, a mixture-of-experts-based model that supports five inference budgets through fixed-prefix layer looping. 
Each budget follows a pre-defined execution path, while the minimum budget retains the unlooped model. Joint pretraining samples one budget per micro-batch and optimizes its next-token prediction loss without additional cross-budget consistency objectives. 
We further preserve these inference budgets through supervised fine-tuning and reinforcement learning, where parameter-group merging and budget-matched multi-teacher on-policy distillation integrate three domain experts into one model. 
After 3.6T tokens of pretraining, increasing inference depth improves Accordion-6B-A0.5B's five-category macro-average across 37 tasks by 3.01 percentage points, with the middle budget capturing 89\% of this gain using half as many additional layer executions.
The final model surpasses each domain expert on its corresponding task-and-budget average, while achieving strong instruction-following performance across all budgets and retaining depth-dependent gains in mathematical reasoning and code generation.
\end{abstract}

\captionsetup{font=accCoverCaption,skip=7pt}
\noindent\begin{minipage}{\textwidth}
  \centering
\includegraphics[width=\linewidth]{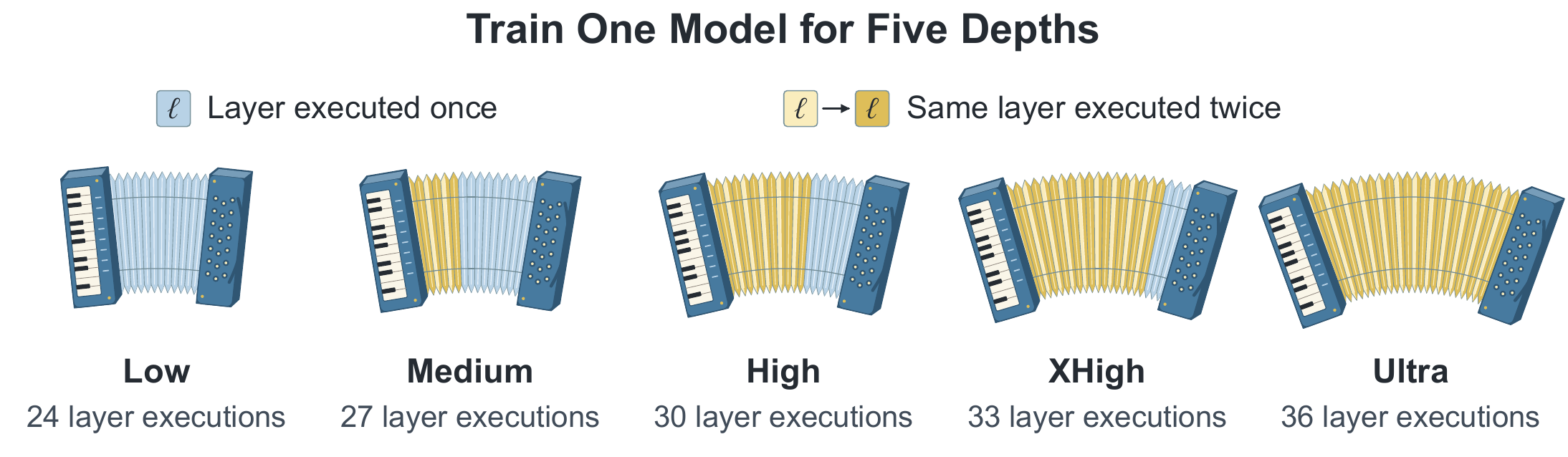}
\captionof{figure}{\textbf{Elastic-depth inference with fixed budget paths.} One set of weights serves five budgets by repeating a prefix of layers 7--18. Blue folds execute once; cream/gold pairs execute the same layer twice.}
\label{fig:accordion}

\end{minipage}

\begingroup
\renewcommand{\thefootnote}{}%
\footnotetext{\textsuperscript{*}\textbf{Correspondence:} Zhenyu Zhang (\href{mailto:zhangzhenyu07@baidu.com}{\texttt{zhangzhenyu07@baidu.com}})}
\endgroup

\newpage
\accUseNeuripsBody
\newcommand{\accOverviewFigure}[1]{%
\begin{figure*}[#1]
  \centering
  \input{figures/overview}
  \caption{\textbf{Fixed budget paths with gated layer looping.} Budgets select a prefix of layers 7--18 for a second execution. Both passes share weights; zero-initialized channel gates scale the second pass before residual addition. Training samples one budget per micro-batch. Inference fixes one budget per request. Blue denotes one execution; cream/gold pairs denote two.}
  \label{fig:overview}
\end{figure*}%
}

\afterpage{%
\begin{figure*}[!t]
  \centering
  \input{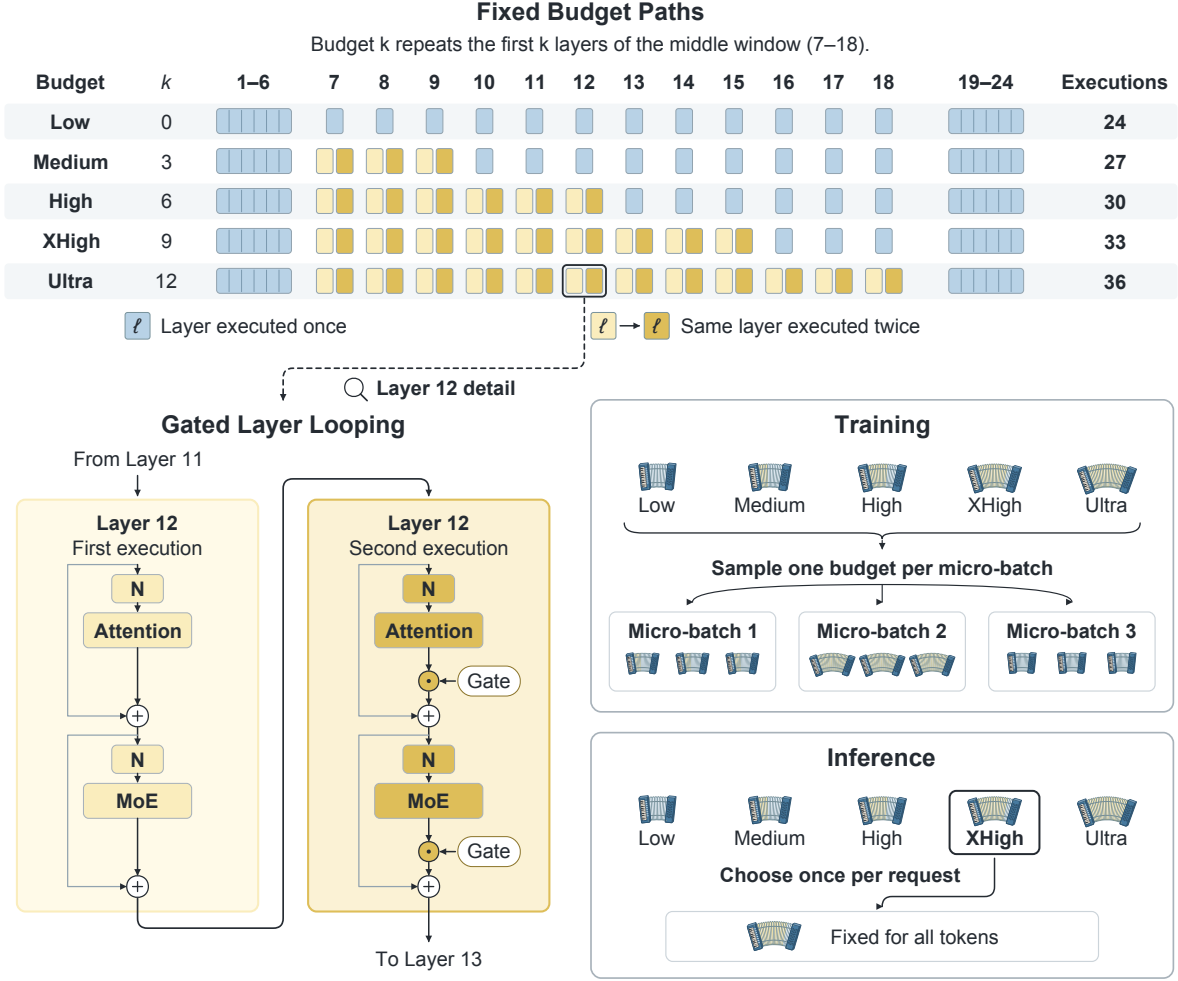}
  \caption{\textbf{Fixed budget paths with gated layer looping.} Budgets select a prefix of layers 7--18 for a second execution. Both passes share weights; zero-initialized channel gates scale the second pass before residual addition. Training samples one budget per micro-batch. Inference fixes one budget per request. Blue denotes one execution; cream/gold pairs denote two.}
  \label{fig:overview}
\end{figure*}%
}
\newcommand{\accOverviewPlacedEarly}{}
\begingroup
\setlength{\parskip}{3pt plus 1pt minus 0.5pt}
\section{Introduction}

Language-model applications require different inference budgets, but training and maintaining separate models for each budget incurs substantial costs.
Elastic inference aims to address this challenge by supporting multiple compute choices within a single model.
Existing approaches achieve this flexibility through nested architectures and subnetworks~\citep{devvrit2023matformer,cai2024flextron,taghibakhshi2025nemotronelastic,taghibakhshi2026starelastic}, reduced execution depth \citep{fan2020layerdrop,schuster2022calm,elhoushi2024layerskip}, or input-dependent computation allocation \citep{raposo2024mod,bae2025mor}.

Looped Transformers provide another approach to elastic inference by repeatedly applying shared parameters to increase effective depth~\citep{dehghani2019universal,lan2020albert,giannou2023looped}.
Recent work explores recurrence structures and training strategies for flexible inference computation~\citep{geiping2025huginn,zhu2025ouro,alabdulmohsin2025rins,bae2025mor,prairie2026parcae}.
However, jointly optimizing multiple inference depths can compromise shallow-path performance, as highlighted by the \emph{one-loop tax} problem~\citep{ladhak2026elr}.
Moreover, practical language models require supervised fine-tuning and reinforcement learning, so elastic inference should remain effective throughout the training pipeline.

To address these challenges, we introduce \textbf{Accordion}, a mixture-of-experts (MoE) Transformer with five fixed inference paths sharing one set of weights (Figure~\ref{fig:accordion}). 
Building on layer looping from Loopie~\citep{gao2026loopie}, Accordion repeats selected Transformer layers consecutively rather than repeating an entire model.
Each budget repeats a fixed prefix within a middle window, concentrating training on the inference paths. Larger budgets add layer executions; the minimum budget retains the unlooped path.
Zero-initialized residual gates control the repeated computations. During pretraining, each micro-batch uniformly samples one budget and optimizes its corresponding path with next-token prediction, without additional cross-budget consistency objectives.

We further extend Accordion to supervised fine-tuning (SFT) and reinforcement learning (RL). SFT continues to jointly optimize the same five inference budgets. 
For RL, we train three domain experts for instruction following, mathematical reasoning, and code generation, each supporting all five budgets. 
Parameter-group merging combines their updates, then multi-teacher on-policy distillation matches each teacher to the student's task domain and inference budget.

In controlled experiments with 80B training tokens, fixed-prefix layer selection achieves the lowest observed validation loss at all four looped budgets among eight evaluated strategies. 
Compared with independently trained budget-specific specialists, Accordion achieves validation losses within 0.003 nats at all three evaluated budgets under equal training tokens, including the unlooped configuration. We further scale Accordion to a 6B-A0.5B MoE model with 3.6T tokens of pretraining at 8K context. Increasing inference depth improves 36 of 37 benchmark scores and raises the five-category average by 3.01 percentage points. The middle budget captures 89\% of this gain with half as many additional layer executions. A further 50B tokens produce Accordion-Base-32K for long-context use (\S\ref{sec:context-extension}); its full benchmark table appears in Appendix~\ref{app:base32k-results}.

Accordion-Instruct retains positive depth-dependent gains on most conversational benchmarks. After domain-specific RL and expert integration, Accordion-RL exceeds each domain expert on its corresponding task-and-budget average. The final model also achieves strong instruction-following performance across all five budgets, outperforming four evaluated public reference models on IFEval. Additional depth continues to improve mathematical reasoning and code generation, although the optimal budget varies across tasks.

\Needspace{6\baselineskip}
Our contributions are:
\begin{itemize}[leftmargin=*,nosep]
  \item \textbf{One model, five inference budgets.} We introduce Accordion, an elastic-depth MoE Transformer that supports five inference budgets through fixed-prefix layer looping. Under equal training tokens, it closely matches budget-specific specialists at three evaluated budgets, including the unlooped path.
  \item \textbf{Depth gains from pretraining through RL.} We extend elastic-depth inference through SFT and RL, integrating three domain experts via parameter-group merging and budget-matched multi-teacher on-policy distillation. Additional depth improves base-model quality and retains benefits in mathematics and code after post-training.
  \item \textbf{A public-data recipe at scale.} We develop Accordion-Base, Accordion-Instruct and Accordion-RL using public datasets across the training pipeline, with detailed data accounting and comprehensive task-level evaluations.
\end{itemize}

\par\endgroup
\providecommand{\accEvidenceBarrier}{\FloatBarrier}
\section{Accordion}
\label{sec:method}

\subsection{Architecture}
\label{sec:architecture}

Consider a pre-norm MoE transformer with $L$ physical layers and hidden dimension $d$.
Let $\mathcal{W}=\{w_1,\ldots,w_m\}$ be an ordered, contiguous middle window, with $w_1<\cdots<w_m$, and let $\mathcal{K}\subseteq\{0,\ldots,m\}$ be the budget menu.
Budget $k$ repeats the prefix $\mathcal{S}_k=\{w_1,\ldots,w_k\}$, with $\mathcal{S}_0=\varnothing$.
Its path $P_k$ executes layer $\ell$ exactly
\begin{equation}
  r_\ell(k)=1+\mathbf{1}\{\ell\in\mathcal{S}_k\}
  \label{eq:executions}
\end{equation}
times before moving to layer $\ell+1$.
Each budget therefore defines a deterministic path with $L+k$ layer executions.
We use $L=24$, $\mathcal{W}=\{7,\ldots,18\}$, and $\mathcal{K}=\{0,3,6,9,12\}$; the corresponding execution counts appear in Table~\ref{tab:cost}.
Layers outside the window always run once, and $P_0$ is an ordinary unlooped MoE.
The menu size and prefix mapping are evaluated in \S\ref{sec:budget-count} and \S\ref{sec:layer-selection}, respectively; the same formulation supports other predefined budget menus.

\ifdefined\accOverviewPlacedEarly
\else

\begin{figure*}[!htbp]
  \centering
  \input{figures/overview}
  \caption{\textbf{Fixed budget paths with gated layer looping.} Budgets select a prefix of layers 7--18 for a second execution. Both passes share weights; zero-initialized channel gates scale the second pass before residual addition. Training samples one budget per micro-batch. Inference fixes one budget per request. Blue denotes one execution; cream/gold pairs denote two.}
  \label{fig:overview}
\end{figure*}%

\fi

At each training micro-batch $\mathcal{B}$, we sample $k\sim\mathrm{Uniform}(\mathcal{K})$ and run only $P_k$ for all sequences in that micro-batch. The sampled budget is shared across all data-parallel and expert-parallel workers. For the main model, each optimizer step contains one micro-batch per data-parallel worker, so one budget is shared by the entire global batch.
All data-parallel and expert-parallel workers derive $k$ from a common random seed and micro-batch counter.
This synchronized sampling follows Huginn~\citep{geiping2025huginn} and keeps MoE all-to-all communication aligned across workers without an additional synchronization message.
For $N$ prediction targets, its next-token loss $\mathcal{L}_{\mathrm{NTP}}(\theta;\mathcal{B},k)$ is
\begin{equation}
  \mathcal{L}_{\mathrm{NTP}}=-\frac{1}{N}\sum_{x\in\mathcal{B}}\sum_t
  \log p_\theta(x_t\mid x_{<t};P_k),
  \label{eq:ntp}
\end{equation}
where $\theta$ contains the shared model weights and gates.
Together with the sequence-level MoE balancing term described below, the training objective is
\begin{equation}
  \mathcal{J}(\theta)=\mathbb{E}_{\mathcal{B},k}\!\left[
    \mathcal{L}_{\mathrm{NTP}}+\lambda_{\mathrm{seq}}\mathcal{L}_{\mathrm{seq}}
  \right].
  \label{eq:joint-objective}
\end{equation}
Here $\lambda_{\mathrm{seq}}=10^{-4}$.
Uniform sampling gives every supported path equal probability during training.
One forward and backward pass through the sampled path provides a stochastic estimate of the joint objective, without evaluating the other paths on that micro-batch.

Each selected layer executes twice in succession, following the layer-looping structure of Loopie (Figure~\ref{fig:overview}).
The repeated unit is an individual layer, rather than a multi-layer block or the full model stack (Appendix Figure~\ref{fig:compare}).
Let $A_\ell$, $M_\ell$, $\mathrm{N}^{A}_\ell$, and $\mathrm{N}^{M}_\ell$ denote attention, MoE, and their respective pre-normalizations. We write $M_\ell(z;b_\ell^{(p)})$ for the shared MoE applied to $z$ with the balancing bias of pass $p\in\{1,2\}$.
For layer input $h$, the first pass is
\begin{align}
  u &= h + A_\ell(\mathrm{N}^{A}_\ell(h)), \nonumber\\
  v &= u + M_\ell(\mathrm{N}^{M}_\ell(u);b_\ell^{(1)}).
  \label{eq:first-pass}
\end{align}
If $\ell\notin\mathcal{S}_k$, the layer returns $v$.
Otherwise, it immediately applies a second pass:
\begin{align}
  u_2 &= v + \tanh(g^{A}_\ell)\odot A_\ell(\mathrm{N}^{A}_\ell(v)), \nonumber\\
  v_2 &= u_2 + \tanh(g^{M}_\ell)\odot M_\ell(\mathrm{N}^{M}_\ell(u_2);b_\ell^{(2)}).
  \label{eq:second-pass}
\end{align}
The layer returns $v_2$.
Its per-channel gates $g^{A}_\ell,g^{M}_\ell\in\mathbb{R}^d$ are initialized to zero and excluded from weight decay.
At initialization, trainable gates make the second pass an identity map.
The hyperbolic tangent bounds the magnitude of each gate by one and has unit derivative at zero.
Attention, experts, normalizations, and the learned router are shared across passes and budgets; gates are layer-specific and shared across budgets.
We use no second-pass embedding or budget-conditioned normalization.
Appendix~\ref{app:training-analysis} analyzes second-pass computation, and \S\ref{sec:second-pass-variants} evaluates budget-conditioned parameter variants.

The two passes apply the same learned routing function to their respective hidden states and may select overlapping sets of experts.
Following the auxiliary-loss-free balancing mechanism of DeepSeek-V3~\citep{deepseek2024v3}, we maintain separate expert-load statistics and balancing biases for the first and second passes.
These pass-specific balancing states are distinct from the shared learned router weights.
We retain the small sequence-level loss $\mathcal{L}_{\mathrm{seq}}$ in Equation~\ref{eq:joint-objective} to encourage balanced expert usage within each sequence.
This term is computed for routing on both the first pass and any executed second pass within the sampled path, and requires no comparison across budgets.
Routing variants are reported in \S\ref{sec:second-pass-routing}, and budget-conditioned variants in \S\ref{sec:second-pass-variants}.
Activation recomputation uses the original sampled path and the corresponding first- or second-pass balancing state.

The user selects $k\in\mathcal{K}$ for each request, and every token follows the corresponding trained path $P_k$ with the same model weights.
Each budget has a static execution graph, so depth remains fixed throughout the request.
The five settings are called Low, Medium, High, XHigh, and Ultra, respectively.

Looping creates no copies of layer weights; relative to the underlying MoE, the gates add only $2md$ effective learned parameters, independent of the selected budget. Checkpoints allocate $2Ld$ gate entries across all physical layers, but gates outside the loopable window remain zero and are never used.
For equal-shaped layers at a fixed sequence length, neglecting the small gating overhead, floating-point operations (FLOPs) are approximately
\begin{equation}
  C(k)\simeq C_{\mathrm{fixed}}+(L+k)C_{\mathrm{layer}},
  \label{eq:cost}
\end{equation}
where $C_{\mathrm{fixed}}$ includes non-repeated embedding and output work.
The same affine dependence applies to model training FLOPs when forward and backward costs are included at a fixed sequence length.
Our uniform menu gives $\mathbb{E}[k]=6$ and hence 30 expected layer executions per micro-batch; the arithmetic cost follows Equation~\ref{eq:cost}.
\S\ref{sec:tax} uses this expectation to match training computation to budget-specific specialists, separately from matching training tokens.
During autoregressive inference, the second attention pass maintains a separate key--value (KV) cache built from its own activations.
Consequently, for our equal-shaped layers, KV storage scales as $(L+k)/L$ relative to $k=0$: $1.25\times$ at $k=6$ and $1.5\times$ at $k=12$.
The analytical costs of \mainmodel{} are given in Table~\ref{tab:cost}; measured runtime also depends on communication and kernel efficiency.

We use the fixed budget paths defined above with top-8 expert routing and a reduced-vocabulary DeepSeek-V3 tokenizer with 128,000 entries and untied input and output embeddings. Attention uses query--key layer normalization. The router applies sigmoid scores and normalizes the selected top-eight weights; pass-specific balancing biases use an update rate of $10^{-3}$.

\subsection{Designing Strong Shared Paths}
\label{sec:shared-path-design}
\label{sec:design}
\label{sec:experimental-setup}
\begingroup
\accLocalParskip{8pt plus 1pt minus 1pt}
We use a 1.61B-parameter MoE to choose how layers and parameters should be shared across inference budgets, then test how closely one shared model matches separately trained specialists.
The design comparisons address four choices: which layers to repeat, how many budgets to train, whether to share the router across passes, and whether to give each budget its own scaling parameters.
All comparisons below use validation loss on the same held-out web set. Each configuration is trained once; the comparisons report its held-out validation loss.
Shared models receive 80B tokens of non-repeated Nemotron-CC-v2 web data, with budgets sampled uniformly unless stated otherwise.
Specialists receive the same token budget in the equal-token comparison; their token counts change only for the FLOPs-matched comparison in \S\ref{sec:tax}.
Appendix Table~\ref{tab:design-model-settings} gives the complete architecture, optimization settings and hardware for these design studies.

\subsubsection{Layer Selection}
\label{sec:layer-selection}
\label{sec:where}
\label{sec:recipe}
We test whether shared inference paths benefit from a fixed layer-selection rule during training or from randomizing the repeated layers across steps.
The eight strategies comprise four fixed and four randomized selection rules, all evaluated at $k\in\{0,3,6,9,12\}$.
They have the same expected training depth of 30 layer executions.
At inference, fixed strategies retain their own selection rules, whereas all randomized strategies use the prefix paths of window 7--18.
The comparison therefore evaluates each training rule together with its specified inference paths.
\textbf{Fixed selection.}
The fixed rules compare the location of the repeated segment.
Within window 7--18, \emph{fixed prefix}, \emph{fixed center} and \emph{fixed suffix} repeat the first, central or last $k$ layers, respectively.
Fixed center uses layers 11--13, 10--15 and 8--16 at $k=3,6,9$, respectively.
A fourth rule, \emph{fixed prefix (4--15)}, moves the window earlier while retaining prefix selection.

\textbf{Randomized selection.}
All four randomized rules use window 7--18.
\emph{Random layers} uniformly samples $k$ individual layers; \emph{random segment} samples a prefix, center or suffix of length $k$; \emph{random blocks} selects $k/3$ of four disjoint three-layer blocks, which need not be adjacent.
These three rules repeat exactly $k$ layers after sampling the budget.
\emph{Layerwise coin} uses the unlooped and fully looped paths with probability 30\% each; on the remaining 40\% of steps, each window layer is independently repeated with probability $0.5$.
This changes the training-budget distribution while preserving the same expected depth.
\emph{Fixed prefix (7--18)} achieves the lowest observed validation loss at all four looped budgets and the lowest five-budget mean (Figure~\ref{fig:layer-selection}).
Its mean advantage over the alternatives is approximately 0.0025--0.0041 nats.
This consistent ranking motivates the fixed-prefix paths used by \method{}.

\begin{figure*}[!htbp]
  \centering
  \includegraphics[width=5.8in]{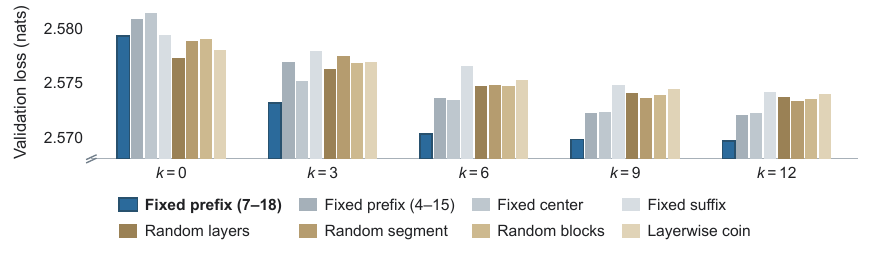}
  \caption{\textbf{Comparison of eight layer-selection strategies.} Accordion's fixed prefix over layers 7--18 is highlighted in blue. Strategies are compared at matched inference budgets.}
  \label{fig:layer-selection}
\end{figure*}

\subsubsection{Number of Training Budgets}
\label{sec:budget-count}
We next test whether offering more inference budgets compromises the quality of the shared paths.
Using the fixed-prefix rule in window 7--18, we train models with three, four or five budgets (Table~\ref{tab:budget-count}).
All three budget sets have $\mathbb{E}[k]=6$, keeping expected training computation fixed.
We compare all models at the shared endpoints, $k=0$ and $k=12$, and compare the three- and five-budget models at $k=6$; the four-budget model is not trained at $k=6$.
The five-budget model has the lowest observed loss at both shared endpoints and lower loss at $k=6$ than the three-budget model.
We therefore use five budgets to provide finer inference control at the same expected training computation.

\begin{table*}[!htbp]
  \centering
\accTableStretch{1.12}
  \caption{\textbf{Comparison of training-budget sets.} Validation losses are in nats; the mean uses the common endpoints $k=0,12$. The four-budget model was not trained at $k=6$. Bold marks the lowest loss in each column.}
  \label{tab:budget-count}
  \accTableFont{10.5}{13}
  \setlength{\tabcolsep}{5pt}
\begin{tabular*}{\textwidth}{@{\hspace{12pt}\extracolsep{\fill}}cccccc@{\hspace{12pt}}}
\toprule
Model & Training budgets & $k=0$ & $k=6$ & $k=12$ & Mean \\
\midrule
Five-budget model & $\{0,3,6,9,12\}$ & \textbf{2.5793} & \textbf{2.5703} & \textbf{2.5696} & \textbf{2.5744} \\
Three-budget model & $\{0,6,12\}$ & 2.5810 & 2.5729 & 2.5721 & 2.5766 \\
Four-budget model & $\{0,4,8,12\}$ & 2.5814 & -- & 2.5719 & 2.5767 \\
\bottomrule
\end{tabular*}
\end{table*}

\subsubsection{Router Sharing}
\label{sec:second-pass-routing}
We test whether the second pass needs a separately learned router to select experts from its updated hidden states.
We compare a shared router, a separate second-pass router, and a separate router that excludes routed experts selected in the first pass.
All three configurations share expert parameters and retain their respective routing rules during evaluation.
The shared router achieves the lowest five-budget mean loss, improving on separate routing by 0.0013 nats and on separate routing with exclusion by 0.0017 nats (Table~\ref{tab:negative}).
A shared router therefore serves every pass and inference budget in \method{}.

\begin{table*}[!htbp]
\centering
\accTableStretch{1.12}
\caption{\textbf{Second-pass routing variants.} Configurations are rows and inference budgets are columns. Validation losses are in nats; bold identifies the lowest loss within each column.}
\label{tab:negative}
\accTableFont{10.5}{13}
\setlength{\tabcolsep}{2pt}
\begin{tabular*}{\textwidth}{@{\hspace{12pt}\extracolsep{\fill}}ccccccc@{\hspace{12pt}}}
\toprule
Routing & $k=0$ & $k=3$ & $k=6$ & $k=9$ & $k=12$ & Mean \\
\midrule
Shared router & \textbf{2.5793} & \textbf{2.5731} & \textbf{2.5703} & \textbf{2.5698} & 2.5696 & \textbf{2.5724} \\
Separate router & 2.5814 & 2.5746 & 2.5715 & 2.5705 & 2.5702 & 2.5737 \\
Separate router + exclusion & 2.5823 & 2.5764 & 2.5718 & 2.5704 & \textbf{2.5694} & 2.5741 \\
\bottomrule
\end{tabular*}
\end{table*}

\subsubsection{Budget-Conditioned Parameters}
\label{sec:second-pass-variants}
We test whether different inference budgets benefit from their own parameters for scaling the second-pass updates.
The baseline shares both gates and normalization parameters across budgets.
Two alternatives condition either the gates or the normalization scales on $k$, leaving the other parameter type budget-independent (Table~\ref{tab:budget-parameters}).
\emph{Budget-conditioned gates} let each budget learn its own attention and MoE residual-update scaling: $\tanh(g_\ell^X)$ becomes $\tanh(g_{\ell,k}^X)$ for $X\in\{A,M\}$.
\emph{Budget-conditioned normalization scales} instead multiply second-pass root-mean-square normalization (RMSNorm) outputs by $(1+s^X_{\ell,k})$.
These scales start at $s^X_{\ell,k}=0$, and the original gates are retained.
The baseline with shared gates and normalization parameters achieves the lowest observed loss at every evaluated budget.
We therefore retain a single budget-independent parameterization for \method{}.

\begin{table*}[!htbp]
  \centering
\accTableStretch{1.12}
  \caption{\textbf{Comparison of budget-conditioned parameters.} Each row identifies which parameter type is budget-conditioned; the other type remains budget-independent. Validation loss is in nats (lower is better). Bold marks the lowest loss within each column.}
  \label{tab:budget-parameters}
  \accTableFont{10.5}{13}
  \setlength{\tabcolsep}{3.5pt}
\begin{tabular*}{\textwidth}{@{\hspace{12pt}\extracolsep{\fill}}ccccccc@{\hspace{12pt}}}
\toprule
Configuration & $k=0$ & $k=3$ & $k=6$ & $k=9$ & $k=12$ & Mean \\
\midrule
Budget-independent gates and norms & \textbf{2.5793} & \textbf{2.5731} & \textbf{2.5703} & \textbf{2.5698} & \textbf{2.5696} & \textbf{2.5724} \\
Budget-conditioned gates & 2.5814 & 2.5761 & 2.5731 & 2.5723 & 2.5721 & 2.5750 \\
Budget-conditioned norm scales & 2.5830 & 2.5769 & 2.5735 & 2.5724 & 2.5722 & 2.5756 \\
\bottomrule
\end{tabular*}
\end{table*}

\subsubsection{Comparison with Budget-Specific Specialists}
\label{sec:tax}

We finally test how closely one shared model matches a separate model trained for each inference budget.
The specialists are trained independently at fixed $k=0,6,12$.
At each budget, the shared model and its corresponding specialist use the same inference graph.
We compare them under two resource constraints: equal training tokens hold data exposure fixed, while equal training FLOPs hold modeled training computation fixed (Figure~\ref{fig:tradeoff}).
A further equal-token control tests whether allocating more training examples to the unlooped path improves its validation loss.

\begin{figure*}[!htbp]
  \centering
  \input{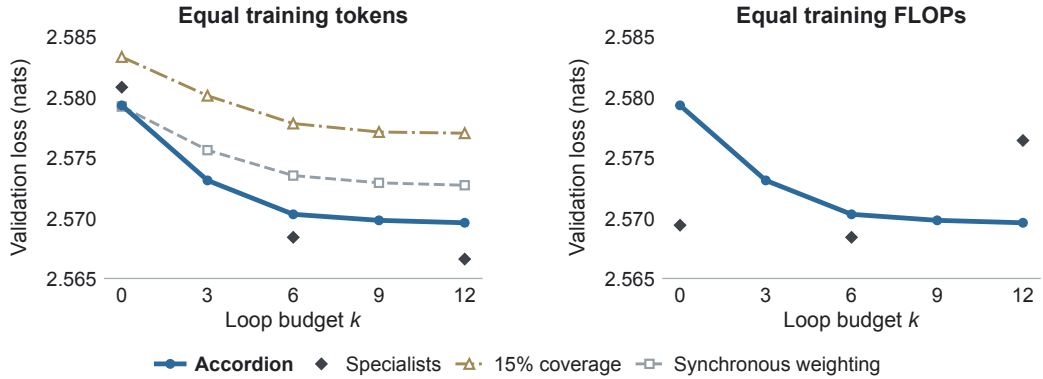}
  \caption{\textbf{Shared models and budget-specific specialists.} Accordion uses uniform budget sampling. Comparisons match training tokens (80B, left) or training FLOPs (right). Each specialist is trained for a single budget. The left panel also includes two controls with increased unlooped-path exposure, defined below.}
  \label{fig:tradeoff}
\end{figure*}

\paragraph{Equal training tokens.}
\label{sec:coverage}
Each model receives 80B training tokens.
Under this constraint, \method{} achieves validation losses within 0.0030 nats of the specialists at all three evaluated budgets (Figure~\ref{fig:tradeoff}, left).
One set of weights thus approximates three specialist operating points while preserving the unlooped path's quality in this comparison.
The $k=0$ comparison addresses the \emph{one-loop tax} discussed by \citet{ladhak2026elr}, where training for recurrent computation degrades single-pass performance.
To test whether the unlooped path needs additional training exposure, a control reserves 15\% of data-parallel replicas for that path.
The synchronous-weighting control instead samples one common budget for all replicas, with probabilities $0.32/0.17/0.17/0.17/0.17$ for $k=0,3,6,9,12$, matching the coverage control's expected marginal distribution. Both controls are token-matched and average 29.1 layer executions, compared with 30 under uniform budget sampling. Uniform budget sampling yields a mean validation loss lower by 0.0067 nats than the replica-coverage control, while retaining single-pass performance without the reserved replicas.

\paragraph{Equal training FLOPs.}
Here we keep the shared model's 80B-token training run fixed and adjust the number of tokens seen by each specialist to match its expected training FLOPs.
A shallower specialist can therefore receive more tokens, while a deeper specialist receives fewer.
Let $C_{\mathrm{train}}(k)$ denote training FLOPs per token at budget $k$.
Under the approximately affine cost model in Equation~\ref{eq:cost}, uniform sampling over the five budgets gives an expected cost of $C_{\mathrm{train}}(6)$ per token.
The specialist at budget $k$ receives approximately
$80\mathrm{B}\,C_{\mathrm{train}}(6)/C_{\mathrm{train}}(k)$
tokens.
The $k=6$ specialist satisfies both resource constraints and is reused in the two comparisons.
At this middle budget, the shared model remains close to the specialist under both matching conditions.
At the largest budget, it improves on the FLOPs-matched specialist by 0.0068 nats.

\par\endgroup
\subsection{Training Pipeline}
\label{sec:training-pipeline}
\begin{figure*}[!htbp]
  \centering
  \input{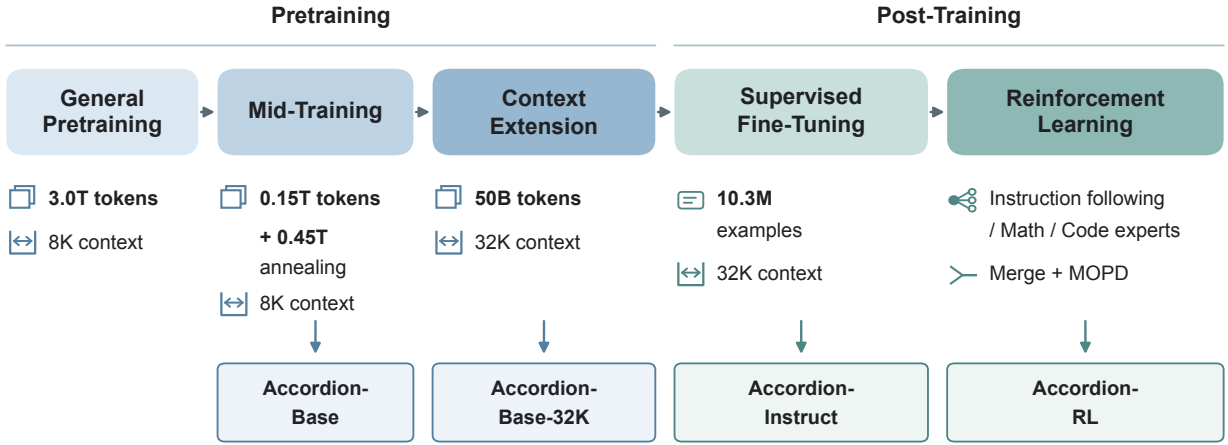}
  \caption{\textbf{Training pipeline for \mainmodel{}.} Accordion-Base is exported after 3.6T tokens at 8K context; another 50B tokens produce Accordion-Base-32K. SFT and RL continue from this extended model and retain all five budget paths. RL combines three domain experts through parameter merging and budget-matched multi-teacher distillation.}
  \label{fig:training-pipeline}
\end{figure*}

\mainmodel{} retains the same five budget paths throughout pretraining and post-training (Figure~\ref{fig:training-pipeline}).
General pretraining and mid-training with annealing produce Accordion-Base at 8K context; a subsequent context-extension stage produces Accordion-Base-32K (\S\ref{sec:pretraining}).
Post-training comprises supervised fine-tuning (SFT) and reinforcement learning (RL) (\S\ref{sec:post-training}).
SFT produces Accordion-Instruct from Accordion-Base-32K.
RL trains independent instruction-following, mathematics and code experts from Accordion-Instruct, then combines them through model merging and budget-matched multi-teacher distillation.
Sections~\ref{sec:pretraining-hparams} and~\ref{sec:rl-expert-training} describe stage-specific optimization, including the expert update and distillation rules.

\FloatBarrier

\section{Pretraining}
\label{sec:pretraining}
\label{sec:experiments}
\newcommand{\pretrainPoolT}{9.57}
\newcommand{\pretrainMaxRepeat}{1.67}
\newcommand{\pretrainGeneralShares}{38.25/22/9/11.5/19.25}
\newcommand{\pretrainMidShares}{24/17/18/14.5/26.5}
\newcommand{\pretrainAnnealShares}{16/12/25/17/30}
\newcommand{\pretrainExtensionShares}{24/6/22.5/18.5/29}

\definecolor{accModelBlue}{HTML}{477A9F}
\definecolor{accTableOurs}{HTML}{EDF4FA}
\definecolor{accTableAccent}{HTML}{315F7D}
\colorlet{accRankFirst}{accModelBlue!24!white}
\colorlet{accRankSecond}{accModelBlue!12!white}
\definecolor{accCategoryKnowledge}{HTML}{C4A57D}
\colorlet{accRankKnowledgeFirst}{accCategoryKnowledge!30!white}
\colorlet{accRankKnowledgeSecond}{accCategoryKnowledge!15!white}
\definecolor{accCategoryCommon}{HTML}{75A894}
\colorlet{accRankCommonFirst}{accCategoryCommon!30!white}
\colorlet{accRankCommonSecond}{accCategoryCommon!15!white}
\definecolor{accCategoryReasoning}{HTML}{A994C0}
\colorlet{accRankReasoningFirst}{accCategoryReasoning!30!white}
\colorlet{accRankReasoningSecond}{accCategoryReasoning!15!white}
\definecolor{accCategoryMath}{HTML}{C5A04B}
\colorlet{accRankMathFirst}{accCategoryMath!30!white}
\colorlet{accRankMathSecond}{accCategoryMath!15!white}
\definecolor{accCategoryCode}{HTML}{C98B8B}
\colorlet{accRankCodeFirst}{accCategoryCode!30!white}
\colorlet{accRankCodeSecond}{accCategoryCode!15!white}
\definecolor{accCategoryOverall}{HTML}{8796A2}
\colorlet{accRankOverallFirst}{accCategoryOverall!30!white}
\colorlet{accRankOverallSecond}{accCategoryOverall!15!white}
\newcommand{\accTableLogo}{\raisebox{-0.08em}[0pt][0pt]{\includegraphics[height=0.85em]{figures/icons/accordion-blue.pdf}}\hspace{0.3em}}
\providecommand{\pending}{\textit{pending}}

Accordion-Base turns additional layer executions into broad capability gains, with the middle budget capturing most of the improvement. We describe the training curriculum and evaluate these gains across tasks, inference costs and context lengths.

\subsection{Data}
\label{sec:pretraining-data}
We construct the pretraining curriculum from publicly released datasets covering web text, mathematics, code and reasoning.
The corpus combines Nemotron pretraining collections~\citep{su2024nemotroncc}, MegaMath~\citep{zhou2025megamath}, FineMath~\citep{allal2025smollm2}, InfiMM-WebMath~\citep{han2024infimm} and StarCoder data~\citep{li2023starcoder}.
The selected training pool contains 835 shards and approximately 4.68T tokens, from which we sample the 3.6T-token curriculum.
Within each source category, shards are sampled across document-length strata; smaller sources are shared across stages with different data seeds.
Appendix Table~\ref{tab:mixture} lists source weights and expected consumption, with accounting details in Appendix~\ref{app:pretraining-evidence}.
We tokenize the data as specified in \S\ref{sec:architecture}.
Reasoning collections are rendered as plain text without role markers, retaining reasoning traces and their delimiters.
Held-out web (Nemotron-CC v2 HQ), mathematics (FineMath-3+) and code (StarCoder Python) shards are excluded from training.

\begin{figure*}[!htbp]
  \centering
  \includegraphics[width=\textwidth]{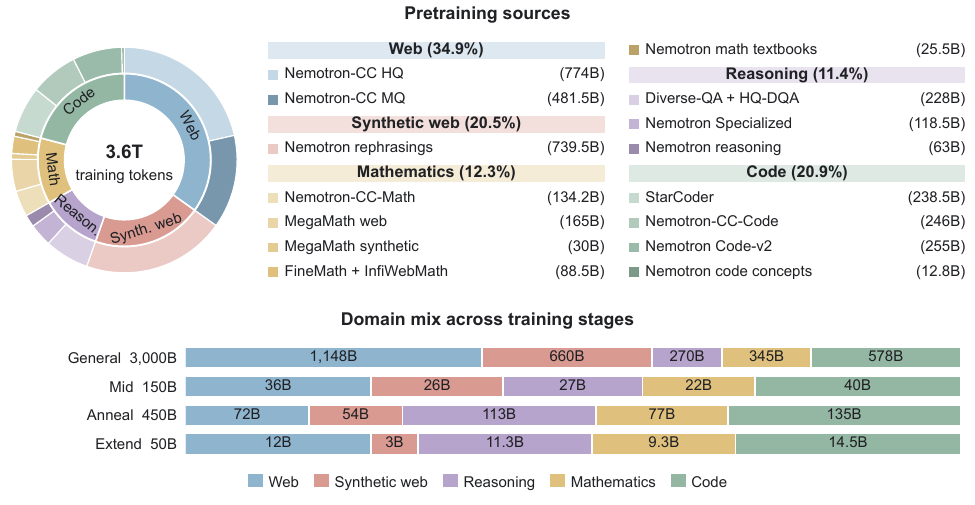}
  \caption{\textbf{Pretraining data composition and curriculum.} Estimated token consumption by domain and source across the 3.6T-token curriculum, followed by 50B-token context extension. Stage bars are normalized to their token budgets; labels are rounded billions of tokens. Exact source quantities appear in Table~\ref{tab:mixture}.}
  \label{fig:pretraining-data}
\end{figure*}

The 3.6T-token curriculum progressively increases exposure to mathematics, code and reasoning (Figure~\ref{fig:pretraining-data}).
General pretraining uses 3.0T tokens, followed by 150B tokens of mid-training and 450B tokens of annealing.
Mid-training introduces the Nemotron reasoning collection (Math-v2, Competitive-Programming and PrismMath) and removes the MegaMath synthetic component.
Annealing keeps these source choices, shifts more weight toward mathematics, code and reasoning, and decays the learning rate.
Table~\ref{tab:mixture} gives the source-level proportions at each stage.
The stage comparisons in \S\ref{sec:training-trajectory} describe this combined curriculum and optimization schedule.
The largest ratio of expected token consumption to available source tokens is approximately \pretrainMaxRepeat{}, excluding context-extension replay.

An additional 50B tokens extend the training context to 32K while maintaining exposure to the annealing distribution.
Half of this budget comes from publicly released long documents of length 16K--128K, covering web text, code, reasoning and mathematical web text.
The other half replays the annealing mixture.
Appendix Table~\ref{tab:context-mixture} gives the allocations and resulting domain proportions.

\subsection{Hyperparameters}
\label{sec:pretraining-hparams}
\textbf{Model configuration.} Our main model, \mainmodel{}, has 6.03B total and 0.54B active parameters. Active counts describe the transformer layers in one unlooped pass and exclude the input embedding and output head; including the output head raises the active count to 0.74B. Table~\ref{tab:hparams} gives its complete architecture and training configuration.

\textbf{Optimization.} All stages train the same five paths using the single-trajectory objective and uniform micro-batch budget sampling defined in \S\ref{sec:architecture}.
We optimize all model weights with AdamW~\citep{loshchilov2019adamw}, excluding second-pass gates from weight decay.
The peak learning rate is $5\times10^{-4}$, chosen from the 1.61B-parameter design studies.
Table~\ref{tab:pretraining-stages} details the stage-specific schedules.
Training uses our layer-looping extension of Megatron-LM~\citep{shoeybi2019megatron}, with expert parallelism within each eight-GPU node.
The first three stages use 8K context and 1,024 sequences per global batch, continuing the optimizer state and step counter across stage boundaries.

\begin{table*}[!htbp]
  \centering
\accTableStretch{1.12}
  \caption{\textbf{Pretraining hyperparameters by stage.} All stages use 8,388,608 tokens per global batch on 512 H800 GPUs. Token budgets are nominal, warmup is measured in optimizer steps, and arrows indicate linear learning-rate decay after warmup.}
  \label{tab:pretraining-stages}
  \accTableFont{10.5}{13}
  \setlength{\tabcolsep}{3pt}
  \begin{tabular*}{\textwidth}{@{\hspace{12pt}\extracolsep{\fill}}*{6}{c}@{\hspace{12pt}}}
    \toprule
    Stage & Tokens (B) & Cumulative & Context & Learning rate & Warmup / optimizer \\
    \midrule
    General & 3,000 & 3.00T & 8K & $5\!\times\!10^{-4}$, constant & 2,000 / initialize \\
    Mid-training & 150 & 3.15T & 8K & $5\!\times\!10^{-4}$, constant & 0 / continue \\
    Annealing & 450 & 3.60T & 8K & $5\!\times\!10^{-4}\!\to\!5\!\times\!10^{-5}$ & 0 / continue \\
    Context extension & 50 & 3.65T & 32K & $5\!\times\!10^{-5}\!\to\!1.5\!\times\!10^{-5}$ & 200 / reset \\
    \bottomrule
  \end{tabular*}
\end{table*}

\begin{table*}[!htbp]
  \centering
  \accTableStretch{1.12}
  \caption{\textbf{Architecture and training configuration of \mainmodel{}.} Active parameters refer to a single unlooped forward pass and exclude the input embedding and output head. Training settings describe the 3.6T-token, 8K curriculum; Table~\ref{tab:pretraining-stages} details the stage schedules and subsequent 32K context extension. WSD denotes warmup--stable--decay.}
  \label{tab:hparams}
  \accTableFont{10.5}{13}
  \setlength{\tabcolsep}{5pt}
  \begin{tabular*}{\textwidth}{@{\extracolsep{\fill}}cc@{}}
    \toprule
    Setting & Value \\
    \midrule
    Layers / hidden size / Q / KV / head dim & 24 / 1536 / 12 / 4 / 128 \\
    Routed / shared / active experts; intermediate size & 128 / 1 / 8; 384 \\
    Total / active parameters; vocabulary & 6.03B / 0.54B; 128{,}000 \\
    Loopable window / budgets & 7--18 / $\{0,3,6,9,12\}$ \\
    \midrule
    Training tokens / context / tokens per step & 3.6T / 8K / 8.39M \\
    Peak LR / final LR / schedule & $5\times10^{-4}$ / 10\% of peak / WSD \\
    Warmup / decay & 2{,}000 steps / final 12.5\% \\
    AdamW $\beta$ / weight decay / gradient clipping & $(0.9,0.95)$ / 0.1 / 1.0 \\
    Weight initialization & std. 0.006, truncated at $\pm3\sigma$ \\
    Gates $g$ & zero initialization, no weight decay \\
    MoE balancing & per-pass bias; sequence auxiliary loss $10^{-4}$ \\
    Precision / GPUs & bfloat16 / $512\times$ H800 \\
    \bottomrule
  \end{tabular*}
\end{table*}

\textbf{Context-extension settings.} Context extension starts from the annealing weights but resets the optimizer and step counter.
With the budget-sampling seed unchanged, resetting the counter also restarts the budget sequence used at the beginning of pretraining.
It retains RoPE~\citep{su2023roformer} base 500,000 and uses context parallelism of degree 2, one sequence per micro-batch and 256 sequences per global batch.
The run lasts 5,960 updates under the schedule in Table~\ref{tab:pretraining-stages}.
Appendix Table~\ref{tab:cost} separates calculated compute and KV costs from measured latency, throughput and peak memory at every inference budget.

\subsection{Evaluation}
\label{sec:scaling}
\label{sec:eval-protocol}
We evaluate all five budgets on the same 37 tasks, using identical configurations and examples.
The primary aggregate groups all tasks into five categories: knowledge and QA, common sense, reasoning, mathematics and code.
For task score $s_t$, task set $\mathcal{T}_c$ in category $c$, and category set $\mathcal{C}$, we compute
\begin{equation}
  \bar{s}_c=\frac{1}{|\mathcal{T}_c|}\sum_{t\in\mathcal{T}_c}s_t,
  \qquad
  S=\frac{1}{|\mathcal{C}|}\sum_{c\in\mathcal{C}}\bar{s}_c.
  \label{eq:benchmark-aggregation}
\end{equation}
Thus each category receives equal weight, regardless of its number of tasks.
The summary in Figure~\ref{fig:public-model-radar} applies this definition to the displayed scores of all 37 tasks in Table~\ref{tab:base-benchmarks}.
Training trajectories and paired uncertainty estimates use the original 18-task core suite; Appendix Table~\ref{tab:aggregation-scopes} defines its task membership and the auxiliary 34-task scope.

\begin{table*}[!htbp]
\centering
\caption{\textbf{Base-model comparison across all 37 benchmarks.} All models follow the protocol in Appendix~\ref{app:base-eval-protocol}. Accordion-Base uses the 3.6T-token, 8K checkpoint. Category and overall scores appear in Figure~\ref{fig:public-model-radar}.}
\label{tab:base-benchmarks}
\accTableFont{10.5}{12}
\accTableStretch{1.0}
\renewcommand{\arraystretch}{1.02}
\setlength{\tabcolsep}{2pt}
\begin{tabular}{@{}>{\centering\arraybackslash}m{108pt}*{5}{>{\centering\arraybackslash}m{\dimexpr(\textwidth-108pt-18\tabcolsep-144pt)/5\relax}}*{4}{>{\centering\arraybackslash}m{36pt}}@{}}
\toprule
 & \multicolumn{5}{c}{\cellcolor{accTableOurs}\color{black}\textbf{\accTableLogo Accordion-Base}} & & & & \\
\multirow{-2}{*}{Benchmark} & \cellcolor{accTableOurs}$k=0$ & \cellcolor{accTableOurs}$k=3$ & \cellcolor{accTableOurs}$k=6$ & \cellcolor{accTableOurs}$k=9$ & \cellcolor{accTableOurs}$k=12$ & \multirow{-2}{*}{\makecell[c]{Qwen3\\0.6B}} & \multirow{-2}{*}{\makecell[c]{Qwen3\\1.7B}} & \multirow{-2}{*}{\makecell[c]{OLMoE\\0125}} & \multirow{-2}{*}{\makecell[c]{Ouro\\1.4B}} \\
\midrule
Pretraining tokens & \multicolumn{5}{c}{3.6T} & 36T & 36T & 5.1T & 7.7T \\
\addlinespace[3pt]
\rowcolor[HTML]{F1ECE5}\multicolumn{10}{c}{\bfseries Knowledge and QA} \\
MMLU & 58.2 & 59.8 & 60.9 & 60.8 & 60.9 & 52.5 & \underline{62.7} & 55.2 & \textbf{67.5} \\
TriviaQA & 51.9 & \cellcolor{accRankKnowledgeSecond}\underline{52.4} & \cellcolor{accRankKnowledgeSecond}\underline{52.4} & 52.3 & \cellcolor{accRankKnowledgeSecond}\underline{52.4} & 20.0 & 36.3 & \textbf{62.7} & 43.8 \\
Natural Questions & 19.1 & 19.3 & \cellcolor{accRankKnowledgeSecond}\underline{19.6} & 19.5 & \cellcolor{accRankKnowledgeSecond}\underline{19.6} & 7.5 & 13.8 & \textbf{25.8} & 17.2 \\
SciQ & 95.5 & 95.4 & 95.5 & 95.7 & 95.6 & 93.6 & \textbf{96.1} & 93.9 & \underline{95.8} \\
MMLU (cloze)$^{\dagger}$ & 43.8 & 44.9 & 45.2 & \cellcolor{accRankKnowledgeSecond}\underline{45.3} & \cellcolor{accRankKnowledgeFirst}\textbf{45.4} & 36.6 & 43.5 & 43.7 & 45.1 \\
MMLU-Pro$^{\dagger}$ & 32.3 & 34.1 & 35.7 & 35.8 & 36.0 & 25.7 & \underline{37.5} & 18.3 & \textbf{48.9} \\
DROP & 4.5 & 4.6 & 4.7 & 4.8 & 4.8 & 4.6 & \underline{15.3} & \textbf{32.1} & 4.7 \\
SQuAD & 56.1 & 56.6 & \cellcolor{accRankKnowledgeSecond}\underline{57.4} & 57.2 & 57.2 & 56.5 & \textbf{59.3} & 42.9 & 57.1 \\
CoQA & 77.5 & 78.9 & 78.9 & 79.3 & \cellcolor{accRankKnowledgeSecond}\underline{79.4} & 73.9 & 79.1 & 71.8 & \textbf{81.5} \\
RACE & \cellcolor{accRankKnowledgeSecond}\underline{38.9} & 37.0 & 37.8 & 37.2 & 37.8 & 35.1 & \textbf{39.5} & 37.9 & 38.3 \\
LAMBADA & 63.9 & 64.7 & 65.5 & 65.8 & \cellcolor{accRankKnowledgeSecond}\underline{65.9} & 53.8 & 63.1 & \textbf{70.5} & 65.1 \\
\addlinespace[3pt]
\rowcolor[HTML]{EDF4F1}\multicolumn{10}{c}{\bfseries Common sense} \\
ARC-C & 58.7 & 58.5 & 59.5 & 59.9 & \cellcolor{accRankCommonFirst}\textbf{60.2} & 45.0 & 55.4 & 58.2 & \underline{60.0} \\
ARC-E & 83.5 & 83.9 & \cellcolor{accRankCommonSecond}\underline{84.0} & 83.9 & \cellcolor{accRankCommonFirst}\textbf{84.1} & 74.7 & 82.2 & 82.5 & \underline{84.0} \\
CommonsenseQA & 63.3 & 65.7 & 66.9 & 67.0 & 66.8 & 62.9 & \underline{75.6} & 65.6 & \textbf{77.0} \\
HellaSwag & 74.3 & 74.9 & \cellcolor{accRankCommonSecond}\underline{75.3} & \cellcolor{accRankCommonSecond}\underline{75.3} & \cellcolor{accRankCommonSecond}\underline{75.3} & 53.6 & 67.1 & \textbf{79.7} & 74.4 \\
WinoGrande & 68.3 & 69.1 & 69.1 & 69.5 & 69.4 & 59.8 & 66.1 & \textbf{71.9} & \underline{70.8} \\
OpenBookQA & 41.4 & 42.4 & 43.2 & \cellcolor{accRankCommonSecond}\underline{43.6} & \cellcolor{accRankCommonSecond}\underline{43.6} & 34.2 & 38.4 & \textbf{44.4} & 41.4 \\
BoolQ & 78.1 & 78.2 & \cellcolor{accRankCommonSecond}\underline{79.1} & \cellcolor{accRankCommonSecond}\underline{79.1} & \cellcolor{accRankCommonSecond}\underline{79.1} & 69.6 & 78.9 & 70.2 & \textbf{79.9} \\
PIQA & 77.8 & 78.0 & \cellcolor{accRankCommonSecond}\underline{78.7} & 78.6 & 78.3 & 69.9 & 75.8 & \textbf{79.7} & 78.6 \\
Social IQa & 44.7 & 45.1 & 45.9 & 45.8 & 45.9 & 43.3 & \textbf{48.5} & \underline{47.2} & 42.9 \\
\addlinespace[3pt]
\rowcolor[HTML]{F1EEF6}\multicolumn{10}{c}{\bfseries Reasoning} \\
BBH & 54.8 & 58.6 & 60.0 & \cellcolor{accRankReasoningSecond}\underline{61.0} & 60.8 & 40.8 & 53.3 & 32.6 & \textbf{70.8} \\
MuSR & 37.2 & 36.9 & 38.6 & 38.8 & \cellcolor{accRankReasoningSecond}\underline{40.2} & 34.7 & \textbf{44.3} & 35.5 & 39.0 \\
AGIEval-en & 25.5 & 26.7 & 27.6 & 27.8 & 27.9 & 26.2 & \underline{32.5} & 25.5 & \textbf{34.9} \\
\addlinespace[3pt]
\rowcolor[HTML]{F8F4E8}\multicolumn{10}{c}{\bfseries Mathematics} \\
GSM8K & 53.1 & 58.0 & 59.4 & 61.7 & 61.3 & 59.2 & \underline{71.3} & 51.1 & \textbf{78.6} \\
MATH (5,000) & 39.8 & 41.0 & \cellcolor{accRankMathSecond}\underline{42.2} & \cellcolor{accRankMathSecond}\underline{42.2} & \cellcolor{accRankMathFirst}\textbf{42.4} & 28.5 & 38.9 & 16.9 & 22.0 \\
MATH500$^{\dagger}$ & 38.2 & \cellcolor{accRankMathSecond}\underline{40.8} & \cellcolor{accRankMathFirst}\textbf{41.4} & 40.2 & 40.6 & 31.8 & 39.4 & 17.2 & 21.4 \\
GSM8K-Platinum & 57.6 & 62.5 & 65.8 & 66.6 & 65.6 & 62.5 & \underline{75.8} & 53.1 & \textbf{80.9} \\
GSM-Plus-mini & 37.7 & 38.3 & 41.1 & 41.5 & 41.5 & 33.2 & \underline{50.4} & 33.8 & \textbf{54.9} \\
MathQA & 46.7 & 47.4 & \cellcolor{accRankMathSecond}\underline{48.1} & 48.0 & 48.0 & 36.8 & 45.2 & 28.0 & \textbf{49.8} \\
Multi-digit arithmetic & 23.3 & 28.9 & 31.5 & 29.3 & 28.6 & 87.3 & \textbf{96.5} & 43.7 & \underline{92.2} \\
ASDiv & 9.8 & 11.2 & 11.7 & \cellcolor{accRankMathFirst}\textbf{12.7} & \cellcolor{accRankMathSecond}\underline{12.2} & 0.0 & 7.9 & 0.4 & 2.6 \\
\addlinespace[3pt]
\rowcolor[HTML]{F7EEEE}\multicolumn{10}{c}{\bfseries Code} \\
HumanEval & 48.8 & 50.0 & 51.2 & \cellcolor{accRankCodeFirst}\textbf{53.7} & \cellcolor{accRankCodeSecond}\underline{53.1} & 28.7 & 50.6 & 11.6 & 35.37 \\
HumanEval+ & 43.9 & 45.1 & 46.3 & \cellcolor{accRankCodeFirst}\textbf{48.8} & \cellcolor{accRankCodeSecond}\underline{47.0} & 22.6 & 43.9 & 11.0 & 32.32 \\
MBPP & 46.6 & 50.2 & 50.2 & 50.4 & 51.8 & 39.0 & \underline{55.6} & 21.0 & \textbf{58.0} \\
MBPP+ & 60.6 & 64.8 & 66.1 & 66.1 & 66.4 & 54.5 & \underline{70.6} & 29.9 & \textbf{76.7} \\
CRUXEval output & 23.1 & 24.8 & 25.1 & \cellcolor{accRankCodeSecond}\underline{25.4} & 25.3 & 16.1 & 21.8 & 13.1 & \textbf{27.5} \\
CRUXEval input & 34.9 & 38.6 & \cellcolor{accRankCodeSecond}\underline{41.3} & 40.9 & 40.6 & 29.1 & 34.3 & 19.4 & \textbf{41.8} \\
\bottomrule
\end{tabular}
\par\vspace{3pt}
\begin{minipage}{\textwidth}
\accTableFont{10.5}{12}
\textbf{Bold}/\underline{underlining}: row best/second-best; fills mark Accordion ranks (ties included). $^{\dagger}$ Diagnostics. MATH variants remain separate. Ouro HumanEval/HumanEval+: trailing prompt whitespace removed; corrected scores used throughout.
\end{minipage}
\end{table*}

We use the Language Model Evaluation Harness~\citep{gao2023harness} (version 0.4.13) with a Hugging Face backend adapted to layer looping.
Scoring uses strict matching for BBH, mathematical-equivalence checking for mathematics and greedy pass@1 for code, whose tests run in isolated CPU subprocesses.
Appendix Table~\ref{tab:eval-protocol} specifies prompting and scoring for each task.
Appendix~\ref{app:table-ranking} gives the ranking conventions and statistical interpretation.

A source-level overlap audit identifies potential benchmark contamination in Diverse-QA.
Reported scores retain the full test sets, including flagged questions; Appendix~\ref{app:pretraining-evidence} gives the audit criteria, counts and limits.

\subsubsection{General Capabilities and Model Comparison}
\label{sec:base-results}
\label{sec:public-base-comparison}
\begin{figure}[H]
  \centering
  \includegraphics[width=6in]{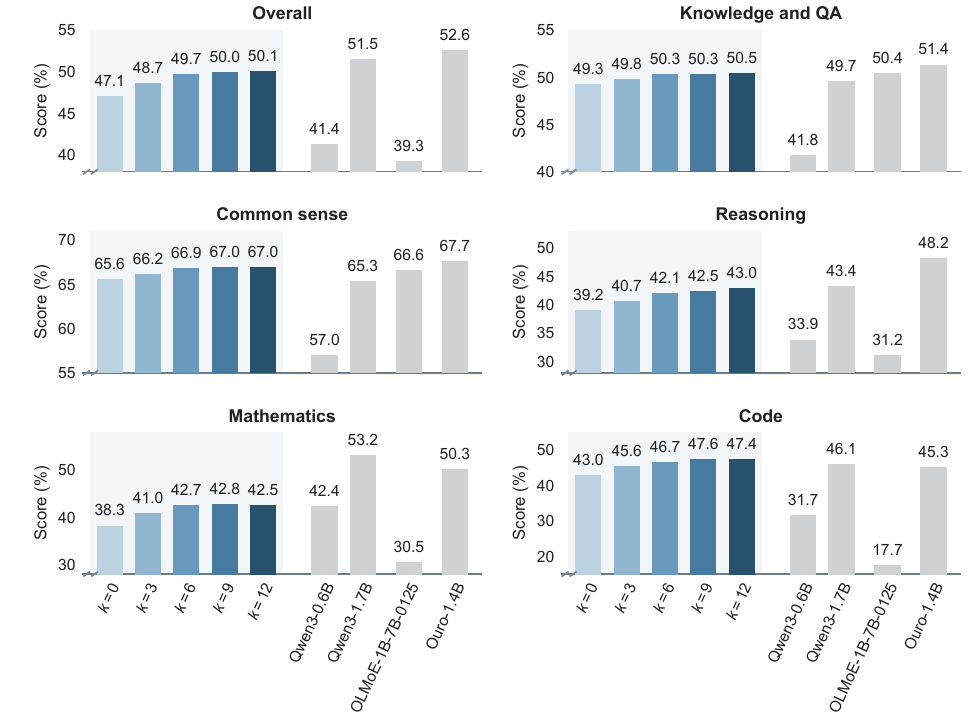}
  \caption{\textbf{Overall and category-level performance of Accordion-Base.} All five budgets and four references follow Appendix~\ref{app:base-eval-protocol}; Ouro's corrected HumanEval scores enter both Code and Overall. Category scores average all benchmarks in their corresponding table group; Overall equally weights the five category means, following Equation~\ref{eq:benchmark-aggregation}. Blue bars denote Accordion-Base at $k=0,3,6,9,12$; gray bars denote Qwen3-0.6B, Qwen3-1.7B, OLMoE-1B-7B-0125 and Ouro-1.4B, from left to right. Panels use separate y-axis ranges; break marks indicate omitted zero baselines. Table~\ref{tab:base-benchmarks} retains the complete benchmark results.}
  \label{fig:public-model-radar}
\end{figure}

Across all five inference budgets, Accordion-Base outperforms all four evaluated references on full MATH and ASDiv, with peak margins of 3.5 and 4.8 percentage points, respectively (Table~\ref{tab:base-benchmarks}).
Increasing the budget from $k=0$ to $k=12$ also improves all five category averages (Figure~\ref{fig:public-model-radar}).
The largest gains occur in code, reasoning and mathematics.
Accordion ranks second in knowledge and QA and common sense at $k=12$, and first in code at $k=9$ (Figure~\ref{fig:public-model-radar}).

The reference models cover complementary parameter scales and architectures. Qwen3-0.6B is a nearby dense reference in active size, whereas OLMoE-0125 is closer in total size and shares the MoE architecture.
We reevaluate Qwen3-0.6B/1.7B~\citep{qwen2025qwen3}, OLMoE-0125~\citep{muennighoff2024olmoe} and Ouro-1.4B~\citep{zhu2025ouro} on the complete suite. Accordion reaches these operating points after 3.6T pretraining tokens; Qwen3-1.7B uses 36T. Training exposure for all models is documented in Table~\ref{tab:public-model-specs}, with Accordion's computational costs in Appendix~\ref{app:cost}. Post-trained models are evaluated separately against four conversational references in Tables~\ref{tab:sft-public} and~\ref{tab:rl-public}.

\subsubsection{Effect of the Inference Budget}
\label{sec:pretrain-budget-effects}
\begin{figure}[!htbp]
\centering
\includegraphics[width=\textwidth]{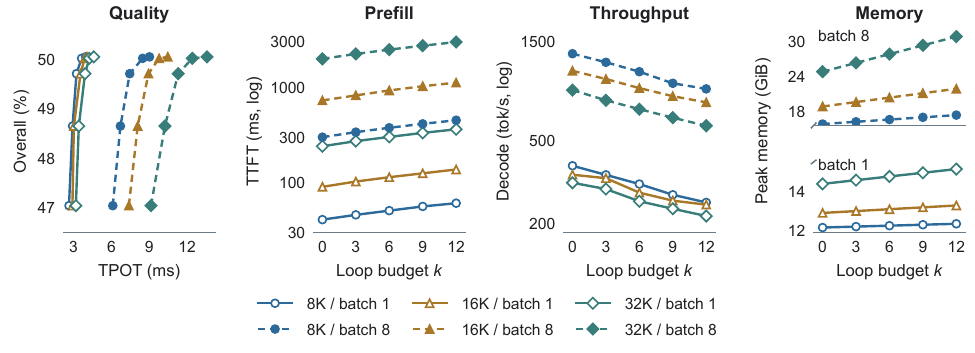}
\caption{\textbf{Quality and inference cost across six workloads.} The quality panel uses the 37-task five-category aggregate versus time per output token; other panels report time to first token, batch decoding throughput and peak allocated GPU memory. Runtime uses Accordion-RL on one H800, averaging two nodes' measurements; quality is evaluated separately. TTFT and throughput use logarithmic axes; memory uses separate batch-1 and batch-8 ranges. Measurement details appear in Table~\ref{tab:cost}.}
\label{fig:base-quality-latency}
\end{figure}

\begin{figure*}[!htbp]
\centering
\includegraphics[width=\textwidth]{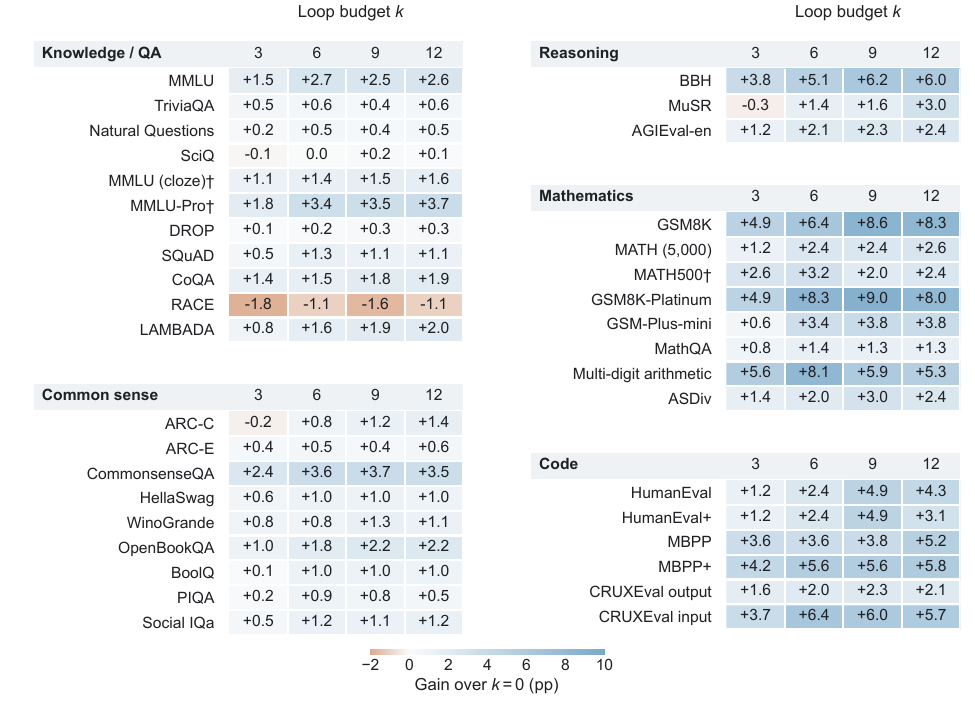}
\caption{\textbf{Benchmark gains across inference budgets.} Accordion-Base gains over $k=0$, in percentage points, for all 37 benchmarks. The baseline column is identically zero and omitted. All four non-baseline budgets share one linear color scale; $\dagger$ marks report-only tasks. Absolute scores appear in Table~\ref{tab:base-benchmarks}; category means appear in Figure~\ref{fig:public-model-radar}.}
\label{fig:base-budget-full}
\label{fig:main-results}
\end{figure*}

With model weights, evaluation tasks and examples fixed, increasing the loop budget from $k=0$ to $k=12$ raises the 37-task aggregate from 47.05 to 50.06 (+3.01 percentage points) and improves 36 of 37 task scores.
The gains are largest in reasoning, mathematics and code (Figure~\ref{fig:base-budget-full}).
The intermediate budget $k=6$ retains 89\% of the endpoint gain, while time per output token increases by approximately 21--24\% across measured workloads. This budget offers most of the quality improvement at a smaller inference cost (Figure~\ref{fig:base-quality-latency}); complete runtime measurements appear in Table~\ref{tab:cost}.

Paired uncertainty estimates for the original 18-task core suite are reported separately in Appendix Table~\ref{tab:pretrain-endpoint-uncertainty}.
The five budgets provide distinct operating points for different workloads (Figure~\ref{fig:base-budget-full}).
Both HumanEval variants peak at $k=9$, as does GSM8K, whereas MBPP+ reaches its highest score at $k=12$.

\phantomsection
\label{sec:domain-gains}
Additional depth also improves next-token prediction on held-out web text, mathematics and Python code. Loss decreases monotonically across all five budgets, with $k=6$ capturing 89--90\% of the endpoint reduction (Appendix Table~\ref{tab:base-validation}). These matched-token evaluations use the same Accordion-Base checkpoint as the benchmark comparison; Appendix~\ref{app:base-validation} gives the protocol.

\subsubsection{How Budget Gains Develop during Training}
\label{sec:analysis}
\label{sec:training-trajectory}
\begin{figure*}[!htbp]
\centering
\includegraphics[width=5.8in]{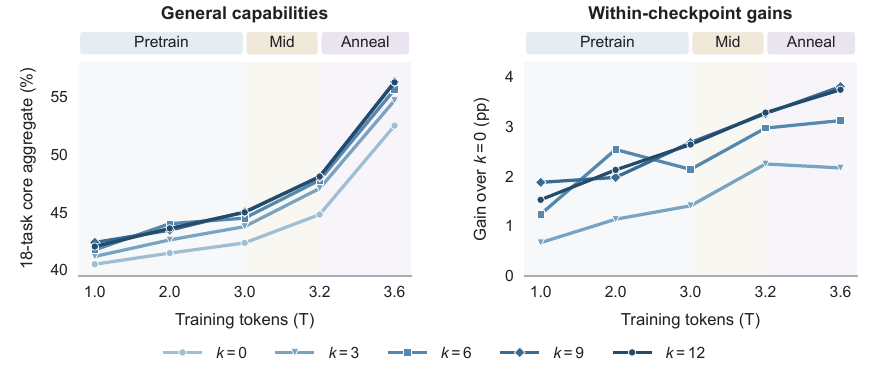}
\caption{\textbf{Performance and budget gains across training.} 18-task core aggregates through the 8K curriculum (left) and gains over each checkpoint's $k=0$ score (right). Checkpoints are equally spaced, with cumulative token counts labeled; horizontal distance does not represent token exposure. Gains use rounded scores.}
\label{fig:training-trajectory}
\end{figure*}

The advantage of looping emerges before post-training and persists across the pretraining curriculum.
All four looped budgets outperform $k=0$ at every evaluated checkpoint (Figure~\ref{fig:training-trajectory}). On this 18-task core aggregate, the endpoint advantage grows from 1.53 points early in training to 3.74 points at the end of annealing. Annealing improves all paths while preserving the benefit of additional inference computation.

\FloatBarrier
\subsubsection{Long-Context Performance}
\label{sec:context-extension}
\begin{figure}[!htbp]
  \centering
  \includegraphics[width=5.8in]{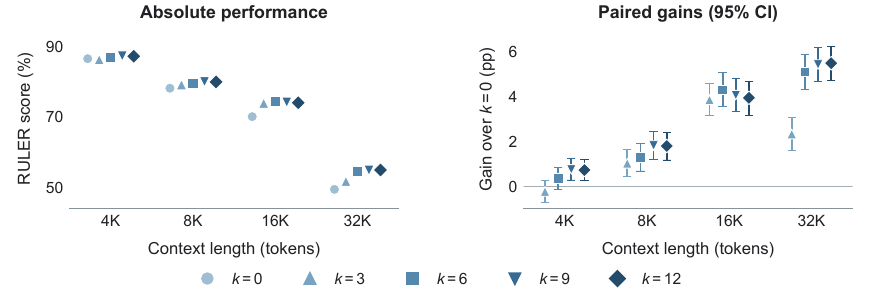}
  \caption{\textbf{Long-context performance of Accordion-Base-32K on RULER.} Scores average 13 tasks with 500 examples per task and context length. Gains over $k=0$ use task-stratified paired bootstrap 95\% confidence intervals (10,000 resamples). Full scores and endpoint intervals appear in Table~\ref{tab:long-context}.}
  \label{fig:long-context}
\end{figure}

On Accordion-Base-32K, additional depth yields larger gains on longer contexts. On the RULER~\citep{hsieh2024ruler} mean, increasing the budget from $k=0$ to $k=12$ adds 0.72 points at 4K and 5.47 points at 32K (Figure~\ref{fig:long-context}). At 32K, variable tracking and multi-key retrieval benefit most from the added computation (Appendix Table~\ref{tab:ruler-tasks-long}).
The five-budget interface adapts to context-dependent compute needs. At 16K, $k=3$ captures nearly all of the endpoint gain; at 32K, larger budgets unlock further gains through $k=9$. Appendix Tables~\ref{tab:long-context}--\ref{tab:ruler-tasks-long} provide complete scores, paired gains and task-level results. The evaluation protocol and consistency checks are described in Appendix~\ref{app:pretraining-evidence}.

The complete general-capability benchmark results for Accordion-Base-32K appear in Appendix Table~\ref{tab:base32k-benchmarks}.

\accEvidenceBarrier

\section{Post-Training}
\label{sec:post-training}
Post-training equips Accordion with conversational reasoning and stronger instruction following while preserving all five inference budgets.
Supervised fine-tuning (SFT) produces Accordion-Instruct; domain-specific reinforcement learning (RL) and expert consolidation produce Accordion-RL.

\subsection{Supervised Fine-Tuning}
\label{sec:posttraining}
Accordion-Instruct supports conversational reasoning at all five inference budgets. Under the primary sampled evaluation, increasing the budget from $k=0$ to $k=12$ improves nine of ten task scores, with the largest gains on competition mathematics and code generation. The same SFT weights support every budget and initialize all three RL experts.

\subsubsection{Data}
\begin{figure*}[!htbp]
\centering
\includegraphics[width=\textwidth]{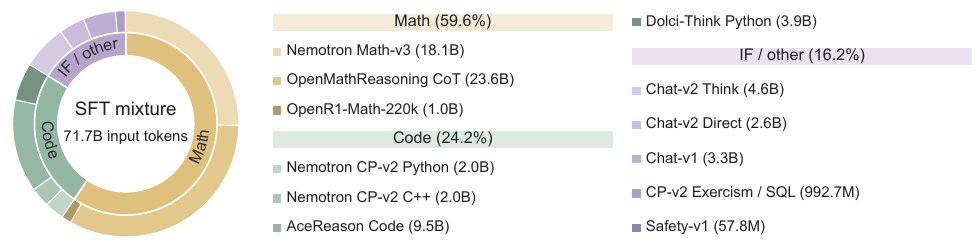}
\caption{\textbf{SFT data composition.} Category and source shares use packed input tokens after applying training epochs, totaling 71.734B; 68.5B carry training loss. Validation is excluded. B/M denote billion/million tokens. Exact source counts appear in Table~\ref{tab:sft-data}.}
\label{fig:sft-data-sources}
\end{figure*}

We fine-tune Accordion-Base-32K on publicly released datasets, with an emphasis on mathematics and code.
Training consumes 71.7B input tokens, of which 68.5B contribute to the masked training loss.
Mathematics and code examples receive one epoch; dialogue, instruction-following, Exercism/SQL and safety examples receive two.
Figure~\ref{fig:sft-data-sources} summarizes the mixture, and Table~\ref{tab:sft-data} gives source-level token counts and epoch allocations.
The sources include Nemotron-SFT-Math-v3~\citep{nvidia2026sftmath}, Nemotron-SFT-Competitive-Programming-v2~\citep{nvidia2025sftcode}, and Nemotron instruction-following/chat~\citep{nvidia2025sftchat,nvidia2025chatv1} and safety~\citep{nvidia2025sftsafety} collections.
We also use OpenR1-Math-220k~\citep{openr12025math}, chain-of-thought traces from OpenMathReasoning~\citep{moshkov2025openmathreasoning}, code examples from AceReason-1.1-SFT~\citep{liu2025acereason}, and correct solutions from Dolci-Think-SFT-Python~\citep{olmo2025olmo3}.

We exclude solutions requiring tool execution and sequences longer than 32K tokens, while retaining multiple solutions to the same problem.
Reasoning examples continue the format introduced by the Nemotron reasoning data during mid-training: \texttt{<think>} and \texttt{</think>} enclose the reasoning trace before the final answer.
Conversations without reasoning receive a \texttt{/no\_think} system instruction.
We pack conversations into 32K-token sequences without truncation; Appendix~\ref{app:sft-details} describes packing and token accounting.
To reduce benchmark overlap, we filter conversations against GSM8K, full MATH, HumanEval, MBPP and AIME 2024.
A conversation is removed when its user messages match at least 20\% of a test question's non-template 13-grams, removing approximately 19,000 conversations in total.
We reserve 0.5\% of examples for validation using a split by problem hash, so different solutions to the same problem cannot appear in both training and validation.

\subsubsection{Hyperparameters}
SFT starts from Accordion-Base-32K and trains all five paths using the uniform micro-batch budget sampling defined in \S\ref{sec:architecture}.
The global batch contains 256 packed 32K-token sequences.
Loss is averaged over assistant tokens, including the final end-of-sequence (EOS) token.
Training runs for 8,614 steps.
The learning rate rises to $2\times10^{-5}$ over 200 warmup steps, then follows a cosine schedule to $2\times10^{-6}$.
Appendix~\ref{app:sft-details} gives the batch and loss configuration.

\subsubsection{Evaluation}
\label{sec:sft-eval}
We evaluate Accordion-Instruct on ten benchmarks at all five inference budgets, using a common conversational protocol that also supports the matched RL comparison.
Eight tasks have public-model comparisons (Table~\ref{tab:sft-public}); the full suite additionally includes IFBench and LiveCodeBench~\citep{jain2024livecodebench}.
The primary evaluation uses vLLM with temperature $T=0.6$ and top-$p=0.95$.
We generate one response per problem, except for AIME~\citep{maaAime}, which averages 16 sampled responses per problem.
Appendix~\ref{app:sft-details} specifies the common response limits and a matched comparison with greedy decoding.
For responses containing a reasoning block, we score the text after the last closing \texttt{</think>} tag; a missing closing tag receives zero.
Responses without reasoning are scored in full.
IFEval~\citep{zhou2023ifeval} and IFBench~\citep{pyatkin2025ifbench} use strict prompt-level accuracy, MMLU-Redux~\citep{gema2024mmluredux} uses answer-letter accuracy, and code tasks use pass@1.
IFEval is scored serially, with language detection and Python randomness seeded to zero.

\begin{table*}[!htbp]
\centering
\caption{\textbf{Accordion-Instruct comparison with public post-trained models (\%).} Reference checkpoints (left to right): Qwen3-0.6B/1.7B, OLMoE-1B-7B-0125-SFT and Ouro-1.4B-Thinking. Bold/underlining identify best/second-best among the nine settings in each row, including ties; category-colored fills highlight Accordion's first/second places. All models use the common sampling and scoring protocol with native chat templates; OLMoE uses 4K context and the others 32K. AIME averages 16 responses per problem; code uses pass@1.}
\label{tab:sft-public}
\accTableFont{10.5}{13}
\accTableStretch{1.12}
\setlength{\tabcolsep}{2pt}
\begin{tabular}{@{}>{\centering\arraybackslash}m{80pt}*{5}{>{\centering\arraybackslash}m{30pt}}*{4}{>{\centering\arraybackslash}m{\dimexpr(\textwidth-230pt-18\tabcolsep)/4\relax}}@{}}
\toprule
 & \multicolumn{5}{c}{\cellcolor{accTableOurs}\color{black}\textbf{\accTableLogo Accordion-Instruct}} & & & & \\
\multirow{-2}{*}{Benchmark} & \cellcolor{accTableOurs}$k=0$ & \cellcolor{accTableOurs}$k=3$ & \cellcolor{accTableOurs}$k=6$ & \cellcolor{accTableOurs}$k=9$ & \cellcolor{accTableOurs}$k=12$ & \multirow{-2}{*}{\makecell[c]{Qwen3\\0.6B}} & \multirow{-2}{*}{\makecell[c]{Qwen3\\1.7B}} & \multirow{-2}{*}{\makecell[c]{OLMoE\\SFT}} & \multirow{-2}{*}{\makecell[c]{Ouro\\1.4B}} \\
\midrule
\addlinespace[3pt]
\rowcolor[HTML]{F1ECE5}\multicolumn{10}{c}{\bfseries Knowledge and instruction following} \\
IFEval & 62.11 & 64.88 & \cellcolor{accRankKnowledgeSecond}\underline{65.99} & 63.40 & 60.07 & 62.85 & \textbf{71.35} & 55.64 & 48.24 \\
MMLU-Redux & 60.88 & 63.47 & 65.82 & 66.17 & 65.16 & 57.69 & \textbf{74.84} & 14.02 & \underline{70.28} \\
\addlinespace[3pt]
\rowcolor[HTML]{F8F4E8}\multicolumn{10}{c}{\bfseries Mathematics} \\
GSM8K & 83.40 & 85.67 & 88.78 & 87.72 & 88.63 & 79.45 & \underline{90.14} & 35.86 & \textbf{92.42} \\
MATH500 & 83.40 & 84.80 & 87.40 & 86.80 & \cellcolor{accRankMathSecond}\underline{89.20} & 72.80 & \underline{89.20} & 13.60 & \textbf{91.20} \\
AIME 2024 & 30.00 & 34.79 & 39.58 & 43.54 & 43.75 & 12.08 & \underline{44.58} & 0.21 & \textbf{52.08} \\
AIME 2025 & 26.88 & 28.96 & 32.92 & \cellcolor{accRankMathSecond}\underline{36.88} & 31.88 & 16.25 & 34.58 & 0.21 & \textbf{41.25} \\
\addlinespace[3pt]
\rowcolor[HTML]{F7EEEE}\multicolumn{10}{c}{\bfseries Code} \\
HumanEval+ & 79.88 & 78.05 & 79.27 & 79.88 & \cellcolor{accRankCodeSecond}\underline{81.71} & 46.34 & 79.27 & 33.54 & \textbf{85.98} \\
MBPP+ & 69.58 & 67.72 & \cellcolor{accRankCodeSecond}\underline{73.02} & 70.90 & 72.75 & 43.65 & 70.90 & 30.42 & \textbf{75.40} \\
\bottomrule
\end{tabular}
\end{table*}

Additional depth improves Accordion-Instruct's mathematical reasoning and code generation, with the largest gains on competition mathematics and LiveCodeBench (Figure~\ref{fig:sft-formats}).
The complete sampled results appear alongside the matched RL results in Appendix Table~\ref{tab:rl-final}. Appendix Table~\ref{tab:sft-decoding-full} includes greedy decoding and paired uncertainty intervals for the change from $k=0$ to $k=12$.
Accordion-Instruct exceeds Qwen3-1.7B on HumanEval+ at $k=12$, MBPP+ at $k=6$ and AIME 2025 at $k=9$ (Table~\ref{tab:sft-public}). On HumanEval+, the margin is 2.44 percentage points. The AIME 2025 comparison also illustrates how a larger budget can change a model's relative standing while keeping its weights fixed.
All models use their native chat templates with the common sampling and scoring settings; OLMoE retains its native 4K context limit, while the others use 32K.
Appendix~\ref{app:sft-details} specifies the reference checkpoints, training stages and common evaluation protocol.

The five paths offer task-specific operating points: AIME 2024 reaches its highest sampled score at $k=12$, AIME 2025 at $k=9$ and MBPP+ at $k=6$ (Figure~\ref{fig:sft-formats}). Appendix~\ref{app:sft-details} compares greedy and sampled decoding and reports response-completion diagnostics.

\begin{figure}[H]
\centering
\includegraphics[width=\textwidth]{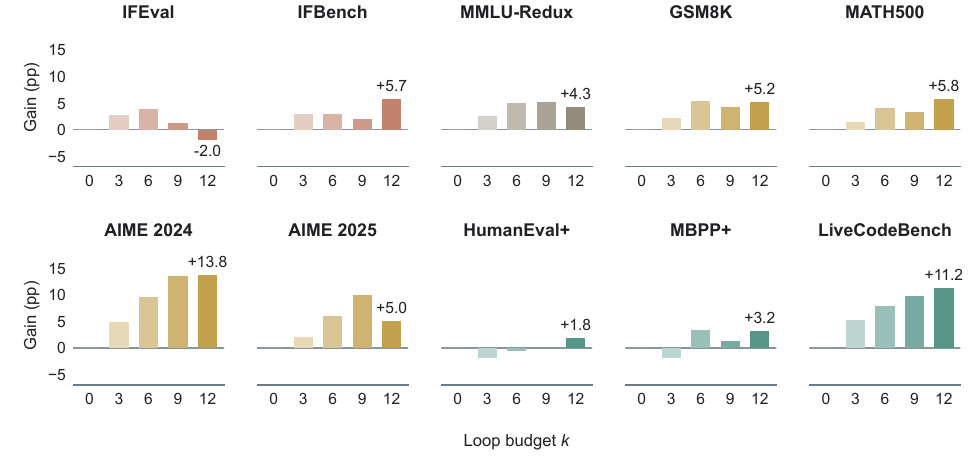}
\caption{\textbf{SFT gains over the unlooped budget.} Score differences from $k=0$ under the sampled conversational protocol. AIME averages 16 responses per problem; other tasks use one. Absolute scores and paired intervals appear in Table~\ref{tab:sft-decoding-full}.}
\label{fig:sft-formats}
\end{figure}

\accEvidenceBarrier

\subsection{Reinforcement Learning}
\label{sec:rl-results}
Accordion-RL improves instruction following at every inference budget and combines the capabilities of three domain experts. It exceeds all four evaluated public references on IFEval at every budget and exceeds each of its own teachers on the corresponding domain average. Training proceeds through domain-specific RL, model merging and budget-matched multi-teacher on-policy distillation (MOPD).

\subsubsection{Data}
\label{sec:rl-data}
We prepare separate instruction-following, mathematics and code pools for expert training, then sample equally from these pools for distillation.
All sources are publicly released problem collections (Appendix Table~\ref{tab:rl-data-sources}).
Instruction following (IF) combines Nemotron-Cascade-RL-Instruction-Following, Nemotron-RL-instruction\_following and its structured-output collection~\citep{nvidia2025cascadeif,nvidia2025rlif,nvidia2025rlstructured}.
Mathematics combines the OpenMathReasoning RL release, OpenR1-Math-220k~\citep{openr12025math}, AceReason-Math, the AoPS subset of Nemotron-Math-v2 and Nemotron-Cascade-RL-Math.
Code uses the competitive-coding and OpenCodeReasoning subsets of Nemotron-RL-coding-competitive\_coding.
We normalize and deduplicate prompts, remove evaluation overlap, filter by difficulty and verify code examples.
Table~\ref{tab:mopd-data} gives the resulting expert pools and the balanced distillation sample.
During MOPD, the student generates its own responses to the sampled prompts; the corresponding RL expert supplies token-level supervision.

\begin{table*}[!htbp]
\centering
\caption{\textbf{Expert training pools and balanced MOPD sampling.} Counts are prompts. MOPD samples equally from the three final expert pools; the student generates its own responses under domain- and budget-matched expert supervision.}
\label{tab:mopd-data}
\accTableFont{11}{13.5}
\accTableStretch{1.15}
\begin{tabular*}{\textwidth}{@{\hspace{12pt}\extracolsep{\fill}}cccc@{\hspace{12pt}}}
\toprule
Domain & Expert pool (prompts) & MOPD sample (prompts) & MOPD share \\
\midrule
Mathematics & 181,405 & 19,200 & $1/3$ \\
Code & 21,679 & 19,200 & $1/3$ \\
Instruction following & 80,291 & 19,200 & $1/3$ \\
\midrule
Total & 283,375 & 57,600 & $1$ \\
\bottomrule
\end{tabular*}
\end{table*}

\subsubsection{Hyperparameters}
\label{sec:rl-expert-training}
The RL pipeline first trains three domain experts, then merges their parameter updates to initialize one student, and finally distills their predictions into that student.
Every expert and the final student retain all five inference paths.

\paragraph{Expert training.}
The three branches start from Accordion-Instruct and optimize instruction following, mathematics and code with group relative policy optimization (GRPO) and DAPO-style training~\citep{shao2024deepseekmath,yu2025dapo}.
Each prompt group is assigned a budget from $\{0,3,6,9,12\}$ by hashing its identifier (MD5 modulo five), giving an approximately uniform allocation. All responses in the group use that budget. A rollout batch can contain all five budgets; groups are partitioned by budget and each nonempty partition receives a separate optimizer update, with the budget order shuffled each training step.
Expert-balancing biases remain frozen.
The IF branch uses an 8,192-token response limit and a soft overlength penalty; mathematics and code allow 28,672 response tokens within the 32K context window.

\begin{figure}[!htbp]
\centering
\includegraphics[width=\textwidth]{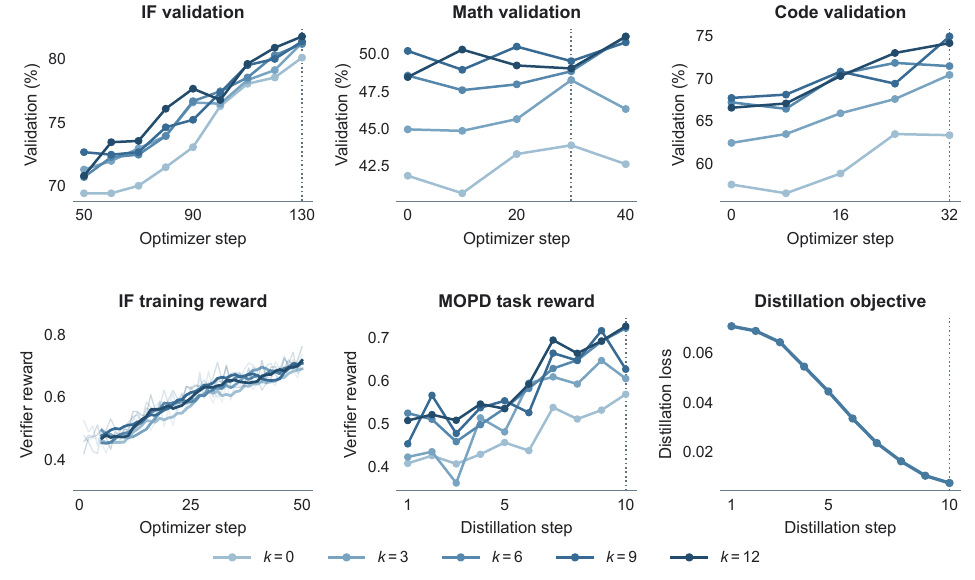}
\caption{\textbf{Expert training improves validation scores; distillation reduces teacher--student discrepancy.} Top row: expert validation with two responses per problem at $T=0.6$ (512 IF/math problems; 387 code problems). Bottom row: IF training reward, MOPD task reward and distillation loss. For IF training reward, solid curves are trailing five-step means over faint raw observations. Dotted lines mark the selected IF/math/code steps 130/30/32 and final MOPD step 10. Each curve represents one run.}
\label{fig:rl-learning}
\end{figure}

\paragraph{Model merging.}
\label{sec:rl-construction}
We use task arithmetic~\citep{ilharco2023editingmodelstaskarithmetic} to combine expert parameter changes relative to the shared SFT initialization, with coefficients that vary by parameter group:
\[
\theta^{\mathrm{merge}}_q=\theta_{\mathrm{SFT},q}+
\sum_j\lambda_{j,q}(\theta_{j,q}-\theta_{\mathrm{SFT},q}),
\]
where $j$ indexes the three training domains and $q$ indexes a parameter group.
For attention, shared MoE experts, loop gates and normalization, we use coefficient 1 for the IF expert and 0.5 for each of the code and mathematics experts.
The remaining trainable parameters use unit coefficients, while balancing biases stay fixed at their SFT values.
The resulting $\theta^{\mathrm{merge}}$ initializes the student; \S\ref{sec:rl-discussion} compares merge variants.

\paragraph{Budget-matched distillation.}
MOPD trains the merged student on its own generated responses, using a domain expert's token probabilities as supervision.
For each response, the teacher is chosen by prompt domain and evaluated at the same loop budget as the student.
Only the token-level distillation signal contributes to the objective; task reward is monitored during training.
Appendix~\ref{app:rl-mopd} gives the rollout configuration, training signal and clipping settings.

\paragraph{Training dynamics and model selection.}
\label{sec:rl-training-dynamics}
Expert selection uses held-out domain validation and entropy-based stopping; Appendix Table~\ref{tab:rl-stage-status} reports the selected steps and validation scores.
The final student, Accordion-RL, is selected after ten distillation steps using downstream comparisons with later steps.
Appendix~\ref{app:rl-mopd} specifies checkpoint selection and the corresponding uncertainty estimates.
All subsequent evaluations use the same Accordion-RL weights at every inference budget.
Expert training and distillation contribute complementary improvements (Figure~\ref{fig:rl-learning}): expert training raises held-out domain validation scores, while token-level distillation reduces teacher--student discrepancy and raises monitored task reward.
Appendix Table~\ref{tab:rl-training-dynamics} summarizes these changes.

\FloatBarrier
\subsubsection{Evaluation}
\label{sec:rl-benchmarks}
\label{sec:rl-eval}
To measure the change from SFT to RL at each inference budget, we evaluate the same ten benchmarks with the same examples, generation settings and scoring procedures used for the sampled SFT evaluation in \S\ref{sec:sft-eval}.
The selected Accordion-RL checkpoint uses this protocol at all five budgets.
Appendix~\ref{app:rl-final-paired} provides the complete matched scores and group-mean changes.

\label{sec:posttraining-comparison}
Accordion-RL exceeds the four evaluated public references on IFEval at every inference budget, with a margin of at least 16.64 percentage points over the strongest reference (Table~\ref{tab:rl-public}). It also achieves the highest HumanEval+ score among the evaluated settings at $k=6$ and narrowly exceeds Qwen3-1.7B on MATH500 at $k=12$.
Relative to matched SFT, instruction-following and knowledge gains persist across the entire budget range: all 15 task--budget comparisons on IFEval, IFBench and MMLU-Redux show higher scores after RL (Appendix Table~\ref{tab:rl-final}).

\begin{table*}[!htbp]
\centering
\caption{\textbf{RL comparison with public post-trained models (\%).} Reference checkpoints (left to right): Qwen3-0.6B/1.7B, OLMoE-1B-7B-0125-SFT and Ouro-1.4B-Thinking. Bold/underlining identify best/second-best among the nine settings in each row, including ties; category-colored fills highlight Accordion's first/second places. All models use the common sampling and scoring protocol with native chat templates; OLMoE uses 4K context and the others 32K. AIME averages 16 responses per problem; code uses pass@1.}
\label{tab:rl-public}
\accTableFont{10.5}{13}
\accTableStretch{1.12}
\renewcommand{\arraystretch}{1.03}
\setlength{\tabcolsep}{2pt}
\begin{tabular}{@{}>{\centering\arraybackslash}m{80pt}*{5}{>{\centering\arraybackslash}m{30pt}}*{4}{>{\centering\arraybackslash}m{\dimexpr(\textwidth-230pt-18\tabcolsep)/4\relax}}@{}}
\toprule
 & \multicolumn{5}{c}{\cellcolor{accTableOurs}\color{black}\textbf{\accTableLogo Accordion-RL}} & & & & \\
\multirow{-2}{*}{Benchmark} & \cellcolor{accTableOurs}$k=0$ & \cellcolor{accTableOurs}$k=3$ & \cellcolor{accTableOurs}$k=6$ & \cellcolor{accTableOurs}$k=9$ & \cellcolor{accTableOurs}$k=12$ & \multirow{-2}{*}{\makecell[c]{Qwen3\\0.6B}} & \multirow{-2}{*}{\makecell[c]{Qwen3\\1.7B}} & \multirow{-2}{*}{\makecell[c]{OLMoE\\SFT}} & \multirow{-2}{*}{\makecell[c]{Ouro\\1.4B}} \\
\midrule
\addlinespace[3pt]
\rowcolor[HTML]{F1ECE5}\multicolumn{10}{c}{\bfseries Knowledge and instruction following} \\
IFEval & 87.99 & \cellcolor{accRankKnowledgeSecond}\underline{88.91} & \cellcolor{accRankKnowledgeFirst}\textbf{89.83} & \cellcolor{accRankKnowledgeFirst}\textbf{89.83} & \cellcolor{accRankKnowledgeFirst}\textbf{89.83} & 62.85 & 71.35 & 55.64 & 48.24 \\
MMLU-Redux & 70.98 & 73.68 & 74.13 & \cellcolor{accRankKnowledgeFirst}\textbf{75.18} & 74.77 & 57.69 & \underline{74.84} & 14.02 & 70.28 \\
\addlinespace[3pt]
\rowcolor[HTML]{F8F4E8}\multicolumn{10}{c}{\bfseries Mathematics} \\
GSM8K & 84.15 & 87.34 & 88.63 & 89.54 & 89.16 & 79.45 & \underline{90.14} & 35.86 & \textbf{92.42} \\
MATH500 & 83.80 & 84.80 & 88.60 & 89.60 & \cellcolor{accRankMathSecond}\underline{89.80} & 72.80 & 89.20 & 13.60 & \textbf{91.20} \\
AIME 2024 & 34.58 & 39.79 & 40.83 & \cellcolor{accRankMathSecond}\underline{45.21} & \cellcolor{accRankMathSecond}\underline{45.21} & 12.08 & 44.58 & 0.21 & \textbf{52.08} \\
AIME 2025 & 27.92 & 32.71 & \cellcolor{accRankMathSecond}\underline{36.88} & 35.83 & 36.67 & 16.25 & 34.58 & 0.21 & \textbf{41.25} \\
\addlinespace[3pt]
\rowcolor[HTML]{F7EEEE}\multicolumn{10}{c}{\bfseries Code} \\
HumanEval+ & 79.88 & 78.05 & \cellcolor{accRankCodeFirst}\textbf{86.59} & 82.93 & 84.15 & 46.34 & 79.27 & 33.54 & \underline{85.98} \\
MBPP+ & 69.58 & 71.43 & 71.43 & 72.49 & \cellcolor{accRankCodeSecond}\underline{73.81} & 43.65 & 70.90 & 30.42 & \textbf{75.40} \\
\bottomrule
\end{tabular}
\end{table*}

\begin{figure*}[!htbp]
\centering
\includegraphics[width=\textwidth]{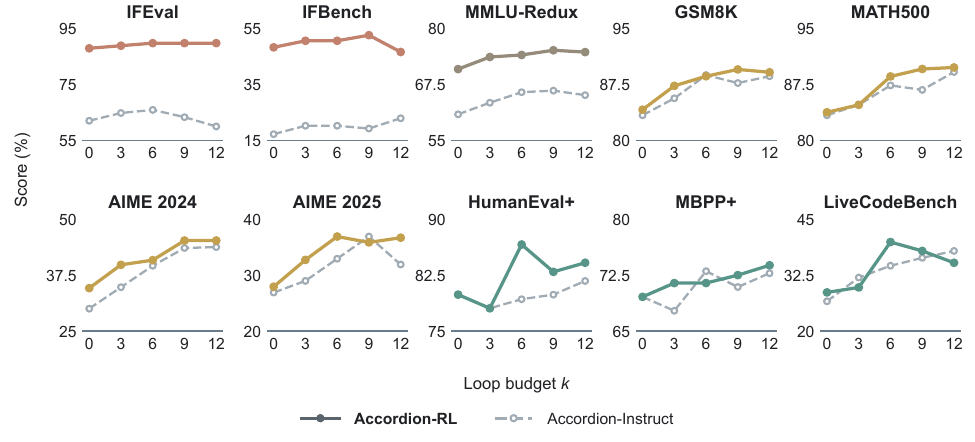}
\caption{\textbf{SFT and RL performance across inference budgets.} Accordion-Instruct and Accordion-RL use matched sampling and scoring on all ten tasks. Exact scores appear in Appendix Table~\ref{tab:rl-final}.}
\label{fig:rl-all-budget-profiles}
\end{figure*}

The largest average gains over SFT occur on IFBench and IFEval, followed by MMLU-Redux. Figure~\ref{fig:rl-all-budget-profiles} compares the matched SFT and RL scores across all five budgets; Appendix Table~\ref{tab:rl-final} gives the exact values.
Beyond instruction following, RL improves AIME 2025 at $k=12$ and both HumanEval+ and LiveCodeBench at $k=6$. LiveCodeBench also has substantially fewer truncated responses after RL. Appendix Figure~\ref{fig:rl-completion} distinguishes this change from accuracy among completed responses.

Increasing Accordion-RL's budget from $k=0$ to $k=12$ improves all four mathematics tasks and all three code tasks. Paired endpoint intervals also support gains on GSM8K, MATH500, MBPP+, LiveCodeBench and MMLU-Redux (Appendix Table~\ref{tab:rl-depth-gains}).
Budget selection exposes task-specific operating points: LiveCodeBench peaks at $k=6$, IFBench at $k=9$ and MATH500 at $k=12$ (Appendix Table~\ref{tab:rl-final}). A single model therefore supports several useful quality--cost choices.

\subsubsection{Discussion}
\label{sec:rl-discussion}
\begin{figure}[!htbp]
\centering
\includegraphics[width=5.6in]{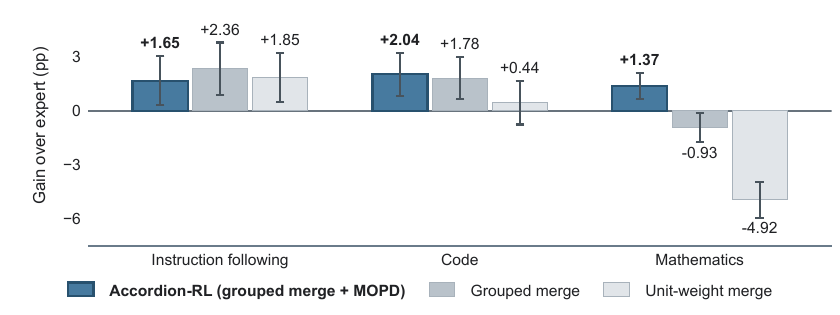}
\caption{\textbf{One student exceeds each domain expert.} Blue bars identify the final Accordion-RL model after grouped merging and MOPD. Gains average task differences relative to each domain expert, then average over budgets; error bars are paired bootstrap 95\% intervals. IF, code and mathematics contain two, three and four tasks; MMLU-Redux is excluded.}
\label{fig:rl-integration}
\end{figure}

Accordion-RL consolidates the three experts into one model and surpasses each teacher on its corresponding domain average, with paired 95\% intervals excluding zero (Figure~\ref{fig:rl-integration}; Appendix Table~\ref{tab:rl-integration-summary}). Each comparison averages task differences equally within each budget and then across all five budgets.

Model merging and distillation make complementary contributions to expert consolidation. Scaling the code and mathematics updates in shared components improves mathematics relative to unit-weight merging while retaining gains over the instruction-following and code experts. Budget-matched distillation adds a further 2.30 points to the mathematics average over this grouped initialization, yielding gains over all three domain experts. Appendix~\ref{app:rl-merging} compares the merge variants and reports the integration gains; Appendix~\ref{app:rl-mopd} specifies distillation and model selection.

\accEvidenceBarrier

\section{What Changes in the Second Pass?}
\label{sec:mechanism}
\label{sec:second-pass-behavior}

Accordion reuses weights while changing both the experts selected and the residual updates written into the hidden state. The effects of these updates also depend on which neighboring repetitions remain. Activation measurements and paired interventions show how shared parameters support distinct computations along the elastic-depth paths.

We probe Accordion-Base-32K at $k=12$ on 1,800 held-out documents, equally divided among web text, mathematics and code, with 1,024 targets per document. An intervention changes second-pass computation and recomputes all downstream activations. We report $\Delta\mathrm{NLL}=\mathrm{NLL}_{\mathrm{edit}}-\mathrm{NLL}_{k=12}$; a positive value means worse prediction on the same targets. Appendix~\ref{app:training-analysis} gives the controls, complete layerwise results and statistical protocol.

\subsection{Shared Weights Support Input-Dependent Expert Selection}
\label{sec:mechanism-selection}

\begin{figure}[!htbp]
\centering
\includegraphics[width=\textwidth]{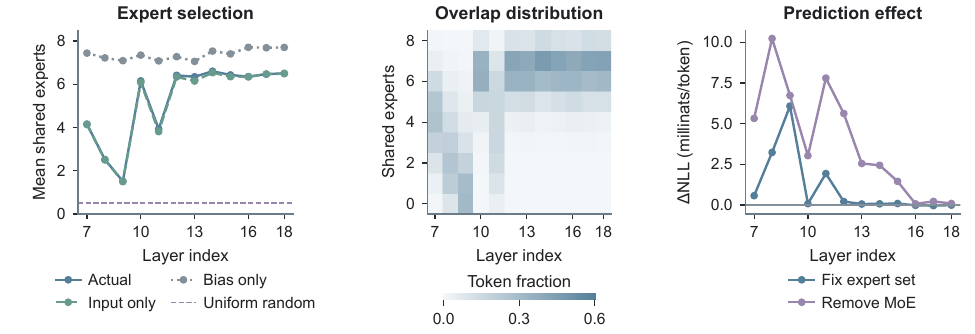}
\caption{\textbf{Input-driven expert switching contributes to prediction.} Left, mean overlap under actual routing, input-only and bias-only controls; independent uniform top-eight draws from 128 experts have expected overlap 0.5. Center, token-level overlap distributions for all twelve layers. Right, fixing first-pass expert sets versus removing second-pass MoE writes. Counts exclude the shared expert. Curves show point estimates; definitions and selected intervals are in Appendix~\ref{app:training-analysis}.}
\label{fig:mechanism-selection-main}
\end{figure}

The second pass selects substantially different experts at specific layers. Layer 9 retains only 1.52 of its eight routed experts on average, and most tokens share at most one expert between passes. Layers 7, 8 and 11 also switch extensively, whereas the remaining repeated layers largely reuse their first-pass selections (Figure~\ref{fig:mechanism-selection-main}).

The updated input drives this change. Changing the balancing bias alone largely preserves expert selection; changing the input with the first-pass bias reproduces actual overlap to within 0.20 experts. At layer 9, fixing the second-pass expert set to its first-pass selection incurs nearly the same NLL penalty as removing the entire second-pass MoE write.

\subsection{Repeated Layers Write Distinct Updates}
\label{sec:mechanism-writes}

\begin{figure}[!htbp]
\centering
\includegraphics[width=\textwidth]{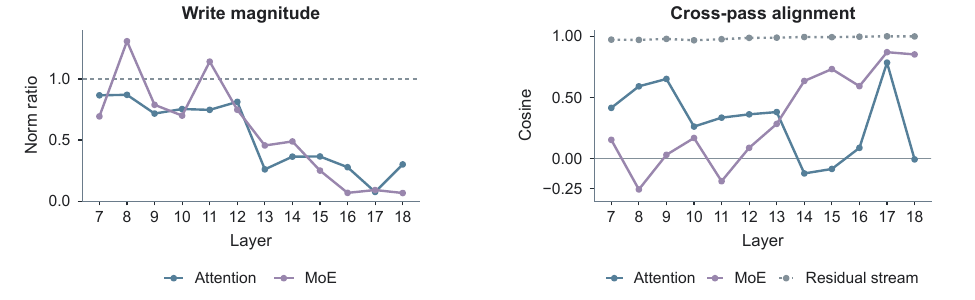}
\caption{\textbf{Shared parameters produce distinct residual updates.} Left, second-pass write norms relative to first-pass norms, with second-pass gates applied; the dashed line denotes equal magnitude. Right, cross-pass cosines of attention writes, MoE writes and residual streams. Residual streams remain closely aligned while attention and MoE writes change direction. Measurements average per-token quantities within each document and then across documents. Accordion-Base-32K, $k=12$.}
\label{fig:mechanism-writes-main}
\end{figure}

Repeated layers preserve the residual-stream direction while producing substantial, differently directed updates. In layers 7--12, gated second-pass MoE writes remain comparable in size to first-pass writes but are nearly orthogonal to them (Figure~\ref{fig:mechanism-writes-main}). Attention writes also change direction, while residual-stream cosines stay above 0.967 across the repeated window.

Even tokens routed to the same eight experts across passes receive distinct outputs, showing that expert reuse also supports input-dependent computation. Appendix~\ref{app:mechanism-routing} gives the conditional measurements and sampling definition.

Expert reselection and distinct second-pass writes persist through SFT and RL. Removing all second-pass writes increases NLL at each stage, measured relative to that stage's own baseline (Appendix~\ref{app:mechanism-stages}).

\subsection{The Effects of Repeated Layers Do Not Simply Add Up}
\label{sec:mechanism-interaction}

Distinct second-pass updates need not make independent contributions to prediction. Work on the Hydra effect shows that ablating one language-model component can change downstream computations, including compensatory responses~\citep{mcgrath2023hydra}. This motivates testing whether the effect of removing several repetitions can be predicted by adding their individual removal effects.

For each three-layer set, we compare three separate interventions, each starting from the intact $k=12$ model and removing one layer's second-pass attention and MoE writes, with one intervention removing all three together. All effects use the same intact baseline, and downstream activations are recomputed after each intervention. For layers 7--9, the three individual NLL increases sum to 24.65 millinats/token, whereas joint removal increases NLL by only 8.07. Across all four tested adjacent triplets, joint removal costs 15--33\% of the summed individual effects; for the four spaced triplets, it costs 96--103\% (Table~\ref{tab:mechanism-interaction-main}).

\begin{table}[H]
\centering
\caption{\textbf{Removing adjacent repetitions together costs less than the sum of separate removals.} All eight tested three-layer sets. ``Sum of singles'' adds NLL increases from three separate interventions; ``Joint effect'' measures one intervention removing all three layers' second-pass attention and MoE writes. Values are in millinats/token relative to the intact $k=12$ model. First-pass execution remains intact. A ratio of one means the two quantities agree. Complete results and uncertainty estimates are in Appendix~\ref{app:mechanism-interactions}.}
\label{tab:mechanism-interaction-main}
\accTableFont{9}{10}\accTableStretch{1}\setlength{\tabcolsep}{3pt}
\begin{tabular*}{\textwidth}{@{\extracolsep{\fill}}lcccccccc@{}}
\toprule
 & \multicolumn{4}{c}{Adjacent layers} & \multicolumn{4}{c}{Spaced layers} \\
\cmidrule(lr){2-5}\cmidrule(l){6-9}
Removed layers & 7--9 & 10--12 & 13--15 & 16--18 & 7, 11, 15 & 8, 12, 16 & 9, 13, 17 & 10, 14, 18 \\
\midrule
Sum of singles & 24.65 & 19.79 & 8.00 & 0.86 & 16.32 & 15.27 & 12.15 & 9.56 \\
Joint effect & 8.07 & 3.61 & 1.23 & 0.25 & 16.49 & 15.81 & 12.19 & 9.13 \\
\midrule
Joint / sum & 0.33 & 0.18 & 0.15 & 0.29 & 1.01 & 1.03 & 1.00 & 0.96 \\
\bottomrule
\end{tabular*}
\end{table}

The sum combines effects measured on different modified paths: each single-layer removal leaves the other repetitions active. Joint removal evaluates a different path, so its loss increase need not equal that sum. The observed discrepancy establishes that single-layer removal effects cannot be treated as independent contributions; it does not by itself identify a compensation mechanism. Removing all second-pass writes likewise costs 13.38 millinats/token, compared with a sum of 53.30 for individual removals. For Accordion, this means that the benefit of extra depth must be assessed on the complete budget path; adding single-layer ablation effects would overestimate the endpoint NLL gain. Appendix~\ref{app:mechanism-interactions} gives further group comparisons and interpretation details.
\phantomsection
\label{sec:mechanism-end}
\accEvidenceBarrier

\section{Related Work}
\label{sec:related}

\subsection{Looped Transformers}

Looped Transformers increase effective depth by repeatedly applying layers with shared weights \citep{dehghani2019universal,lan2020albert}. This family includes latent recurrent language models, which refine hidden states before predicting each token. The repeated unit can be a whole stack \citep{zhu2025ouro,nanbeige2026nanbeige42}, an internal block \citep{geiping2025huginn,wang2026smelt}, or an individual layer \citep{gao2026loopie}. Stack and block recurrence pass representations through multiple layers per iteration; layer looping reuses each layer locally before advancing. Accordion builds on local layer reuse, assigning each budget a fixed prefix of a middle window while retaining the unlooped path.

\subsection{Elastic Inference}

Elastic inference requires shared weights to support several execution paths, implemented through nested subnetworks \citep{devvrit2023matformer,cai2024flextron,taghibakhshi2025nemotronelastic,taghibakhshi2026starelastic}, reduced depth \citep{fan2020layerdrop,schuster2022calm,elhoushi2024layerskip}, or sampled recurrence depths \citep{geiping2025huginn}. Recurrence can also be introduced into pretrained models with frozen weights through damped updates \citep{chen2026trainingfree}.

For jointly trained recurrent depths, the central challenge is preserving shallow-path quality while benefiting from deeper computation. LoopFormer aligns trajectories through shortcut consistency \citep{jeddi2026loopformer}. Stochastic RINS improves unlooped inference through recurrence dropout and budget-specific adapters \citep{alabdulmohsin2025rins}, while Elastic Latent Recurrence (ELR) addresses the \emph{one-loop tax} through gated updates and within-update depth mixing, with loop conditioning supporting deeper budgets \citep{ladhak2026elr}. Accordion addresses this trade-off through path design: fixed-prefix paths share gates across budgets and are trained by uniformly sampling one path per pretraining micro-batch.
Budget selection is a separate choice: learned routing allocates depth per token \citep{raposo2024mod,bae2025mor,yu2026tloopformer,you2026tah2}. Accordion exposes depth as a caller-selected budget, using one fixed path for every token in a request.

\section{Conclusion}
\label{sec:conclusion}
Accordion makes inference depth a selectable property of a single model. Fixed-prefix layer looping supports five trained execution paths, and joint training preserves these choices through pretraining, SFT and RL. The base model obtains most of its additional-depth gain at the middle budget; its 32K extension retains depth gains on long-context tasks. Budget-matched multi-teacher distillation combines three domain experts into a student that exceeds each expert on its domain's task-and-budget average. The resulting model outperforms the four evaluated public references on IFEval at every budget and retains depth-dependent gains in mathematical reasoning and code generation. Mechanistic probes show that repeated layers reselect experts and write distinct updates, with prediction effects that depend on neighboring repetitions. Together, these results establish a training recipe based on public data for combining adjustable inference depth with domain expertise in one set of weights.

\begingroup
\renewcommand{\normalsize}{\fontsize{10.3}{13}\selectfont}
\normalsize
\bibliographystyle{plainnat}
\bibliography{references}

\begin{thebibliography}{93}
\providecommand{\natexlab}[1]{#1}
\providecommand{\url}[1]{\texttt{#1}}
\expandafter\ifx\csname urlstyle\endcsname\relax
  \providecommand{\doi}[1]{doi: #1}\else
  \providecommand{\doi}{doi: \begingroup \urlstyle{rm}\Url}\fi

\bibitem[Alabdulmohsin and Zhai(2025)]{alabdulmohsin2025rins}
Ibrahim Alabdulmohsin and Xiaohua Zhai.
\newblock {Recursive Inference Scaling: A Winning Path to Scalable Inference in
  Language and Multimodal Systems}.
\newblock \emph{arXiv preprint arXiv:2502.07503}, 2025.
\newblock URL \url{https://arxiv.org/abs/2502.07503}.

\bibitem[Amini et~al.(2019)Amini, Gabriel, Lin, Koncel-Kedziorski, Choi, and
  Hajishirzi]{amini2019mathqa}
Aida Amini, Saadia Gabriel, Peter Lin, Rik Koncel-Kedziorski, Yejin Choi, and
  Hannaneh Hajishirzi.
\newblock {MathQA: Towards Interpretable Math Word Problem Solving with
  Operation-Based Formalisms}.
\newblock arXiv preprint arXiv:1905.13319, 2019.
\newblock URL \url{https://arxiv.org/abs/1905.13319}.

\bibitem[Austin et~al.(2021)Austin, Odena, Nye, Bosma, Michalewski, Dohan,
  et~al.]{austin2021mbpp}
Jacob Austin, Augustus Odena, Maxwell Nye, Maarten Bosma, Henryk Michalewski,
  David Dohan, et~al.
\newblock {Program Synthesis with Large Language Models}.
\newblock arXiv preprint arXiv:2108.07732, 2021.
\newblock URL \url{https://arxiv.org/abs/2108.07732}.

\bibitem[Bae et~al.(2025)Bae, Kim, Bayat, Kim, Ha, Schuster, Fisch,
  Harutyunyan, Ji, Courville, et~al.]{bae2025mor}
Sangmin Bae, Yujin Kim, Reza Bayat, Sungnyun Kim, Jiyoun Ha, Tal Schuster, Adam
  Fisch, Hrayr Harutyunyan, Ziwei Ji, Aaron Courville, et~al.
\newblock {Mixture-of-Recursions: Learning Dynamic Recursive Depths for
  Adaptive Token-Level Computation}.
\newblock \emph{arXiv preprint arXiv:2507.10524}, 2025.
\newblock URL \url{https://arxiv.org/abs/2507.10524}.

\bibitem[Ben~Allal et~al.(2025)Ben~Allal, Lozhkov, Bakouch, Martín~Blázquez,
  Penedo, Tunstall, Marafioti, Kydlíček, Piqueres~Lajarín, Srivastav,
  et~al.]{allal2025smollm2}
Loubna Ben~Allal, Anton Lozhkov, Elie Bakouch, Gabriel Martín~Blázquez,
  Guilherme Penedo, Lewis Tunstall, Andrés Marafioti, Hynek Kydlíček,
  Agustín Piqueres~Lajarín, Vaibhav Srivastav, et~al.
\newblock {SmolLM2: When Smol Goes Big -- Data-Centric Training of a Small
  Language Model}.
\newblock \emph{arXiv preprint arXiv:2502.02737}, 2025.
\newblock URL \url{https://arxiv.org/abs/2502.02737}.

\bibitem[{BIG-bench authors}(2023)]{srivastava2023beyond}
{BIG-bench authors}.
\newblock {Beyond the Imitation Game: Quantifying and extrapolating the
  capabilities of language models}.
\newblock \emph{Transactions on Machine Learning Research}, 2023.
\newblock ISSN 2835-8856.
\newblock URL \url{https://openreview.net/forum?id=uyTL5Bvosj}.

\bibitem[Bisk et~al.(2019)Bisk, Zellers, Le~Bras, Gao, and Choi]{bisk2019piqa}
Yonatan Bisk, Rowan Zellers, Ronan Le~Bras, Jianfeng Gao, and Yejin Choi.
\newblock {PIQA: Reasoning about Physical Commonsense in Natural Language}.
\newblock arXiv preprint arXiv:1911.11641, 2019.
\newblock URL \url{https://arxiv.org/abs/1911.11641}.

\bibitem[Cai et~al.(2024)Cai, Muralidharan, Heinrich, Yin, Wang, Kautz, and
  Molchanov]{cai2024flextron}
Ruisi Cai, Saurav Muralidharan, Greg Heinrich, Hongxu Yin, Zhangyang Wang, Jan
  Kautz, and Pavlo Molchanov.
\newblock {Flextron: Many-in-One Flexible Large Language Model}.
\newblock \emph{arXiv preprint arXiv:2406.10260}, 2024.
\newblock URL \url{https://arxiv.org/abs/2406.10260}.

\bibitem[Chen et~al.(2026)Chen, Li, Liang, Lao, and Liu]{chen2026trainingfree}
Lizhang Chen, Jonathan Li, Chen Liang, Ni~Lao, and Qiang Liu.
\newblock {Training-Free Looped Transformers}.
\newblock \emph{arXiv preprint arXiv:2605.23872}, 2026.
\newblock URL \url{https://arxiv.org/abs/2605.23872}.

\bibitem[Chen et~al.(2021)Chen, Tworek, Jun, Yuan, de~Oliveira~Pinto, Kaplan,
  et~al.]{chen2021humaneval}
Mark Chen, Jerry Tworek, Heewoo Jun, Qiming Yuan, Henrique~Ponde
  de~Oliveira~Pinto, Jared Kaplan, et~al.
\newblock {Evaluating Large Language Models Trained on Code}.
\newblock arXiv preprint arXiv:2107.03374, 2021.
\newblock URL \url{https://arxiv.org/abs/2107.03374}.

\bibitem[Clark et~al.(2019)Clark, Lee, Chang, Kwiatkowski, Collins, and
  Toutanova]{clark2019boolq}
Christopher Clark, Kenton Lee, Ming-Wei Chang, Tom Kwiatkowski, Michael
  Collins, and Kristina Toutanova.
\newblock {BoolQ: Exploring the Surprising Difficulty of Natural Yes/No
  Questions}.
\newblock arXiv preprint arXiv:1905.10044, 2019.
\newblock URL \url{https://arxiv.org/abs/1905.10044}.

\bibitem[Clark et~al.(2018)Clark, Cowhey, Etzioni, Khot, Sabharwal, Schoenick,
  and Tafjord]{clark2018arc}
Peter Clark, Isaac Cowhey, Oren Etzioni, Tushar Khot, Ashish Sabharwal, Carissa
  Schoenick, and Oyvind Tafjord.
\newblock {Think you have Solved Question Answering? Try ARC, the AI2 Reasoning
  Challenge}.
\newblock arXiv preprint arXiv:1803.05457, 2018.
\newblock URL \url{https://arxiv.org/abs/1803.05457}.

\bibitem[Cobbe et~al.(2021)Cobbe, Kosaraju, Bavarian, Chen, Jun, Kaiser,
  et~al.]{cobbe2021gsm8k}
Karl Cobbe, Vineet Kosaraju, Mohammad Bavarian, Mark Chen, Heewoo Jun, Lukasz
  Kaiser, et~al.
\newblock {Training Verifiers to Solve Math Word Problems}.
\newblock arXiv preprint arXiv:2110.14168, 2021.
\newblock URL \url{https://arxiv.org/abs/2110.14168}.

\bibitem[{DeepSeek-AI}(2024)]{deepseek2024v3}
{DeepSeek-AI}.
\newblock {DeepSeek-V3 Technical Report}.
\newblock \emph{arXiv preprint arXiv:2412.19437}, 2024.
\newblock URL \url{https://arxiv.org/abs/2412.19437}.

\bibitem[Dehghani et~al.(2018)Dehghani, Gouws, Vinyals, Uszkoreit, and
  Kaiser]{dehghani2019universal}
Mostafa Dehghani, Stephan Gouws, Oriol Vinyals, Jakob Uszkoreit, and {\L}ukasz
  Kaiser.
\newblock {Universal Transformers}.
\newblock \emph{arXiv preprint arXiv:1807.03819}, 2018.
\newblock URL \url{https://arxiv.org/abs/1807.03819}.

\bibitem[Devvrit et~al.(2023)Devvrit, Kudugunta, Kusupati, Dettmers, Chen,
  Dhillon, Tsvetkov, Hajishirzi, Kakade, Farhadi, et~al.]{devvrit2023matformer}
Devvrit, Sneha Kudugunta, Aditya Kusupati, Tim Dettmers, Kaifeng Chen, Inderjit
  Dhillon, Yulia Tsvetkov, Hannaneh Hajishirzi, Sham Kakade, Ali Farhadi,
  et~al.
\newblock {MatFormer: Nested Transformer for Elastic Inference}.
\newblock \emph{arXiv preprint arXiv:2310.07707}, 2023.
\newblock URL \url{https://arxiv.org/abs/2310.07707}.

\bibitem[Dua et~al.(2019)Dua, Wang, Dasigi, Stanovsky, Singh, and
  Gardner]{dua2019drop}
Dheeru Dua, Yizhong Wang, Pradeep Dasigi, Gabriel Stanovsky, Sameer Singh, and
  Matt Gardner.
\newblock {DROP: A Reading Comprehension Benchmark Requiring Discrete Reasoning
  Over Paragraphs}.
\newblock arXiv preprint arXiv:1903.00161, 2019.
\newblock URL \url{https://arxiv.org/abs/1903.00161}.

\bibitem[Elhoushi et~al.(2024)Elhoushi, Shrivastava, Liskovich, Hosmer, Wasti,
  Lai, Mahmoud, Acun, Agarwal, Roman, et~al.]{elhoushi2024layerskip}
Mostafa Elhoushi, Akshat Shrivastava, Diana Liskovich, Basil Hosmer, Bram
  Wasti, Liangzhen Lai, Anas Mahmoud, Bilge Acun, Saurabh Agarwal, Ahmed Roman,
  et~al.
\newblock {LayerSkip: Enabling Early Exit Inference and Self-Speculative
  Decoding}.
\newblock \emph{arXiv preprint arXiv:2404.16710}, 2024.
\newblock URL \url{https://arxiv.org/abs/2404.16710}.

\bibitem[Fan et~al.(2019)Fan, Grave, and Joulin]{fan2020layerdrop}
Angela Fan, Edouard Grave, and Armand Joulin.
\newblock {Reducing Transformer Depth on Demand with Structured Dropout}.
\newblock \emph{arXiv preprint arXiv:1909.11556}, 2019.
\newblock URL \url{https://arxiv.org/abs/1909.11556}.

\bibitem[Gao et~al.(2023)Gao, Tow, Abbasi, Biderman, Black, DiPofi,
  et~al.]{gao2023harness}
Leo Gao, Jonathan Tow, Baber Abbasi, Stella Biderman, Sid Black, Anthony
  DiPofi, et~al.
\newblock {A Framework for Few-Shot Language Model Evaluation}, 2023.
\newblock URL \url{https://github.com/EleutherAI/lm-evaluation-harness}.

\bibitem[Gao et~al.(2026)Gao, Chen, Xiao, Yang, Tao, Zhou, and
  Dai]{gao2026loopie}
Zitian Gao, Yilong Chen, Yihao Xiao, Xinyu Yang, Ran Tao, Joey Zhou, and Bryan
  Dai.
\newblock {Loop the Loopies!}
\newblock \emph{arXiv preprint arXiv:2607.16051}, 2026.
\newblock URL \url{https://arxiv.org/abs/2607.16051}.

\bibitem[Geiping et~al.(2025)Geiping, McLeish, Jain, Kirchenbauer, Singh,
  Bartoldson, Kailkhura, Bhatele, and Goldstein]{geiping2025huginn}
Jonas Geiping, Sean McLeish, Neel Jain, John Kirchenbauer, Siddharth Singh,
  Brian~R. Bartoldson, Bhavya Kailkhura, Abhinav Bhatele, and Tom Goldstein.
\newblock {Scaling up Test-Time Compute with Latent Reasoning: A Recurrent
  Depth Approach}.
\newblock \emph{arXiv preprint arXiv:2502.05171}, 2025.
\newblock URL \url{https://arxiv.org/abs/2502.05171}.

\bibitem[Gema et~al.(2024)Gema, Leang, Hong, Devoto, Mancino, Saxena,
  et~al.]{gema2024mmluredux}
Aryo~Pradipta Gema, Joshua Ong~Jun Leang, Giwon Hong, Alessio Devoto, Alberto
  Carlo~Maria Mancino, Rohit Saxena, et~al.
\newblock {Are We Done with MMLU?}
\newblock arXiv preprint arXiv:2406.04127, 2024.
\newblock URL \url{https://arxiv.org/abs/2406.04127}.

\bibitem[Giannou et~al.(2023)Giannou, Rajput, yong Sohn, Lee, Lee, and
  Papailiopoulos]{giannou2023looped}
Angeliki Giannou, Shashank Rajput, Jy~yong Sohn, Kangwook Lee, Jason~D. Lee,
  and Dimitris Papailiopoulos.
\newblock {Looped Transformers as Programmable Computers}.
\newblock \emph{arXiv preprint arXiv:2301.13196}, 2023.
\newblock URL \url{https://arxiv.org/abs/2301.13196}.

\bibitem[Gu et~al.(2024)Gu, Rozi{\`e}re, Leather, Solar-Lezama, Synnaeve, and
  Wang]{gu2024cruxeval}
Alex Gu, Baptiste Rozi{\`e}re, Hugh Leather, Armando Solar-Lezama, Gabriel
  Synnaeve, and Sida~I. Wang.
\newblock {CRUXEval: A Benchmark for Code Reasoning, Understanding and
  Execution}.
\newblock arXiv preprint arXiv:2401.03065, 2024.
\newblock URL \url{https://arxiv.org/abs/2401.03065}.

\bibitem[Han et~al.(2024)Han, Jian, Hu, Liu, Wang, Fan, Ai, Huang, He, Yang,
  et~al.]{han2024infimm}
Xiaotian Han, Yiren Jian, Xuefeng Hu, Haogeng Liu, Yiqi Wang, Qihang Fan, Yuang
  Ai, Huaibo Huang, Ran He, Zhenheng Yang, et~al.
\newblock {InfiMM-WebMath-40B: Advancing Multimodal Pre-Training for Enhanced
  Mathematical Reasoning}.
\newblock \emph{arXiv preprint arXiv:2409.12568}, 2024.
\newblock URL \url{https://arxiv.org/abs/2409.12568}.

\bibitem[Hendrycks et~al.(2021{\natexlab{a}})Hendrycks, Burns, Basart, Zou,
  Mazeika, Song, and Steinhardt]{hendrycks2021mmlu}
Dan Hendrycks, Collin Burns, Steven Basart, Andy Zou, Mantas Mazeika, Dawn
  Song, and Jacob Steinhardt.
\newblock {Measuring Massive Multitask Language Understanding}.
\newblock In \emph{International Conference on Learning Representations},
  2021{\natexlab{a}}.
\newblock URL \url{https://arxiv.org/abs/2009.03300}.

\bibitem[Hendrycks et~al.(2021{\natexlab{b}})Hendrycks, Burns, Kadavath, Arora,
  Basart, Tang, Song, and Steinhardt]{hendrycks2021math}
Dan Hendrycks, Collin Burns, Saurav Kadavath, Akul Arora, Steven Basart, Eric
  Tang, Dawn Song, and Jacob Steinhardt.
\newblock {Measuring Mathematical Problem Solving With the MATH Dataset}.
\newblock arXiv preprint arXiv:2103.03874, 2021{\natexlab{b}}.
\newblock URL \url{https://arxiv.org/abs/2103.03874}.

\bibitem[Hsieh et~al.(2024)Hsieh, Sun, Kriman, Acharya, Rekesh, Jia, Zhang, and
  Ginsburg]{hsieh2024ruler}
Cheng-Ping Hsieh, Simeng Sun, Samuel Kriman, Shantanu Acharya, Dima Rekesh, Fei
  Jia, Yang Zhang, and Boris Ginsburg.
\newblock {RULER: What's the Real Context Size of Your Long-Context Language
  Models?}
\newblock \emph{arXiv preprint arXiv:2404.06654}, 2024.
\newblock URL \url{https://arxiv.org/abs/2404.06654}.

\bibitem[Ilharco et~al.(2023)Ilharco, Ribeiro, Wortsman, Gururangan, Schmidt,
  Hajishirzi, and Farhadi]{ilharco2023editingmodelstaskarithmetic}
Gabriel Ilharco, Marco~Tulio Ribeiro, Mitchell Wortsman, Suchin Gururangan,
  Ludwig Schmidt, Hannaneh Hajishirzi, and Ali Farhadi.
\newblock Editing models with task arithmetic, 2023.
\newblock URL \url{https://arxiv.org/abs/2212.04089}.

\bibitem[Jain et~al.(2024)Jain, Han, Gu, Li, Yan, Zhang, Wang, Solar-Lezama,
  Sen, and Stoica]{jain2024livecodebench}
Naman Jain, King Han, Alex Gu, Wen-Ding Li, Fanjia Yan, Tianjun Zhang, Sida
  Wang, Armando Solar-Lezama, Koushik Sen, and Ion Stoica.
\newblock {LiveCodeBench: Holistic and Contamination Free Evaluation of Large
  Language Models for Code}.
\newblock \emph{arXiv preprint arXiv:2403.07974}, 2024.
\newblock URL \url{https://arxiv.org/abs/2403.07974}.

\bibitem[Jeddi et~al.(2026)Jeddi, Ciccone, and Taati]{jeddi2026loopformer}
Ahmadreza Jeddi, Marco Ciccone, and Babak Taati.
\newblock {LoopFormer: Elastic-Depth Looped Transformers for Latent Reasoning
  via Shortcut Modulation}.
\newblock \emph{arXiv preprint arXiv:2602.11451}, 2026.
\newblock URL \url{https://arxiv.org/abs/2602.11451}.

\bibitem[Joshi et~al.(2017)Joshi, Choi, Weld, and
  Zettlemoyer]{joshi2017triviaqa}
Mandar Joshi, Eunsol Choi, Daniel~S. Weld, and Luke Zettlemoyer.
\newblock {TriviaQA: A Large Scale Distantly Supervised Challenge Dataset for
  Reading Comprehension}.
\newblock arXiv preprint arXiv:1705.03551, 2017.
\newblock URL \url{https://arxiv.org/abs/1705.03551}.

\bibitem[Kwiatkowski et~al.(2019)Kwiatkowski, Palomaki, Redfield, Collins,
  Parikh, Alberti, et~al.]{kwiatkowski2019naturalquestions}
Tom Kwiatkowski, Jennimaria Palomaki, Olivia Redfield, Michael Collins, Ankur
  Parikh, Chris Alberti, et~al.
\newblock {Natural Questions: A Benchmark for Question Answering Research}.
\newblock \emph{Transactions of the Association for Computational Linguistics},
  7:\penalty0 452--466, 2019.
\newblock \doi{10.1162/tacl_a_00276}.
\newblock URL \url{https://aclanthology.org/Q19-1026/}.

\bibitem[Kwon et~al.(2023)Kwon, Li, Zhuang, Sheng, Zheng, Yu, Gonzalez, Zhang,
  and Stoica]{kwon2023vllm}
Woosuk Kwon, Zhuohan Li, Siyuan Zhuang, Ying Sheng, Lianmin Zheng, Cody~Hao Yu,
  Joseph~E. Gonzalez, Hao Zhang, and Ion Stoica.
\newblock {Efficient Memory Management for Large Language Model Serving with
  PagedAttention}.
\newblock arXiv preprint arXiv:2309.06180, 2023.
\newblock URL \url{https://arxiv.org/abs/2309.06180}.

\bibitem[Ladhak et~al.(2026)Ladhak, Acharya, Serano, Sun, Ginsburg, and
  Puvvada]{ladhak2026elr}
Faisal Ladhak, Shantanu Acharya, Santiago~Akle Serano, Simeng Sun, Boris
  Ginsburg, and Krishna~C Puvvada.
\newblock {One Model to Loop Them All: Elastic Latent Recurrence for Language
  Models}.
\newblock \emph{arXiv preprint arXiv:2610.12582}, 2026.
\newblock URL \url{https://arxiv.org/abs/2610.12582}.

\bibitem[Lai et~al.(2017)Lai, Xie, Liu, Yang, and Hovy]{lai2017race}
Guokun Lai, Qizhe Xie, Hanxiao Liu, Yiming Yang, and Eduard Hovy.
\newblock {RACE: Large-scale ReAding Comprehension Dataset From Examinations}.
\newblock arXiv preprint arXiv:1704.04683, 2017.
\newblock URL \url{https://arxiv.org/abs/1704.04683}.

\bibitem[Lan et~al.(2019)Lan, Chen, Goodman, Gimpel, Sharma, and
  Soricut]{lan2020albert}
Zhenzhong Lan, Mingda Chen, Sebastian Goodman, Kevin Gimpel, Piyush Sharma, and
  Radu Soricut.
\newblock {ALBERT: A Lite BERT for Self-supervised Learning of Language
  Representations}.
\newblock \emph{arXiv preprint arXiv:1909.11942}, 2019.
\newblock URL \url{https://arxiv.org/abs/1909.11942}.

\bibitem[Li et~al.(2024)Li, Cui, Zhao, Kong, and Bi]{li2024gsmplus}
Qintong Li, Leyang Cui, Xueliang Zhao, Lingpeng Kong, and Wei Bi.
\newblock {GSM-Plus: A Comprehensive Benchmark for Evaluating the Robustness of
  LLMs as Mathematical Problem Solvers}.
\newblock arXiv preprint arXiv:2402.19255, 2024.
\newblock URL \url{https://arxiv.org/abs/2402.19255}.

\bibitem[Li et~al.(2023)Li, Ben~Allal, Zi, Muennighoff, Kocetkov, Mou, Marone,
  Akiki, Li, Chim, et~al.]{li2023starcoder}
Raymond Li, Loubna Ben~Allal, Yangtian Zi, Niklas Muennighoff, Denis Kocetkov,
  Chenghao Mou, Marc Marone, Christopher Akiki, Jia Li, Jenny Chim, et~al.
\newblock {StarCoder: may the source be with you!}
\newblock \emph{arXiv preprint arXiv:2305.06161}, 2023.
\newblock URL \url{https://arxiv.org/abs/2305.06161}.

\bibitem[Lightman et~al.(2023)Lightman, Kosaraju, Burda, Edwards, Baker, Lee,
  Leike, Schulman, Sutskever, and Cobbe]{lightman2023math500}
Hunter Lightman, Vineet Kosaraju, Yura Burda, Harri Edwards, Bowen Baker, Teddy
  Lee, Jan Leike, John Schulman, Ilya Sutskever, and Karl Cobbe.
\newblock {Let's Verify Step by Step}.
\newblock arXiv preprint arXiv:2305.20050, 2023.
\newblock URL \url{https://arxiv.org/abs/2305.20050}.

\bibitem[Liu et~al.(2023)Liu, Xia, Wang, and Zhang]{liu2023evalplus}
Jiawei Liu, Chunqiu~Steven Xia, Yuyao Wang, and Lingming Zhang.
\newblock {Is Your Code Generated by ChatGPT Really Correct? Rigorous
  Evaluation of Large Language Models for Code Generation}.
\newblock arXiv preprint arXiv:2305.01210, 2023.
\newblock URL \url{https://arxiv.org/abs/2305.01210}.

\bibitem[Liu et~al.(2025)Liu, Yang, Chen, Lee, Shoeybi, Catanzaro, and
  Ping]{liu2025acereason}
Zihan Liu, Zhuolin Yang, Yang Chen, Chankyu Lee, Mohammad Shoeybi, Bryan
  Catanzaro, and Wei Ping.
\newblock {AceReason-Nemotron 1.1: Advancing Math and Code Reasoning through
  SFT and RL Synergy}.
\newblock arXiv preprint arXiv:2506.13284, 2025.
\newblock URL \url{https://arxiv.org/abs/2506.13284}.

\bibitem[Loshchilov and Hutter(2019)]{loshchilov2019adamw}
Ilya Loshchilov and Frank Hutter.
\newblock {Decoupled Weight Decay Regularization}.
\newblock In \emph{International Conference on Learning Representations}, 2019.
\newblock URL \url{https://arxiv.org/abs/1711.05101}.

\bibitem[{Mathematical Association of America}(n.d.)]{maaAime}
{Mathematical Association of America}.
\newblock {American Invitational Mathematics Examination (AIME)}.
\newblock American Mathematics Competitions, n.d.
\newblock URL \url{https://maa.org/student-programs/amc/}.
\newblock Accessed October 20, 2026.

\bibitem[McGrath et~al.(2023)McGrath, Rahtz, Kramar, Mikulik, and
  Legg]{mcgrath2023hydra}
Thomas McGrath, Matthew Rahtz, Janos Kramar, Vladimir Mikulik, and Shane Legg.
\newblock {The Hydra Effect: Emergent Self-repair in Language Model
  Computations}.
\newblock \emph{arXiv preprint arXiv:2307.15771}, 2023.
\newblock URL \url{https://arxiv.org/abs/2307.15771}.

\bibitem[Miao et~al.(2020)Miao, Liang, and Su]{miao2020asdiv}
Shen-yun Miao, Chao-Chun Liang, and Keh-Yih Su.
\newblock {A Diverse Corpus for Evaluating and Developing English Math Word
  Problem Solvers}.
\newblock In \emph{Proceedings of the 58th Annual Meeting of the Association
  for Computational Linguistics}, pages 975--984, 2020.
\newblock \doi{10.18653/v1/2020.acl-main.92}.
\newblock URL \url{https://aclanthology.org/2020.acl-main.92/}.

\bibitem[Mihaylov et~al.(2018)Mihaylov, Clark, Khot, and
  Sabharwal]{mihaylov2018openbookqa}
Todor Mihaylov, Peter Clark, Tushar Khot, and Ashish Sabharwal.
\newblock {Can a Suit of Armor Conduct Electricity? A New Dataset for Open Book
  Question Answering}.
\newblock arXiv preprint arXiv:1809.02789, 2018.
\newblock URL \url{https://arxiv.org/abs/1809.02789}.

\bibitem[Moshkov et~al.(2025)Moshkov, Hanley, Sorokin, Toshniwal, Henkel,
  Schifferer, Du, and Gitman]{moshkov2025openmathreasoning}
Ivan Moshkov, Darragh Hanley, Ivan Sorokin, Shubham Toshniwal, Christof Henkel,
  Benedikt Schifferer, Wei Du, and Igor Gitman.
\newblock {AIMO-2 Winning Solution: Building State-of-the-Art Mathematical
  Reasoning Models with OpenMathReasoning dataset}.
\newblock arXiv preprint arXiv:2504.16891, 2025.
\newblock URL \url{https://arxiv.org/abs/2504.16891}.

\bibitem[Muennighoff et~al.(2024)Muennighoff, Soldaini, Groeneveld, Lo,
  Morrison, Min, et~al.]{muennighoff2024olmoe}
Niklas Muennighoff, Luca Soldaini, Dirk Groeneveld, Kyle Lo, Jacob Morrison,
  Sewon Min, et~al.
\newblock {OLMoE}: Open mixture-of-experts language models.
\newblock arXiv preprint arXiv:2409.02060, 2024.
\newblock URL \url{https://arxiv.org/abs/2409.02060}.

\bibitem[{Nanbeige Lab} et~al.(2026){Nanbeige Lab}, Yang, Huang, Lan, Chen,
  Zhou, Song, Cao, Zhu, Niu, Wang, Huang, Liang, Le, Feng, Sun, Gu, Zhang, Luo,
  Song, Xing, Wen, Xu, Chen, and Li]{nanbeige2026nanbeige42}
{Nanbeige Lab}, Chen Yang, Chengrui Huang, Fufeng Lan, Hanhui Chen, Hao Zhou,
  Huatong Song, Jiaqi Cao, Jiaying Zhu, Jinlin Niu, Kai Wang, Lisheng Huang,
  Qiliang Liang, Ran Le, Ruixiang Feng, Shuang Sun, Tao Gu, Tao Zhang, Tianyu
  Luo, Yang Song, Yun Xing, Yuntao Wen, Ziyao Xu, Zongchao Chen, and Zongqiang
  Li.
\newblock {Nanbeige4.2-3B: Unlocking Agentic Capabilities in a Compact Model}.
\newblock arXiv preprint arXiv:2607.22083, 2026.
\newblock URL \url{https://arxiv.org/abs/2607.22083}.

\bibitem[{NVIDIA}(2025{\natexlab{a}})]{nvidia2025cascadeif}
{NVIDIA}.
\newblock {Nemotron-Cascade-RL-Instruction-Following}.
\newblock Hugging Face dataset, 2025{\natexlab{a}}.
\newblock URL
  \url{https://huggingface.co/datasets/nvidia/Nemotron-Cascade-RL-Instruction-Following}.

\bibitem[{NVIDIA}(2025{\natexlab{b}})]{nvidia2025chatv1}
{NVIDIA}.
\newblock {Nemotron-Instruction-Following-Chat-v1}.
\newblock Hugging Face dataset, 2025{\natexlab{b}}.
\newblock URL
  \url{https://huggingface.co/datasets/nvidia/Nemotron-Instruction-Following-Chat-v1}.

\bibitem[{NVIDIA}(2025{\natexlab{c}})]{nvidia2025rlif}
{NVIDIA}.
\newblock {Nemotron-RL-instruction\_following}.
\newblock Hugging Face dataset, 2025{\natexlab{c}}.
\newblock URL
  \url{https://huggingface.co/datasets/nvidia/Nemotron-RL-instruction_following}.

\bibitem[{NVIDIA}(2025{\natexlab{d}})]{nvidia2025rlstructured}
{NVIDIA}.
\newblock {Nemotron-RL-instruction\_following-structured\_outputs}.
\newblock Hugging Face dataset, 2025{\natexlab{d}}.
\newblock URL
  \url{https://huggingface.co/datasets/nvidia/Nemotron-RL-instruction_following-structured_outputs}.

\bibitem[{NVIDIA}(2025{\natexlab{e}})]{nvidia2025sftchat}
{NVIDIA}.
\newblock {Nemotron-SFT-Instruction-Following-Chat-v2}.
\newblock Hugging Face dataset, 2025{\natexlab{e}}.
\newblock URL
  \url{https://huggingface.co/datasets/nvidia/Nemotron-SFT-Instruction-Following-Chat-v2}.

\bibitem[{NVIDIA}(2025{\natexlab{f}})]{nvidia2025sftcode}
{NVIDIA}.
\newblock {Nemotron-SFT-Competitive-Programming-v2}.
\newblock Hugging Face dataset, 2025{\natexlab{f}}.
\newblock URL
  \url{https://huggingface.co/datasets/nvidia/Nemotron-SFT-Competitive-Programming-v2}.

\bibitem[{NVIDIA}(2025{\natexlab{g}})]{nvidia2025sftsafety}
{NVIDIA}.
\newblock {Nemotron-SFT-Safety-v1}.
\newblock Hugging Face dataset, 2025{\natexlab{g}}.
\newblock URL
  \url{https://huggingface.co/datasets/nvidia/Nemotron-SFT-Safety-v1}.

\bibitem[{NVIDIA}(2026)]{nvidia2026sftmath}
{NVIDIA}.
\newblock {Nemotron-SFT-Math-v3}.
\newblock Hugging Face dataset, 2026.
\newblock URL
  \url{https://huggingface.co/datasets/nvidia/Nemotron-SFT-Math-v3}.

\bibitem[{Open R1 Team}(2025)]{openr12025math}
{Open R1 Team}.
\newblock {OpenR1-Math-220k}.
\newblock Hugging Face dataset, 2025.
\newblock URL \url{https://huggingface.co/datasets/open-r1/OpenR1-Math-220k}.

\bibitem[Paperno et~al.(2016)Paperno, Kruszewski, Lazaridou, Pham, Bernardi,
  Pezzelle, Baroni, Boleda, and Fernández]{paperno2016lambada}
Denis Paperno, Germán Kruszewski, Angeliki Lazaridou, Quan~Ngoc Pham,
  Raffaella Bernardi, Sandro Pezzelle, Marco Baroni, Gemma Boleda, and Raquel
  Fernández.
\newblock {The LAMBADA dataset: Word prediction requiring a broad discourse
  context}.
\newblock arXiv preprint arXiv:1606.06031, 2016.
\newblock URL \url{https://arxiv.org/abs/1606.06031}.

\bibitem[Prairie et~al.(2026)Prairie, Novack, Berg-Kirkpatrick, and
  Fu]{prairie2026parcae}
Hayden Prairie, Zachary Novack, Taylor Berg-Kirkpatrick, and Daniel~Y. Fu.
\newblock {Parcae: Scaling Laws For Stable Looped Language Models}.
\newblock \emph{arXiv preprint arXiv:2604.12946}, 2026.
\newblock URL \url{https://arxiv.org/abs/2604.12946}.

\bibitem[Pyatkin et~al.(2025)Pyatkin, Malik, Graf, Ivison, Huang, Dasigi,
  Lambert, and Hajishirzi]{pyatkin2025ifbench}
Valentina Pyatkin, Saumya Malik, Victoria Graf, Hamish Ivison, Shengyi Huang,
  Pradeep Dasigi, Nathan Lambert, and Hannaneh Hajishirzi.
\newblock {Generalizing Verifiable Instruction Following}.
\newblock arXiv preprint arXiv:2507.02833, 2025.
\newblock URL \url{https://arxiv.org/abs/2507.02833}.

\bibitem[{Qwen Team}(2025)]{qwen2025qwen3}
{Qwen Team}.
\newblock {Qwen3} technical report.
\newblock arXiv preprint arXiv:2505.09388, 2025.
\newblock URL \url{https://arxiv.org/abs/2505.09388}.

\bibitem[Rajpurkar et~al.(2016)Rajpurkar, Zhang, Lopyrev, and
  Liang]{rajpurkar2016squad}
Pranav Rajpurkar, Jian Zhang, Konstantin Lopyrev, and Percy Liang.
\newblock {SQuAD: 100,000+ Questions for Machine Comprehension of Text}.
\newblock arXiv preprint arXiv:1606.05250, 2016.
\newblock URL \url{https://arxiv.org/abs/1606.05250}.

\bibitem[Raposo et~al.(2024)Raposo, Ritter, Richards, Lillicrap, Humphreys, and
  Santoro]{raposo2024mod}
David Raposo, Sam Ritter, Blake Richards, Timothy Lillicrap, Peter~Conway
  Humphreys, and Adam Santoro.
\newblock {Mixture-of-Depths: Dynamically allocating compute in
  transformer-based language models}.
\newblock \emph{arXiv preprint arXiv:2404.02258}, 2024.
\newblock URL \url{https://arxiv.org/abs/2404.02258}.

\bibitem[Reddy et~al.(2019)Reddy, Chen, and Manning]{reddy2019coqa}
Siva Reddy, Danqi Chen, and Christopher~D. Manning.
\newblock {CoQA: A Conversational Question Answering Challenge}.
\newblock \emph{Transactions of the Association for Computational Linguistics},
  7:\penalty0 249--266, 2019.
\newblock \doi{10.1162/tacl_a_00266}.
\newblock URL \url{https://aclanthology.org/Q19-1016/}.

\bibitem[Sakaguchi et~al.(2019)Sakaguchi, Le~Bras, Bhagavatula, and
  Choi]{sakaguchi2019winogrande}
Keisuke Sakaguchi, Ronan Le~Bras, Chandra Bhagavatula, and Yejin Choi.
\newblock {WinoGrande: An Adversarial Winograd Schema Challenge at Scale}.
\newblock arXiv preprint arXiv:1907.10641, 2019.
\newblock URL \url{https://arxiv.org/abs/1907.10641}.

\bibitem[Sap et~al.(2019)Sap, Rashkin, Chen, LeBras, and
  Choi]{sap2019socialiqa}
Maarten Sap, Hannah Rashkin, Derek Chen, Ronan LeBras, and Yejin Choi.
\newblock {SocialIQA: Commonsense Reasoning about Social Interactions}.
\newblock arXiv preprint arXiv:1904.09728, 2019.
\newblock URL \url{https://arxiv.org/abs/1904.09728}.

\bibitem[Schuster et~al.(2022)Schuster, Fisch, Gupta, Dehghani, Bahri, Tran,
  Tay, and Metzler]{schuster2022calm}
Tal Schuster, Adam Fisch, Jai Gupta, Mostafa Dehghani, Dara Bahri, Vinh~Q.
  Tran, Yi~Tay, and Donald Metzler.
\newblock {Confident Adaptive Language Modeling}.
\newblock \emph{arXiv preprint arXiv:2207.07061}, 2022.
\newblock URL \url{https://arxiv.org/abs/2207.07061}.

\bibitem[Shao et~al.(2024)Shao, Wang, Zhu, Xu, Song, Bi,
  et~al.]{shao2024deepseekmath}
Zhihong Shao, Peiyi Wang, Qihao Zhu, Runxin Xu, Junxiao Song, Xiao Bi, et~al.
\newblock {DeepSeekMath: Pushing the Limits of Mathematical Reasoning in Open
  Language Models}.
\newblock arXiv preprint arXiv:2402.03300, 2024.
\newblock URL \url{https://arxiv.org/abs/2402.03300}.

\bibitem[Sheng et~al.(2024)Sheng, Zhang, Ye, Wu, Zhang, Zhang, Peng, Lin, and
  Wu]{sheng2024hybridflow}
Guangming Sheng, Chi Zhang, Zilingfeng Ye, Xibin Wu, Wang Zhang, Ru~Zhang,
  Yanghua Peng, Haibin Lin, and Chuan Wu.
\newblock {HybridFlow: A Flexible and Efficient RLHF Framework}.
\newblock arXiv preprint arXiv:2409.19256, 2024.
\newblock URL \url{https://arxiv.org/abs/2409.19256}.

\bibitem[Shoeybi et~al.(2019)Shoeybi, Patwary, Puri, LeGresley, Casper, and
  Catanzaro]{shoeybi2019megatron}
Mohammad Shoeybi, Mostofa Patwary, Raul Puri, Patrick LeGresley, Jared Casper,
  and Bryan Catanzaro.
\newblock {Megatron-LM: Training Multi-Billion Parameter Language Models Using
  Model Parallelism}.
\newblock \emph{arXiv preprint arXiv:1909.08053}, 2019.
\newblock URL \url{https://arxiv.org/abs/1909.08053}.

\bibitem[Sprague et~al.(2023)Sprague, Ye, Bostrom, Chaudhuri, and
  Durrett]{sprague2023musr}
Zayne Sprague, Xi~Ye, Kaj Bostrom, Swarat Chaudhuri, and Greg Durrett.
\newblock {MuSR: Testing the Limits of Chain-of-thought with Multistep Soft
  Reasoning}.
\newblock arXiv preprint arXiv:2310.16049, 2023.
\newblock URL \url{https://arxiv.org/abs/2310.16049}.

\bibitem[Su et~al.(2024)Su, Kong, Lin, Jennings, Norick, Kliegl, Patwary,
  Shoeybi, and Catanzaro]{su2024nemotroncc}
Dan Su, Kezhi Kong, Ying Lin, Joseph Jennings, Brandon Norick, Markus Kliegl,
  Mostofa Patwary, Mohammad Shoeybi, and Bryan Catanzaro.
\newblock {Nemotron-CC: Transforming Common Crawl into a Refined Long-Horizon
  Pretraining Dataset}.
\newblock \emph{arXiv preprint arXiv:2412.02595}, 2024.
\newblock URL \url{https://arxiv.org/abs/2412.02595}.

\bibitem[Su et~al.(2021)Su, Lu, Pan, Murtadha, Wen, and Liu]{su2023roformer}
Jianlin Su, Yu~Lu, Shengfeng Pan, Ahmed Murtadha, Bo~Wen, and Yunfeng Liu.
\newblock {RoFormer: Enhanced Transformer with Rotary Position Embedding}.
\newblock arXiv preprint arXiv:2104.09864, 2021.
\newblock URL \url{https://arxiv.org/abs/2104.09864}.

\bibitem[Suzgun et~al.(2022)Suzgun, Scales, Schärli, Gehrmann, Tay, Chung,
  et~al.]{suzgun2022bbh}
Mirac Suzgun, Nathan Scales, Nathanael Schärli, Sebastian Gehrmann, Yi~Tay,
  Hyung~Won Chung, et~al.
\newblock {Challenging BIG-Bench Tasks and Whether Chain-of-Thought Can Solve
  Them}.
\newblock arXiv preprint arXiv:2210.09261, 2022.
\newblock URL \url{https://arxiv.org/abs/2210.09261}.

\bibitem[Taghibakhshi et~al.(2025)Taghibakhshi, Sreenivas, Muralidharan, Cai,
  Chochowski, Mahabaleshwarkar, Suhara, Olabiyi, Korzekwa, Patwary,
  et~al.]{taghibakhshi2025nemotronelastic}
Ali Taghibakhshi, Sharath~Turuvekere Sreenivas, Saurav Muralidharan, Ruisi Cai,
  Marcin Chochowski, Ameya~Sunil Mahabaleshwarkar, Yoshi Suhara, Oluwatobi
  Olabiyi, Daniel Korzekwa, Mostofa Patwary, et~al.
\newblock {Nemotron Elastic: Towards Efficient Many-in-One Reasoning LLMs}.
\newblock \emph{arXiv preprint arXiv:2511.16664}, 2025.
\newblock URL \url{https://arxiv.org/abs/2511.16664}.

\bibitem[Taghibakhshi et~al.(2026)Taghibakhshi, Cai, Muralidharan, Sreenivas,
  Vavre, Mahabaleshwarkar, Kartal, Liang, Chochowski, Chen,
  et~al.]{taghibakhshi2026starelastic}
Ali Taghibakhshi, Ruisi Cai, Saurav Muralidharan, Sharath~Turuvekere Sreenivas,
  Aditya Vavre, Ameya~Sunil Mahabaleshwarkar, Bilal Kartal, Sheldon Liang,
  Marcin Chochowski, Zijia Chen, et~al.
\newblock {Star Elastic: Many-in-One Reasoning LLMs with Efficient Budget
  Control}.
\newblock \emph{arXiv preprint arXiv:2605.07182}, 2026.
\newblock URL \url{https://arxiv.org/abs/2605.07182}.

\bibitem[Talmor et~al.(2019)Talmor, Herzig, Lourie, and
  Berant]{talmor2019commonsenseqa}
Alon Talmor, Jonathan Herzig, Nicholas Lourie, and Jonathan Berant.
\newblock {CommonsenseQA: A Question Answering Challenge Targeting Commonsense
  Knowledge}.
\newblock In \emph{Proceedings of the 2019 Conference of the North American
  Chapter of the Association for Computational Linguistics: Human Language
  Technologies, Volume 1 (Long and Short Papers)}, pages 4149--4158, 2019.
\newblock \doi{10.18653/v1/N19-1421}.
\newblock URL \url{https://aclanthology.org/N19-1421/}.

\bibitem[{Team Olmo} et~al.(2025){Team Olmo}, Ettinger, Bertsch, Kuehl, Graham,
  Heineman, et~al.]{olmo2025olmo3}
{Team Olmo}, Allyson Ettinger, Amanda Bertsch, Bailey Kuehl, David Graham,
  David Heineman, et~al.
\newblock {Olmo 3}.
\newblock arXiv preprint arXiv:2512.13961, 2025.
\newblock URL \url{https://arxiv.org/abs/2512.13961}.

\bibitem[Vendrow et~al.(2025)Vendrow, Vendrow, Beery, and
  Madry]{vendrow2025platinum}
Joshua Vendrow, Edward Vendrow, Sara Beery, and Aleksander Madry.
\newblock {Do Large Language Model Benchmarks Test Reliability?}
\newblock arXiv preprint arXiv:2502.03461, 2025.
\newblock URL \url{https://arxiv.org/abs/2502.03461}.

\bibitem[Wang et~al.(2026)Wang, Zhang, Luo, Wu, Liu, Liu, Huang, Yan, and
  Li]{wang2026smelt}
Shaowen Wang, Ge~Zhang, Kairong Luo, Yuhao Wu, Shaofan Liu, Jiaheng Liu, Wenhao
  Huang, Shen Yan, and Jian Li.
\newblock {SMELT: Scaling Laws for Compute-Matched MoE Looped Transformers}.
\newblock \emph{arXiv preprint arXiv:2609.01343}, 2026.
\newblock URL \url{https://arxiv.org/abs/2609.01343}.

\bibitem[Wang et~al.(2024)Wang, Ma, Zhang, Ni, Chandra, Guo,
  et~al.]{wang2024mmlupro}
Yubo Wang, Xueguang Ma, Ge~Zhang, Yuansheng Ni, Abhranil Chandra, Shiguang Guo,
  et~al.
\newblock {MMLU-Pro: A More Robust and Challenging Multi-Task Language
  Understanding Benchmark}.
\newblock arXiv preprint arXiv:2406.01574, 2024.
\newblock URL \url{https://arxiv.org/abs/2406.01574}.

\bibitem[Welbl et~al.(2017)Welbl, Liu, and Gardner]{welbl2017sciq}
Johannes Welbl, Nelson~F. Liu, and Matt Gardner.
\newblock {Crowdsourcing Multiple Choice Science Questions}.
\newblock arXiv preprint arXiv:1707.06209, 2017.
\newblock URL \url{https://arxiv.org/abs/1707.06209}.

\bibitem[You et~al.(2026)You, Fu, Feng, Lv, Ning, Ding, and Wang]{you2026tah2}
Yichen You, Tianyu Fu, Aosong Feng, Xingtai Lv, Xuefei Ning, Ning Ding, and
  Yu~Wang.
\newblock {Improving Test-Time Scaling with Adaptive Looped Transformers}.
\newblock arXiv preprint arXiv:2609.35748, 2026.
\newblock URL \url{https://arxiv.org/abs/2609.35748}.

\bibitem[Yu et~al.(2026)Yu, Zhang, Cui, and Zhao]{yu2026tloopformer}
Mingqian Yu, Wenpeng Zhang, Shaobo Cui, and Peilin Zhao.
\newblock {T-LoopFormer: Token-Level Elastic-Depth Looped Transformers with
  Dynamic Routing}.
\newblock \emph{arXiv preprint arXiv:2609.15160}, 2026.
\newblock URL \url{https://arxiv.org/abs/2609.15160}.

\bibitem[Yu et~al.(2025)Yu, Zhang, Zhu, Yuan, Zuo, Yue, et~al.]{yu2025dapo}
Qiying Yu, Zheng Zhang, Ruofei Zhu, Yufeng Yuan, Xiaochen Zuo, Yu~Yue, et~al.
\newblock {DAPO: An Open-Source LLM Reinforcement Learning System at Scale}.
\newblock arXiv preprint arXiv:2503.14476, 2025.
\newblock URL \url{https://arxiv.org/abs/2503.14476}.

\bibitem[Zellers et~al.(2019)Zellers, Holtzman, Bisk, Farhadi, and
  Choi]{zellers2019hellaswag}
Rowan Zellers, Ari Holtzman, Yonatan Bisk, Ali Farhadi, and Yejin Choi.
\newblock {HellaSwag: Can a Machine Really Finish Your Sentence?}
\newblock arXiv preprint arXiv:1905.07830, 2019.
\newblock URL \url{https://arxiv.org/abs/1905.07830}.

\bibitem[Zhong et~al.(2023)Zhong, Cui, Guo, Liang, Lu, Wang, Saied, Chen, and
  Duan]{zhong2023agieval}
Wanjun Zhong, Ruixiang Cui, Yiduo Guo, Yaobo Liang, Shuai Lu, Yanlin Wang, Amin
  Saied, Weizhu Chen, and Nan Duan.
\newblock {AGIEval: A Human-Centric Benchmark for Evaluating Foundation
  Models}.
\newblock arXiv preprint arXiv:2304.06364, 2023.
\newblock URL \url{https://arxiv.org/abs/2304.06364}.

\bibitem[Zhou et~al.(2025)Zhou, Wang, Ranjan, Cheng, Tang, He, Liu, and
  Xing]{zhou2025megamath}
Fan Zhou, Zengzhi Wang, Nikhil Ranjan, Zhoujun Cheng, Liping Tang, Guowei He,
  Zhengzhong Liu, and Eric~P. Xing.
\newblock {MegaMath: Pushing the Limits of Open Math Corpora}.
\newblock \emph{arXiv preprint arXiv:2504.02807}, 2025.
\newblock URL \url{https://arxiv.org/abs/2504.02807}.

\bibitem[Zhou et~al.(2023)Zhou, Lu, Mishra, Brahma, Basu, Luan, Zhou, and
  Hou]{zhou2023ifeval}
Jeffrey Zhou, Tianjian Lu, Swaroop Mishra, Siddhartha Brahma, Sujoy Basu,
  Yi~Luan, Denny Zhou, and Le~Hou.
\newblock {Instruction-Following Evaluation for Large Language Models}.
\newblock arXiv preprint arXiv:2311.07911, 2023.
\newblock URL \url{https://arxiv.org/abs/2311.07911}.

\bibitem[Zhu et~al.(2025)Zhu, Wang, Hua, Zhang, Li, Que, Wei, Wen, Yin, Xing,
  et~al.]{zhu2025ouro}
Rui-Jie Zhu, Zixuan Wang, Kai Hua, Tianyu Zhang, Ziniu Li, Haoran Que, Boyi
  Wei, Zixin Wen, Fan Yin, He~Xing, et~al.
\newblock {Scaling Latent Reasoning via Looped Language Models}.
\newblock \emph{arXiv preprint arXiv:2510.25741}, 2025.
\newblock URL \url{https://arxiv.org/abs/2510.25741}.

\end{thebibliography}
\par\endgroup

\par
\clearpage
\appendix
\accUseCompactAppendix
\Needspace*{6\baselineskip}
\setcounter{topnumber}{5}
\setcounter{totalnumber}{8}

\section{Comparison of Execution Structures}
\label{sec:loop-structures}

\begin{figure}[H]
  \centering
  \input{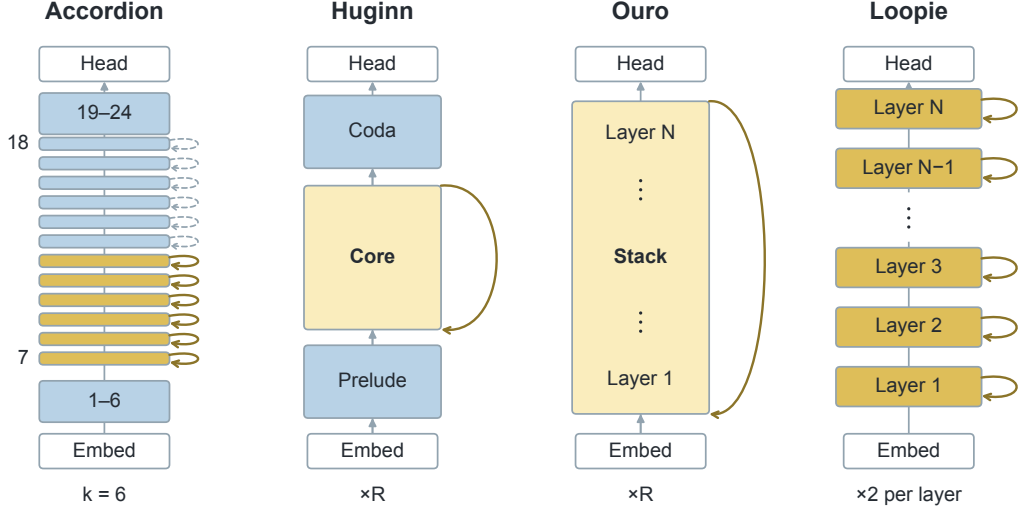}
  \caption{\textbf{Execution structures for depth reuse.} Execution proceeds bottom to top. Accordion repeats a middle-layer prefix ($k=6$ shown); Huginn repeats a core; Ouro repeats the full stack; Loopie repeats each MoE layer. $R$ denotes recurrence count. Blue denotes single execution and cream/gold repeated execution. Peer layer counts are schematic.}
  \label{fig:compare}
\end{figure}

Looped language models differ in the unit of repeated computation: Huginn repeats a shared core~\citep{geiping2025huginn}, Ouro repeats the full stack~\citep{zhu2025ouro}, and Loopie repeats individual layers~\citep{gao2026loopie} (Figure~\ref{fig:compare}).
\method{} adopts local layer looping and jointly trains a fixed set of prefix paths (Figure~\ref{fig:overview}).

\FloatBarrier
\section{Pretraining Data and Evaluation Details}
\label{app:pretraining-evidence}
Accordion-Base is the 8K model obtained after 3.6T training tokens. A further 50B tokens over 5,960 updates produce Accordion-Base-32K, which initializes SFT. Both models have complete results on the same 37 general-capability benchmarks at all five budgets. The main benchmark comparison reports Accordion-Base. Section~\ref{sec:context-extension} reports long-context results for Accordion-Base-32K; Appendix~\ref{app:base32k-results} gives its full benchmark table and detailed evaluation protocol.

\subsection{Data Accounting}

Available tokens in Table~\ref{tab:mixture} denote approximate source-pool sizes.
Expected consumption sums each stage's token budget weighted by its source sampling fraction; expected repetitions express this consumption in source-pool equivalents.
The source pools total approximately \pretrainPoolT{}T tokens under the rounded quantities listed here, while the selected 835 shards contain approximately 4.68T tokens.
The 3.6T training budget samples from the selected shards; stage-specific seeds 1234/1235/1236 change document order for shared sources.
Within each source category, shards are stratified by average document length.
We reserve one source shard each from Nemotron-CC v2 HQ, FineMath-3+ and StarCoder Python for validation, excluding these shards from training.
Context extension combines selected long documents with replay from the annealing distribution. Table~\ref{tab:context-mixture} specifies the target domain sampling proportions for this 50B-token stage.
\begin{table}[H]
  \centering
\accTableStretch{1.12}
  \caption{\textbf{Source-level mixture for the 3.6T/8K stages.} Available pools, expected consumption and repetitions use the documented rounded source quantities. General/mid/anneal columns each sum to 100\%; consumed tokens sum to 3,600B. The separate 50B context-extension mixture is excluded from consumption and repetition estimates.}
  \label{tab:mixture}
  \accTableFont{10.5}{13}
  \renewcommand{\arraystretch}{1.23}
  \setlength{\tabcolsep}{3pt}
  \begin{tabular*}{\textwidth}{@{\extracolsep{\fill}}l*{6}{c}@{}}
    \toprule
    Source & \makecell[c]{Available\\(B tokens)} & \makecell[c]{General\\3,000B} & \makecell[c]{Mid\\150B} & \makecell[c]{Anneal\\450B} & \makecell[c]{Expected\\used (B)} & \makecell[c]{Expected\\repetitions} \\
    \midrule
    Nemotron-CC web, high quality (v2/v2.1) & 1,243 & 23\% & 17\% & 13\% & 774 & 0.62 \\
    Nemotron-CC web, medium quality (v2/v2.1) & 2,161 & 15.25\% & 7\% & 3\% & 481.5 & 0.22 \\
    Nemotron-CC synthetic rephrasings & 3,503 & 22\% & 17\% & 12\% & 739.5 & 0.21 \\
    Nemotron-CC Diverse-QA + HQ-DQA & 679 & 6.5\% & 7\% & 5\% & 228 & 0.34 \\
    Nemotron Specialized (RQA, STEM-SFT, InfiniByte) & 228 & 2.5\% & 5\% & 8\% & 118.5 & 0.52 \\
    Nemotron reasoning (Math-v2, Comp.-Prog., PrismMath) & 102 & 0\% & 6\% & 12\% & 63 & 0.62 \\
    Nemotron-CC-Math & 195 & 3.5\% & 4.5\% & 5\% & 134.25 & 0.69 \\
    MegaMath web & 231 & 4.5\% & 5\% & 5\% & 165 & 0.71 \\
    MegaMath synthetic & 52 & 1\% & 0\% & 0\% & 30 & 0.58 \\
    FineMath-3+ and InfiWebMath-3+ & 53 & 2\% & 4\% & 5\% & 88.5 & 1.67 \\
    Math textbooks (Nemotron Specialized) & 24 & 0.5\% & 1\% & 2\% & 25.5 & 1.06 \\
    StarCoder data & 228 & 5.5\% & 10\% & 13\% & 238.5 & 1.05 \\
    Nemotron-CC-Code & 410 & 7\% & 6\% & 6\% & 246 & 0.60 \\
    Nemotron code synthetic (Code-v2) & 450 & 6.5\% & 10\% & 10\% & 255 & 0.57 \\
    Code concepts (Nemotron Specialized v1.1) & 9 & 0.25\% & 0.5\% & 1\% & 12.75 & 1.42 \\
    \bottomrule
  \end{tabular*}
\end{table}

\begin{table}[H]
\centering
\caption{\textbf{Context-extension data allocation.} Long-document and replay allocations each total 25B tokens. Domain shares combine both components; replay allocations follow the annealing mixture.}
\label{tab:context-mixture}
\accTableFont{10.5}{13}
\accTableStretch{1.12}
\setlength{\tabcolsep}{6pt}
\begin{tabular*}{\textwidth}{@{\hspace{12pt}\extracolsep{\fill}}lccccc@{\hspace{12pt}}}
\toprule
Metric & Web & Synthetic web & Reasoning & Mathematics & Code \\
\midrule
Long documents (B tokens) & 8 & 0 & 5 & 5 & 7 \\
Replay (B tokens) & 4 & 3 & 6.25 & 4.25 & 7.5 \\
Share (\%) & 24 & 6 & 22.5 & 18.5 & 29 \\
\bottomrule
\end{tabular*}
\end{table}

The audit cited in \S\ref{sec:eval-protocol} scans the 302B-token Diverse-QA source pool with the training tokenizer, excluding 13-grams shared by multiple questions within a benchmark. This pool is the Diverse-QA component of the combined 679B-token Diverse-QA + HQ-DQA row in Table~\ref{tab:mixture}. The reported high-overlap threshold requires coverage of at least 50\% of a question's non-template 13-grams. It flags 73 of 1,319 GSM8K questions and 3 of 500 MATH500 questions. These counts characterize overlap in the source pool; actual exposure depends on the sampled training stream. Reported benchmark scores include the flagged questions and are computed on the full evaluation sets.

\subsection{Evaluation Protocol and Aggregation}
\label{app:base-eval-protocol}
Accordion-Base and the public reference models are evaluated on the same 37 tasks, with Accordion retaining all five inference budgets (Table~\ref{tab:base-benchmarks}).
Table~\ref{tab:eval-protocol} gives the core prompting and scoring settings.
For Ouro-1.4B only, we remove trailing whitespace from HumanEval/HumanEval+ prompts before generation; all other settings are unchanged, and these corrected scores enter all benchmark tables and aggregates.
The extended suite includes SciQ~\citep{welbl2017sciq}, SocialIQA~\citep{sap2019socialiqa}, MuSR~\citep{sprague2023musr}, AGIEval~\citep{zhong2023agieval}, GSM8K-Platinum~\citep{vendrow2025platinum}, GSM-Plus~\citep{li2024gsmplus}, MathQA~\citep{amini2019mathqa}, ASDiv~\citep{miao2020asdiv} and both CRUXEval tasks~\citep{gu2024cruxeval}.
The reading tasks are DROP~\citep{dua2019drop}, SQuAD~\citep{rajpurkar2016squad}, CoQA~\citep{reddy2019coqa}, RACE~\citep{lai2017race} and LAMBADA~\citep{paperno2016lambada}; the report-only MMLU-Pro task follows \citet{wang2024mmlupro}.
The three summary scopes are specified in Table~\ref{tab:aggregation-scopes}.
ASDiv and arithmetic remain in the secondary suite; MATH500, cloze MMLU and MMLU-Pro are reported individually and enter only the full-table summary.

\begin{table}[H]
\centering
\caption{\textbf{Core evaluation protocol for Accordion-Base and public references.} All five budgets use identical tasks, prompts and scoring. Shot counts specify in-context examples; CoT denotes chain-of-thought prompting. Evaluation includes full MATH~\citep{hendrycks2021math} with 5,000 problems and a separately reported MATH500~\citep{lightman2023math500} score.}
\label{tab:eval-protocol}
\accTableFont{10.5}{13}
\accTableStretch{1.15}
\setlength{\tabcolsep}{5pt}
\begin{tabularx}{\textwidth}{@{\hspace{\tabcolsep}}cX@{}}
\toprule
\makebox[65pt][c]{Category} & \begin{tabular}{@{}m{\dimexpr\linewidth-146pt-2\tabcolsep\relax}>{\centering\arraybackslash}m{146pt}@{}}
{\centering Tasks and prompting\par} & Scoring
\end{tabular} \\
\midrule
\cellcolor{accCategoryKnowledge!15!white}\raisebox{-0.5\normalbaselineskip}[0pt][0pt]{Knowledge} &
\begin{tabular}{@{}m{\dimexpr\linewidth-146pt-2\tabcolsep\relax}>{\centering\arraybackslash}m{146pt}@{}}
\mbox{MMLU~\citep{hendrycks2021mmlu} (5-shot)} & \mbox{Multiple-choice accuracy} \\
\mbox{TriviaQA~\citep{joshi2017triviaqa}}\newline
\mbox{Natural Questions~\citep{kwiatkowski2019naturalquestions} (5-shot)} & \mbox{Closed-book exact match}
\end{tabular} \\
\midrule
\cellcolor{accCategoryCommon!15!white}\raisebox{-1.25\normalbaselineskip}[0pt][0pt]{Common sense} &
\begin{tabular}{@{}m{\dimexpr\linewidth-146pt-2\tabcolsep\relax}>{\centering\arraybackslash}m{146pt}@{}}
\mbox{ARC-C / ARC-E~\citep{clark2018arc} (25-/8-shot)}\newline
\mbox{CommonsenseQA~\citep{talmor2019commonsenseqa}}\newline
\mbox{HellaSwag~\citep{zellers2019hellaswag} (10-shot)}\newline
\mbox{WinoGrande~\citep{sakaguchi2019winogrande} (5-shot)}\newline
\mbox{OpenBookQA~\citep{mihaylov2018openbookqa}}\newline
\mbox{BoolQ~\citep{clark2019boolq}, PIQA~\citep{bisk2019piqa} (0-shot)} &
\mbox{Task-specific harness scoring}
\end{tabular} \\
\midrule
\cellcolor{accCategoryReasoning!15!white}\raisebox{-0.3\normalbaselineskip}[0pt][0pt]{Reasoning} &
\begin{tabular}{@{}m{\dimexpr\linewidth-146pt-2\tabcolsep\relax}>{\centering\arraybackslash}m{146pt}@{}}
\mbox{BBH~\citep{suzgun2022bbh} (3-shot CoT)} &
\makecell[c]{Strict answer matching\\aggregate over subtasks}
\end{tabular} \\
\midrule
\cellcolor{accCategoryMath!15!white}\raisebox{-0.5\normalbaselineskip}[0pt][0pt]{Mathematics} &
\begin{tabular}{@{}m{\dimexpr\linewidth-146pt-2\tabcolsep\relax}>{\centering\arraybackslash}m{146pt}@{}}
\mbox{GSM8K~\citep{cobbe2021gsm8k} (3-shot CoT)} & \mbox{Flexible answer extraction} \\
Full MATH (4-shot) & \mbox{Mathematical-equivalence checking}
\end{tabular} \\
\midrule
\cellcolor{accCategoryCode!15!white}\raisebox{-0.5\normalbaselineskip}[0pt][0pt]{Code} &
\begin{tabular}{@{}m{\dimexpr\linewidth-146pt-2\tabcolsep\relax}>{\centering\arraybackslash}m{146pt}@{}}
\mbox{HumanEval~\citep{chen2021humaneval} /}\newline
\mbox{HumanEval+~\citep{liu2023evalplus} (0-shot)}\newline
\mbox{MBPP~\citep{austin2021mbpp} / MBPP+ (3-shot)} &
\makecell[c]{Greedy pass@1\\CPU subprocess test execution}
\end{tabular} \\
\bottomrule
\end{tabularx}
\end{table}

\begin{table}[H]
\centering
\caption{\textbf{Task and category scopes of the pretraining summaries.} All three use the equal-category averaging rule in Equation~\ref{eq:benchmark-aggregation}. The core aggregate is used for training trajectories.}
\label{tab:aggregation-scopes}
\accTableFont{10.5}{13}
\accTableStretch{1.12}
\begin{tabularx}{\textwidth}{@{}>{\centering\arraybackslash}m{100pt}cX@{}}
\toprule
Summary & Tasks & \multicolumn{1}{c}{Category scope} \\
\midrule
Core aggregate & 18 & Five: knowledge, common sense, reasoning, mathematics and code. \\
\addlinespace[3pt]
Secondary aggregate & 34 & Six: the core categories plus reading, with 16 additional tasks overall. \\
\addlinespace[3pt]
Full-table summary & 37 & Five display groups; knowledge and reading are combined under Knowledge and QA. Every displayed task is included. \\
\bottomrule
\end{tabularx}
\end{table}

MBPP and MBPP+ score different program sets (500 and 378 programs, respectively). Their absolute scores therefore reflect both test strength and the composition of the scored sets.
The reference-model category scores use the same evaluation suite as Accordion.
Tasks marked $^{\dagger}$ are excluded from the 18-task and 34-task aggregates.
The 37-task summary in Figure~\ref{fig:public-model-radar} uses the displayed benchmark scores and includes every row in each display group; its Overall is calculated from the five category means before rounding.
Core and secondary aggregates follow the fixed category definitions in Table~\ref{tab:aggregation-scopes}. Paired endpoint intervals use the 18-task core scope; the 37-task aggregate and its 89\% gain-retention estimate are descriptive summaries of the full-suite scores.
Benchmark scores use one decimal place (two for corrected Ouro HumanEval/HumanEval+); source data retain original precision.

\phantomsection
\label{app:table-ranking}
Across performance tables, \textbf{bold} identifies the best distinct displayed value within the comparison group specified in each caption. The second-best value is \underline{underlined} when the group contains at least four available candidates; missing scores do not count. Ties at the displayed precision share the same marking. In public-model comparisons, fills in the corresponding category color highlight Accordion's first and second places, with a stronger tint for first place. These descriptive ranks do not indicate statistical significance; unavailable results, counts and confidence intervals are not ranked.

\subsection{Held-Out Language-Modeling Loss}
\label{app:base-validation}

Increasing inference depth monotonically reduces held-out loss in web text, mathematics and code on identical validation tokens (Table~\ref{tab:base-validation}). Each domain uses the held-out source shard specified above and the same validation indices as during pretraining; Table~\ref{tab:validation-settings} gives the evaluation settings. These losses use Accordion-Base before context extension; Accordion-Base-32K is reported separately in Appendix~\ref{app:base32k-results}.
Token, target and loss-mask checksums agree across budgets and with a repeated evaluation of the annealing model. All 15 repeated losses match the original annealing results at the reported precision, confirming consistency of the evaluation inputs and reported losses.

\begin{table}[H]
\centering
\caption{\textbf{Held-out validation loss of Accordion-Base.} Lower is better. Each domain uses identical validation tokens across budgets. Values are batch-mean point estimates rounded to four decimal places; sequence-level records for paired intervals are unavailable. The endpoint reductions reported in \S\ref{sec:domain-gains} use unrounded records. Bold and underlining mark the lowest and second-lowest loss in each row.}
\label{tab:base-validation}
\accTableFont{10.5}{13}
\setlength{\tabcolsep}{8pt}
\accTableStretch{1.12}
\begin{tabular}{@{}cccccc@{}}
\toprule
\multirow{2}{*}[-2.5pt]{Domain} & \multicolumn{5}{c}{Accordion-Base} \\
\cmidrule(lr){2-6}
 & $k=0$ & $k=3$ & $k=6$ & $k=9$ & $k=12$ \\
\midrule
Web & 2.3059 & 2.2976 & 2.2935 & \underline{2.2921} & \textbf{2.2919} \\
Math & 1.3505 & 1.3419 & 1.3379 & \underline{1.3367} & \textbf{1.3366} \\
Code & 0.8693 & 0.8629 & 0.8596 & \underline{0.8586} & \textbf{0.8584} \\
\bottomrule
\end{tabular}
\end{table}

\begin{table}[H]
\centering
\caption{\textbf{Held-out loss evaluation settings.} The same settings and validation sequences are used for all five inference budgets in each domain.}
\label{tab:validation-settings}
\accTableFont{10.5}{13}
\accTableStretch{1.12}
\setlength{\tabcolsep}{6pt}
\begin{tabular*}{\textwidth}{@{\extracolsep{\fill}}cccc@{}}
\toprule
Setting & Value & Setting & Value \\
\midrule
Sequences per domain & 1,400 & Expert parallelism degree & 1 \\
Tokens per sequence & 8,192 & Micro-batch / global batch (sequences) & 4 / 28 \\
Loss tokens per domain & 11,468,800 & Evaluation iterations & 50 \\
GPUs & 7 & Random seed & 1234 \\
\bottomrule
\end{tabular*}
\end{table}

\subsection{Common-Protocol Model Specifications}
The comparison uses Accordion-Base, Qwen3-0.6B/1.7B, OLMoE-0125 and Ouro-1.4B under the protocol in Appendix~\ref{app:base-eval-protocol}.
The OLMoE training configuration combines 5,033B first-stage tokens with a 100B-token annealing stage, for approximately 5.1T tokens.
Ouro-1.4B is configured for four recurrent steps with early exit at the third step. It uses unfused attention, with one example per batch on generation tasks.
\begin{table}[H]
  \centering
\accTableStretch{1.12}
  \caption{\textbf{Base-model comparison specifications.} Parameter counts are in billions and retain source conventions. Active shared weights are not multiplied by recurrence count. Accordion's active count includes its output head (0.54B without it) and describes an unlooped pass. Training exposure excludes subsequent SFT/RL.}
  \label{tab:public-model-specs}
  \accTableFont{10.5}{13}
  \setlength{\tabcolsep}{2pt}
\begin{tabular*}{\textwidth}{@{\extracolsep{\fill}}cccccc@{}}
\toprule
Property & \color{black}\bfseries \accTableLogo Accordion-Base & Qwen3-0.6B & Qwen3-1.7B & OLMoE-0125 & Ouro-1.4B \\
\midrule
Architecture & Layer-loop MoE & Dense & Dense & MoE & LoopLM \\
Total parameters (B) & 6.03 & 0.6 & 1.7 & 6.9 & 1.4 \\
Active parameters (B) & 0.74 & 0.6 & 1.7 & 1.3 & 1.4 \\
Training tokens & 3.6T & 36T & 36T & $\sim$5.1T & 7.7T \\
Execution & $k=0,3,6,9,12$ & 1 pass & 1 pass & 1 pass & R=4 \\
\bottomrule
\end{tabular*}
\end{table}

\subsection{Paired Evidence for Depth Gains}
Additional depth improves all 18 core-task point estimates, with paired 95\% intervals excluding zero for 12 tasks (Table~\ref{tab:pretrain-endpoint-uncertainty}). The core aggregate increases from 52.49 at $k=0$ to 56.23 at $k=12$ (+3.74 points; paired 95\% interval $[3.17,4.34]$).
Within this scope, $k=6$ retains 83\% of the endpoint gain as a descriptive ratio.

\begin{table}[H]
\centering
\caption{\textbf{Paired endpoint gains on the core suite.} Accordion-Base differences between $k=12$ and $k=0$ (percentage points), with paired bootstrap 95\% confidence intervals reflecting evaluation-sample uncertainty.}
\label{tab:pretrain-endpoint-uncertainty}
\accTableFont{10.5}{13}
\accTableStretch{1.0}
\setlength{\tabcolsep}{3pt}
\begin{tabular*}{\textwidth}{@{\extracolsep{\fill}}ccc@{\hspace{18pt}}ccc@{}}
\toprule
Task & Gain & 95\% CI & Task & Gain & 95\% CI \\
\midrule
MMLU & $+2.63$ & $[+2.02,+3.21]$ & BoolQ & $+0.98$ & $[+0.03,+1.96]$ \\
TriviaQA & $+0.55$ & $[+0.19,+0.91]$ & PIQA & $+0.49$ & $[-0.33,+1.31]$ \\
Natural Questions & $+0.50$ & $[-0.11,+1.14]$ & BBH & $+6.01$ & $[+5.01,+7.06]$ \\
ARC-C & $+1.45$ & $[+0.09,+2.73]$ & GSM8K & $+8.26$ & $[+6.07,+10.61]$ \\
ARC-E & $+0.59$ & $[-0.04,+1.26]$ & Full MATH & $+2.64$ & $[+1.54,+3.74]$ \\
CommonsenseQA & $+3.52$ & $[+1.80,+5.41]$ & HumanEval & $+4.27$ & $[-1.22,+9.76]$ \\
HellaSwag & $+0.97$ & $[+0.65,+1.27]$ & HumanEval+ & $+3.05$ & $[-2.44,+8.54]$ \\
WinoGrande & $+1.10$ & $[-0.63,+3.00]$ & MBPP & $+5.20$ & $[+2.40,+8.40]$ \\
OpenBookQA & $+2.20$ & $[+0.40,+4.00]$ & MBPP+ & $+5.82$ & $[+2.38,+9.26]$ \\
\bottomrule
\end{tabular*}
\end{table}

\subsection{Model-Design Study Configuration}
\label{app:design-study-config}
The model-design ablations in \S\ref{sec:shared-path-design} evaluate shared-path design choices in a 1.61B-parameter configuration trained on 80B tokens. Table~\ref{tab:design-model-settings} specifies this experimental configuration; \S\ref{sec:pretraining} specifies the main Accordion model.

\begin{table}[H]
  \centering
  \accTableStretch{1.12}
  \caption{\textbf{Configuration for the 1.61B-parameter design studies.} Active parameters refer to a single unlooped forward pass and exclude the input embedding and output head. WSD denotes warmup--stable--decay.}
  \label{tab:design-model-settings}
  \accTableFont{10.5}{13}
  \renewcommand{\arraystretch}{1.08}
  \setlength{\tabcolsep}{5pt}
  \begin{tabular*}{\textwidth}{@{\extracolsep{\fill}}cc@{}}
    \toprule
    Setting & Value \\
    \midrule
    Layers / hidden size / Q / KV / head dim & 24 / 768 / 6 / 2 / 128 \\
    Routed / shared / active experts; intermediate size & 128 / 1 / 8; 192 \\
    Total / active parameters; vocabulary & 1.61B / 0.14B; 129{,}280 \\
    Loopable window / budgets & 7--18 / $\{0,3,6,9,12\}$ \\
    \midrule
    Training tokens / context / tokens per step & 80B / 8K / 2.1M \\
    Peak LR / final LR / schedule & $7.07\times10^{-4}$ / 1\% of peak / WSD \\
    Warmup / decay & 1{,}000 steps / final 10\% \\
    AdamW $\beta$ / weight decay / gradient clipping & $(0.9,0.95)$ / 0.1 / 1.0 \\
    Weight initialization & std. 0.006, truncated at $\pm3\sigma$ \\
    Gates $g$ & zero initialization, no weight decay \\
    MoE balancing & per-pass bias; sequence auxiliary loss $10^{-4}$ \\
    Precision / GPUs & bfloat16 / $64\times$ H800 \\
    \bottomrule
  \end{tabular*}
\end{table}

\subsection{Accordion-Base-32K Evaluation}
\label{app:base32k-results}
Accordion-Base-32K retains the five-budget interface after context extension. Table~\ref{tab:base32k-benchmarks} reports all 185 task scores; Table~\ref{tab:base32k-summary} summarizes the five capability categories. Table~\ref{tab:base32k-validation} reports held-out losses on the same 8K sequences used for Accordion-Base. Long-context results are discussed in \S\ref{sec:context-extension}, with the full protocol and task scores below.

\begin{table}[H]
\centering
\caption{\textbf{Category-level performance of Accordion-Base-32K.} Categories average the displayed scores in Table~\ref{tab:base32k-benchmarks}, including the Ouro correction. Reference columns follow the main-text order.}
\label{tab:base32k-summary}
\accTableFont{10.5}{13}
\accTableStretch{1.05}
\setlength{\tabcolsep}{2pt}
\begin{tabular*}{\textwidth}{@{\extracolsep{\fill}}l*{9}{c}@{}}
\toprule
 & \multicolumn{5}{c}{Accordion-Base-32K} & \multicolumn{2}{c}{Qwen3} & OLMoE & Ouro \\
Category & $k=0$ & $k=3$ & $k=6$ & $k=9$ & $k=12$ & 0.6B & 1.7B & 0125 & 1.4B \\
\midrule
Knowledge and QA & 49.27 & 49.94 & 50.34 & 50.43 & \cellcolor{accRankKnowledgeSecond}\underline{50.55} & 41.80 & 49.65 & 50.44 & \textbf{51.36} \\
Common sense & 65.28 & 66.18 & 66.61 & 66.56 & \cellcolor{accRankCommonSecond}\underline{66.62} & 57.00 & 65.33 & 66.60 & \textbf{67.67} \\
Reasoning & 39.13 & 40.60 & 42.30 & 42.43 & 42.57 & 33.90 & \underline{43.37} & 31.20 & \textbf{48.23} \\
Mathematics & 40.25 & 41.83 & 43.40 & 43.85 & 43.66 & 42.41 & \textbf{53.18} & 30.53 & \underline{50.30} \\
Code & 40.87 & 43.52 & 44.20 & 44.95 & 44.65 & 31.67 & \textbf{46.13} & 17.67 & \underline{45.28} \\
\bottomrule
\end{tabular*}
\end{table}

\begin{table}[H]
\centering
\caption{\textbf{Base-model comparison across all 37 benchmarks.} All models follow the protocol in Appendix~\ref{app:base-eval-protocol}. Accordion-Base-32K uses 3.65T training tokens, including 50B for context extension. Category scores appear in Table~\ref{tab:base32k-summary}.}
\label{tab:base32k-benchmarks}
\accTableFont{10.5}{12}
\accTableStretch{1.0}
\setlength{\tabcolsep}{2pt}
\begin{tabular}{@{}>{\centering\arraybackslash}m{108pt}*{5}{>{\centering\arraybackslash}m{\dimexpr(\textwidth-108pt-18\tabcolsep-144pt)/5\relax}}*{4}{>{\centering\arraybackslash}m{36pt}}@{}}
\toprule
 & \multicolumn{5}{c}{\cellcolor{accTableOurs}\color{black}\textbf{\accTableLogo Accordion-Base-32K}} & & & & \\
\multirow{-2}{*}{Benchmark} & \cellcolor{accTableOurs}$k=0$ & \cellcolor{accTableOurs}$k=3$ & \cellcolor{accTableOurs}$k=6$ & \cellcolor{accTableOurs}$k=9$ & \cellcolor{accTableOurs}$k=12$ & \multirow{-2}{*}{\makecell[c]{Qwen3\\0.6B}} & \multirow{-2}{*}{\makecell[c]{Qwen3\\1.7B}} & \multirow{-2}{*}{\makecell[c]{OLMoE\\0125}} & \multirow{-2}{*}{\makecell[c]{Ouro\\1.4B}} \\
\midrule
Pretraining tokens & \multicolumn{5}{c}{3.65T} & 36T & 36T & 5.1T & 7.7T \\
\addlinespace[3pt]
\rowcolor[HTML]{F1ECE5}\multicolumn{10}{c}{\bfseries Knowledge and QA} \\
MMLU & 58.4 & 60.3 & 61.1 & 61.3 & 61.4 & 52.5 & \underline{62.7} & 55.2 & \textbf{67.5} \\
TriviaQA & 52.1 & \cellcolor{accRankKnowledgeSecond}\underline{52.8} & 52.7 & 52.6 & 52.7 & 20.0 & 36.3 & \textbf{62.7} & 43.8 \\
Natural Questions & 19.2 & 19.8 & 20.0 & 19.9 & \cellcolor{accRankKnowledgeSecond}\underline{20.1} & 7.5 & 13.8 & \textbf{25.8} & 17.2 \\
SciQ & 95.8 & \cellcolor{accRankKnowledgeSecond}\underline{95.9} & 95.8 & 95.7 & 95.6 & 93.6 & \textbf{96.1} & 93.9 & 95.8 \\
MMLU (cloze)$^{\dagger}$ & 43.9 & 44.7 & \cellcolor{accRankKnowledgeSecond}\underline{45.4} & \cellcolor{accRankKnowledgeFirst}\textbf{45.5} & \cellcolor{accRankKnowledgeFirst}\textbf{45.5} & 36.6 & 43.5 & 43.7 & 45.1 \\
MMLU-Pro$^{\dagger}$ & 31.8 & 34.9 & 35.4 & 36.0 & 36.0 & 25.7 & \underline{37.5} & 18.3 & \textbf{48.9} \\
DROP & 4.4 & 4.5 & 4.6 & 4.7 & 4.7 & 4.6 & \underline{15.3} & \textbf{32.1} & 4.7 \\
SQuAD & 57.1 & 56.8 & 57.9 & 57.7 & \cellcolor{accRankKnowledgeSecond}\underline{58.3} & 56.5 & \textbf{59.3} & 42.9 & 57.1 \\
CoQA & 78.1 & 78.9 & 79.2 & 79.3 & \cellcolor{accRankKnowledgeSecond}\underline{79.7} & 73.9 & 79.1 & 71.8 & \textbf{81.5} \\
RACE & 37.3 & 36.3 & 36.5 & 36.8 & 36.9 & 35.1 & \textbf{39.5} & 37.9 & \underline{38.3} \\
LAMBADA & 63.9 & 64.4 & 65.1 & \cellcolor{accRankKnowledgeSecond}\underline{65.2} & 65.1 & 53.8 & 63.1 & \textbf{70.5} & 65.1 \\
\addlinespace[3pt]
\rowcolor[HTML]{EDF4F1}\multicolumn{10}{c}{\bfseries Common sense} \\
ARC-C & 57.9 & 58.9 & 59.0 & \cellcolor{accRankCommonFirst}\textbf{60.0} & \cellcolor{accRankCommonSecond}\underline{59.9} & 45.0 & 55.4 & 58.2 & \textbf{60.0} \\
ARC-E & 82.2 & 83.0 & 82.8 & 82.9 & \cellcolor{accRankCommonSecond}\underline{83.1} & 74.7 & 82.2 & 82.5 & \textbf{84.0} \\
CommonsenseQA & 64.1 & 66.2 & 68.0 & 67.1 & 67.3 & 62.9 & \underline{75.6} & 65.6 & \textbf{77.0} \\
HellaSwag & 74.3 & 75.0 & \cellcolor{accRankCommonSecond}\underline{75.4} & \cellcolor{accRankCommonSecond}\underline{75.4} & \cellcolor{accRankCommonSecond}\underline{75.4} & 53.6 & 67.1 & \textbf{79.7} & 74.4 \\
WinoGrande & 66.6 & 67.8 & 69.2 & 68.4 & 68.7 & 59.8 & 66.1 & \textbf{71.9} & \underline{70.8} \\
OpenBookQA & 41.6 & 42.8 & 42.8 & \cellcolor{accRankCommonSecond}\underline{43.4} & 42.8 & 34.2 & 38.4 & \textbf{44.4} & 41.4 \\
BoolQ & 78.7 & \cellcolor{accRankCommonFirst}\textbf{79.9} & \cellcolor{accRankCommonSecond}\underline{79.8} & 79.6 & \cellcolor{accRankCommonFirst}\textbf{79.9} & 69.6 & 78.9 & 70.2 & \textbf{79.9} \\
PIQA & 77.6 & 77.4 & 77.8 & 78.0 & 78.2 & 69.9 & 75.8 & \textbf{79.7} & \underline{78.6} \\
Social IQa & 44.5 & 44.6 & 44.7 & 44.2 & 44.3 & 43.3 & \textbf{48.5} & \underline{47.2} & 42.9 \\
\addlinespace[3pt]
\rowcolor[HTML]{F1EEF6}\multicolumn{10}{c}{\bfseries Reasoning} \\
BBH & 54.8 & 58.2 & 61.0 & \cellcolor{accRankReasoningSecond}\underline{61.3} & \cellcolor{accRankReasoningSecond}\underline{61.3} & 40.8 & 53.3 & 32.6 & \textbf{70.8} \\
MuSR & 36.9 & 37.6 & 38.0 & 37.7 & 38.1 & 34.7 & \textbf{44.3} & 35.5 & \underline{39.0} \\
AGIEval-en & 25.7 & 26.0 & 27.9 & 28.3 & 28.3 & 26.2 & \underline{32.5} & 25.5 & \textbf{34.9} \\
\addlinespace[3pt]
\rowcolor[HTML]{F8F4E8}\multicolumn{10}{c}{\bfseries Mathematics} \\
GSM8K & 55.5 & 60.0 & 61.5 & 62.6 & 62.8 & 59.2 & \underline{71.3} & 51.1 & \textbf{78.6} \\
MATH (5,000) & 39.7 & 39.9 & \cellcolor{accRankMathSecond}\underline{41.1} & \cellcolor{accRankMathFirst}\textbf{42.3} & \cellcolor{accRankMathFirst}\textbf{42.3} & 28.5 & 38.9 & 16.9 & 22.0 \\
MATH500$^{\dagger}$ & 40.0 & 41.8 & 41.6 & \cellcolor{accRankMathFirst}\textbf{42.6} & \cellcolor{accRankMathSecond}\underline{42.4} & 31.8 & 39.4 & 17.2 & 21.4 \\
GSM8K-Platinum & 58.2 & 62.0 & 65.1 & 65.6 & 65.3 & 62.5 & \underline{75.8} & 53.1 & \textbf{80.9} \\
GSM-Plus-mini & 38.5 & 38.5 & 41.5 & 42.2 & 41.8 & 33.2 & \underline{50.4} & 33.8 & \textbf{54.9} \\
MathQA & 46.4 & 47.2 & 47.9 & 47.7 & \cellcolor{accRankMathSecond}\underline{48.0} & 36.8 & 45.2 & 28.0 & \textbf{49.8} \\
Multi-digit arithmetic & 32.2 & 32.0 & 35.8 & 33.0 & 32.2 & 87.3 & \textbf{96.5} & 43.7 & \underline{92.2} \\
ASDiv & 11.5 & 13.2 & 12.7 & \cellcolor{accRankMathFirst}\textbf{14.8} & \cellcolor{accRankMathSecond}\underline{14.5} & 0.0 & 7.9 & 0.4 & 2.6 \\
\addlinespace[3pt]
\rowcolor[HTML]{F7EEEE}\multicolumn{10}{c}{\bfseries Code} \\
HumanEval & 38.4 & 42.7 & 42.7 & \cellcolor{accRankCodeSecond}\underline{44.5} & 43.3 & 28.7 & \textbf{50.6} & 11.6 & 35.37 \\
HumanEval+ & 36.0 & 39.0 & 39.6 & \cellcolor{accRankCodeSecond}\underline{40.2} & 39.6 & 22.6 & \textbf{43.9} & 11.0 & 32.32 \\
MBPP & 48.6 & 51.6 & 51.8 & 53.8 & 53.4 & 39.0 & \underline{55.6} & 21.0 & \textbf{58.0} \\
MBPP+ & 63.0 & 64.8 & 66.4 & 66.9 & 67.2 & 54.5 & \underline{70.6} & 29.9 & \textbf{76.7} \\
CRUXEval output & 23.3 & 23.5 & \cellcolor{accRankCodeSecond}\underline{25.6} & 25.0 & 25.5 & 16.1 & 21.8 & 13.1 & \textbf{27.5} \\
CRUXEval input & 35.9 & \cellcolor{accRankCodeSecond}\underline{39.5} & 39.1 & 39.3 & 38.9 & 29.1 & 34.3 & 19.4 & \textbf{41.8} \\
\bottomrule
\end{tabular}
\par\vspace{3pt}
\begin{minipage}{\textwidth}
\accTableFont{10.5}{12}
\textbf{Bold}/\underline{underlining}: row best/second-best; fills mark Accordion ranks (ties included). $^{\dagger}$ Diagnostics. MATH variants remain separate. Ouro HumanEval/HumanEval+: trailing prompt whitespace removed; corrected scores used throughout.
\end{minipage}
\end{table}

\begin{table}[H]
\centering
\caption{\textbf{Accordion-Base-32K held-out loss.} The 8K evaluation settings match Table~\ref{tab:base-validation}; lower is better.}
\label{tab:base32k-validation}
\accTableFont{10.5}{13}
\accTableStretch{1.12}
\begin{tabular*}{0.85\textwidth}{@{\extracolsep{\fill}}cccccc@{}}
\toprule
Domain & $k=0$ & $k=3$ & $k=6$ & $k=9$ & $k=12$ \\
\midrule
Web & 2.3007 & 2.2923 & 2.2883 & \underline{2.2869} & \textbf{2.2867} \\
Math & 1.3436 & 1.3347 & 1.3308 & \underline{1.3296} & \textbf{1.3295} \\
Code & 0.8681 & 0.8618 & 0.8586 & \underline{0.8576} & \textbf{0.8574} \\
\bottomrule
\end{tabular*}
\end{table}

\subsubsection{Long-Context Evaluation Protocol}

\textbf{Tasks and generation.} We evaluate 13 RULER tasks: eight needle-retrieval tasks, variable tracking, two word-aggregation tasks and two QA tasks. Inputs use the training tokenizer at 4K, 8K, 16K and 32K, with greedy generation.
The scorer measures the fraction of reference targets recovered for retrieval, tracking and aggregation; QA receives credit when any reference answer is found. Scores are averaged equally across the 13 tasks. Accordion-Base-32K uses 500 examples per task and context length at all five budgets $k=0,3,6,9,12$, with budget contrasts paired on identical examples.

\begin{table}[H]
\centering
\accTableStretch{1.12}
\caption{\textbf{RULER for Accordion-Base-32K.} Scores average 13 tasks (\%), using 500 examples per task and context length. The last column gives the paired $k=0$ to $k=12$ gain [95\% CI] in percentage points. Bold and underline mark the highest and second-highest scores within each row.}
\label{tab:long-context}
\accTableFont{10.5}{13}
\renewcommand{\arraystretch}{1.20}
\setlength{\tabcolsep}{2pt}
\begin{tabular*}{\textwidth}{@{\extracolsep{\fill}}ccccccc@{}}
\toprule
Context & $k=0$ & $k=3$ & $k=6$ & $k=9$ & $k=12$ & $\Delta_{12}$ [95\% CI] \\
\midrule
4K & 86.52 & 86.29 & 86.87 & \textbf{87.27} & \underline{87.23} & $+0.72\;[0.24,\,1.20]$ \\
8K & 78.15 & 79.18 & 79.44 & \textbf{79.97} & \underline{79.95} & $+1.79\;[1.17,\,2.40]$ \\
16K & 70.10 & 73.94 & \textbf{74.38} & \underline{74.15} & 74.03 & $+3.93\;[3.16,\,4.66]$ \\
32K & 49.49 & 51.80 & 54.58 & \underline{54.90} & \textbf{54.95} & $+5.47\;[4.69,\,6.23]$ \\
\bottomrule
\end{tabular*}
\end{table}

\textbf{Paired uncertainty estimates.} Confidence intervals use task-stratified paired bootstrap with 10,000 resamples: examples are resampled within each task, the same indices are used for both budgets, and the 13 task gains are averaged equally. Gains are calculated before rounding. Intervals quantify evaluation-sample uncertainty conditional on the trained checkpoint.
The scorer matches reference answers anywhere in the generated continuation after lowercasing and control-character normalization.
Context and output together can exceed the nominal sequence length by approximately 50 tokens.

Endpoint-gain intervals exclude zero at all four context lengths (Table~\ref{tab:long-context}). At 32K, the improvement from $k=3$ to $k=9$ is concentrated in variable tracking and multi-key retrieval. Tables~\ref{tab:ruler-tasks-short} and~\ref{tab:ruler-tasks-long} retain all task scores.

\begin{table}[H]
\centering
\caption{\textbf{Complete RULER task scores at 4K and 8K.} Scores are percentages over 500 examples per task and budget. S: single-needle; MK: multi-key; MQ: multi-query; MV: multi-value; VT: variable tracking; CWE/FWE: common/frequent-word extraction; QA1/2: SQuAD/HotpotQA. Bold/underline mark the highest/second-highest scores per row, including ties.}
\label{tab:ruler-tasks-short}
\accTableFont{10.5}{13}
\accTableStretch{1.0}
\setlength{\tabcolsep}{3pt}
\begin{minipage}[t]{0.485\textwidth}
\begin{tabular*}{\linewidth}{@{\extracolsep{\fill}}cccccc@{}}
\toprule
\multicolumn{6}{c}{\textbf{4K context}} \\

Task & $k=0$ & $k=3$ & $k=6$ & $k=9$ & $k=12$ \\
\midrule
S1 & \textbf{100.00} & \textbf{100.00} & \textbf{100.00} & \textbf{100.00} & \textbf{100.00} \\
S2 & \textbf{100.00} & \textbf{100.00} & \textbf{100.00} & \textbf{100.00} & \textbf{100.00} \\
S3 & \textbf{97.20} & 93.20 & 95.00 & \underline{95.40} & \underline{95.40} \\
MK1 & 95.60 & \textbf{96.60} & \underline{96.40} & \textbf{96.60} & 96.00 \\
MK2 & 99.40 & \underline{99.80} & \textbf{100.00} & \textbf{100.00} & \textbf{100.00} \\
MK3 & 97.20 & 99.20 & \underline{99.60} & \textbf{99.80} & \textbf{99.80} \\
MQ & 94.55 & \textbf{95.45} & 94.30 & 94.40 & \underline{94.65} \\
MV & \textbf{95.40} & \underline{92.05} & 91.55 & 91.95 & 91.85 \\
VT & 89.92 & 92.08 & 93.04 & \underline{94.40} & \textbf{94.52} \\
CWE & \textbf{60.28} & 58.04 & 58.46 & 59.40 & \underline{59.50} \\
FWE & 84.20 & 78.53 & 84.93 & \textbf{86.00} & \underline{85.53} \\
QA1 & 67.00 & \underline{70.80} & 70.60 & \underline{70.80} & \textbf{71.20} \\
QA2 & 44.00 & \textbf{46.00} & 45.40 & \underline{45.80} & 45.60 \\
\bottomrule
\end{tabular*}
\end{minipage}
\hfill
\begin{minipage}[t]{0.485\textwidth}
\begin{tabular*}{\linewidth}{@{\extracolsep{\fill}}cccccc@{}}
\toprule
\multicolumn{6}{c}{\textbf{8K context}} \\

Task & $k=0$ & $k=3$ & $k=6$ & $k=9$ & $k=12$ \\
\midrule
S1 & \textbf{100.00} & \textbf{100.00} & \textbf{100.00} & \textbf{100.00} & \textbf{100.00} \\
S2 & \textbf{100.00} & \textbf{100.00} & \textbf{100.00} & \textbf{100.00} & \textbf{100.00} \\
S3 & \textbf{88.40} & 83.40 & 83.00 & \underline{83.60} & \underline{83.60} \\
MK1 & 87.20 & 91.60 & 93.80 & \underline{94.80} & \textbf{95.00} \\
MK2 & 97.60 & \underline{99.20} & \textbf{99.40} & 99.00 & 98.80 \\
MK3 & 90.00 & \textbf{96.00} & 94.40 & 95.60 & \underline{95.80} \\
MQ & 90.00 & \textbf{92.10} & 90.70 & 91.15 & \underline{91.35} \\
MV & \textbf{88.20} & \underline{86.75} & 86.65 & 85.75 & 86.35 \\
VT & 68.48 & 73.08 & 78.28 & \textbf{81.52} & \underline{80.28} \\
CWE & 42.40 & \textbf{44.54} & 43.46 & \underline{43.60} & 43.40 \\
FWE & \textbf{72.73} & 69.07 & 69.07 & \underline{69.13} & \underline{69.13} \\
QA1 & 47.80 & 50.40 & 50.40 & \textbf{51.00} & \underline{50.60} \\
QA2 & 43.20 & 43.20 & 43.60 & \underline{44.40} & \textbf{45.00} \\
\bottomrule
\end{tabular*}
\end{minipage}
\end{table}

\begin{table}[H]
\centering
\caption{\textbf{Complete RULER task scores at 16K and 32K.} The protocol, task abbreviations and ranking marks follow Table~\ref{tab:ruler-tasks-short}. Each task and budget uses 500 examples.}
\label{tab:ruler-tasks-long}
\accTableFont{10.5}{13}
\accTableStretch{1.0}
\setlength{\tabcolsep}{3pt}
\begin{minipage}[t]{0.485\textwidth}
\begin{tabular*}{\linewidth}{@{\extracolsep{\fill}}cccccc@{}}
\toprule
\multicolumn{6}{c}{\textbf{16K context}} \\

Task & $k=0$ & $k=3$ & $k=6$ & $k=9$ & $k=12$ \\
\midrule
S1 & \textbf{100.00} & \textbf{100.00} & \textbf{100.00} & \textbf{100.00} & \textbf{100.00} \\
S2 & \textbf{100.00} & \textbf{100.00} & \textbf{100.00} & \textbf{100.00} & \textbf{100.00} \\
S3 & \textbf{97.80} & \textbf{97.80} & 97.40 & \underline{97.60} & \textbf{97.80} \\
MK1 & 69.20 & 72.20 & 77.80 & \underline{78.80} & \textbf{79.20} \\
MK2 & 86.80 & 91.60 & \textbf{93.80} & 92.80 & \underline{93.20} \\
MK3 & 69.00 & \textbf{81.00} & 79.20 & \underline{80.00} & 77.80 \\
MQ & 76.70 & \textbf{83.25} & \underline{83.20} & 81.70 & 82.95 \\
MV & 81.95 & \textbf{87.80} & \underline{87.65} & 86.40 & 86.20 \\
VT & 35.88 & \underline{52.00} & \textbf{52.64} & 50.16 & 47.76 \\
CWE & \underline{35.02} & \textbf{36.36} & 33.24 & 32.88 & 33.02 \\
FWE & 73.33 & 72.87 & 72.27 & \underline{73.60} & \textbf{73.67} \\
QA1 & 48.60 & 47.80 & 50.20 & \underline{50.40} & \textbf{50.60} \\
QA2 & 37.00 & 38.60 & \underline{39.60} & \underline{39.60} & \textbf{40.20} \\
\bottomrule
\end{tabular*}
\end{minipage}
\hfill
\begin{minipage}[t]{0.485\textwidth}
\begin{tabular*}{\linewidth}{@{\extracolsep{\fill}}cccccc@{}}
\toprule
\multicolumn{6}{c}{\textbf{32K context}} \\

Task & $k=0$ & $k=3$ & $k=6$ & $k=9$ & $k=12$ \\
\midrule
S1 & \textbf{99.00} & 95.20 & \underline{95.40} & 95.20 & 95.20 \\
S2 & 91.20 & 92.00 & \textbf{93.60} & \underline{93.40} & 93.00 \\
S3 & 95.00 & 92.80 & 94.80 & \underline{95.40} & \textbf{95.60} \\
MK1 & 52.00 & 49.80 & \textbf{57.80} & \underline{57.00} & 56.80 \\
MK2 & 41.80 & 57.00 & \textbf{63.80} & \underline{61.40} & 60.80 \\
MK3 & 18.20 & \underline{29.00} & 28.00 & \textbf{29.20} & 28.60 \\
MQ & 56.55 & 59.55 & 60.60 & \underline{62.65} & \textbf{63.45} \\
MV & 51.10 & 53.75 & \underline{57.80} & 57.65 & \textbf{58.10} \\
VT & 9.20 & 16.48 & 30.76 & \underline{33.80} & \textbf{33.88} \\
CWE & \underline{12.86} & \textbf{14.90} & 9.94 & 8.10 & 8.40 \\
FWE & \textbf{58.40} & 52.73 & 53.20 & \underline{56.27} & 55.93 \\
QA1 & 25.20 & 27.80 & \underline{30.20} & 29.40 & \textbf{30.60} \\
QA2 & 32.80 & 32.40 & 33.60 & \textbf{34.20} & \underline{34.00} \\
\bottomrule
\end{tabular*}
\end{minipage}
\end{table}

\paragraph{Evaluation consistency.}
All budgets use the same Accordion-Base-32K weights, tokenizer, examples and scoring rule, with exact agreement in example identities verified across budgets. A cross-implementation check repeats the unlooped evaluation on three 4K tasks using identical examples and the original batch partition; verbatim prediction agreement is 488/500 for multi-value retrieval, 477/500 for common-word extraction and 498/500 for variable tracking; absolute task-score differences are at most 0.25 points.
Budgets $k=3,9$ use a revised inference implementation. The earlier implementation used for the other budgets was not archived, and cross-implementation validation is restricted to the three reported 4K tasks.

\accEvidenceBarrier
\accEvidenceBarrier
\begingroup
\setlength{\parskip}{4pt plus 1pt minus 1pt}
\setlength{\intextsep}{7pt}
\captionsetup{font=small,skip=5pt}
\section{Second-Pass Mechanism and Supporting Probes}
\label{app:training-analysis}
\newcommand{\accMechanismTable}[3]{%
\begin{table}[H]\centering
\caption{#1}\label{#2}
\accTableFont{9}{10}\accTableStretch{1}\setlength{\tabcolsep}{3pt}
#3\end{table}}
\newcommand{\accMechanismFigure}[4][\textwidth]{%
\begin{figure}[H]\centering
\includegraphics[width=#1]{figures/generated/mechanism_#2.pdf}
\caption{#3}\label{#4}\end{figure}}

\subsection{Probe Protocol and Internal Representations}
\label{app:mechanism-setup}
\label{app:mechanism-validation}
\label{app:mechanism-routing}
\label{app:mechanism-updates}
\label{app:mechanism-attention}

We probe Accordion-Base-32K at $k=12$ on 1,800 held-out documents, with 600 each from web text, mathematics and code. Documents have 1,024--15,999 tokens, are sampled without replacement and are deduplicated on their first 1,024 tokens. Each contributes those targets after an end-of-document token. Expert overlap counts the intersection of the two top-eight routed sets, excluding the shared expert. The routing and attention diagnostics adapt SMELT~\citep{wang2026smelt} to in-place layer repetition.

Interventions edit second-pass writes, preserve first-pass execution and recompute downstream activations. We compare teacher-forced NLL with the intact $k=12$ baseline using matched bf16 execution and fixed expert summation order. Intervals use 2,000 paired document bootstrap resamples, stratified by domain, with 95\% percentile bounds. Recording preserves logits; removing all second-pass writes exactly recovers $k=0$, and suffix removals recover intermediate budgets. Fixed-prefix execution makes a repeated layer's upstream computation identical across budgets that include it. Documents were split by a deterministic hash into exploration and confirmation halves; whole-layer removal effects differ by at most 0.3 millinats/token between halves. These probes measure inference-time sensitivity within one training run.

Input-only routing changes the input while retaining the first-pass bias; bias-only routing changes the bias while retaining the first-pass input (Figure~\ref{fig:mechanism-selection-main}). Fixing expert sets preserves second-pass inputs and recomputes mixture weights from second-pass scores. Fixing sets and weights additionally copies first-pass mixture weights.

\accMechanismFigure{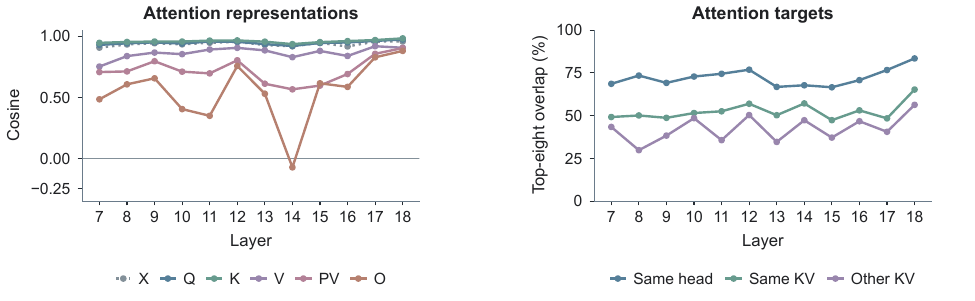}{\textbf{Attention readouts change despite similar target selection.} Left, cross-pass cosines before gating: normalized input ($X$), queries ($Q$), keys ($K$), values ($V$), attention-weighted values ($PV$) and output projection ($O$). Right, top-eight target overlap for the same head across passes (Same head), and within-pass controls using another head in the same or a different KV group (Same KV / Other KV), at query positions at least 64.}{fig:mechanism-attention-detail}

Even with identical routed sets, $\|r_2-r_1\|/\|r_1\|$ ranges from 0.39 to 0.87 across layers, where $r$ is the routed-expert output before gating. These conditional groups range from 861 tokens at layer 9 to 256,756 at layer 14. Their means weight tokens within each group; both inputs and mixture weights can change despite identical expert sets.

Mean gate magnitudes are 0.36--0.69 in layers 7--12 and 0.06--0.11 in layers 16--18, with second-pass training exposures of 60--80\% and 20\%, respectively. Position, gate magnitude and exposure covary under fixed-prefix training.

Write-norm ratios are computed per token using its first-pass write norm plus $10^{-12}$, then averaged within and across documents; no first-pass write has zero norm.

Attention readouts change despite similar target selection (Figure~\ref{fig:mechanism-attention-detail}). Queries and keys retain cosines of 0.92--0.98, while output-projection cosines span $-0.07$--$0.88$. The same head retains 72\% of its top-eight targets, versus 42\% for another KV group and 53\% for the same KV group.

Mean attention mass on the initial EOD token increases by 0.137 in the second pass across the twelve layers. Changed readouts thus coexist with similar target selection and increased attention to this sink position.

\Needspace*{5\baselineskip}
\subsection{Write Interventions and Persistence through Post-Training}
\label{app:mechanism-interventions}
\label{app:mechanism-stages}

Layerwise interventions localize prediction sensitivity across the twelve repeated layers and test the learned write magnitude (Table~\ref{tab:mechanism-edits-compact}). In layers 7--12, both attenuation and amplification raise NLL for attention, MoE and combined writes. Scaling multiplies the already gated second-pass write by $\alpha\in\{0.5,1.5\}$; the intact model has $\alpha=1$. At layer 9, fixing the expert set costs 6.08 [5.78, 6.38] millinats/token, versus 6.73 [6.42, 7.06] for removing the MoE write (95\% intervals).

Second-pass computation contributes to prediction across domains and through post-training. Removing all writes from Accordion-Base-32K costs 14.25, 15.26 and 10.63 millinats/token in web text, mathematics and code, respectively.

Across Base-32K, Instruct and RL, removal increases NLL by 13.38, 9.66 and 7.22 millinats/token, with positive 95\% intervals at every stage (Table~\ref{tab:mechanism-stages}). Common raw-text inputs provide a shared reference for comparing these computations. The largest cross-stage layerwise differences are 0.37 shared experts and 0.12 in MoE norm ratio. Within this persistent profile, removing layer 13's combined write costs 4.3, 8.3 and 6.8 millinats/token, respectively.

\accMechanismTable{\textbf{Complete layerwise intervention effects.} $\Delta$NLL in millinats/token. ``Both'' always refers to second-pass attention and MoE writes, with first-pass execution preserved. The lower block of each panel gives scaling factors.}{tab:mechanism-edits-compact}{
\renewcommand{\arraystretch}{1.40}
\begin{tabular*}{\textwidth}{@{\extracolsep{\fill}}lcccccc@{}}
\toprule
Edit / layer & 7 & 8 & 9 & 10 & 11 & 12 \\
\midrule
Remove attention & 1.86 & 3.11 & 2.71 & 2.34 & 2.45 & 1.98 \\
Remove MoE & 5.32 & 10.23 & 6.73 & 3.03 & 7.80 & 5.62 \\
Remove both writes & 7.94 & 9.07 & 7.64 & 7.28 & 6.86 & 5.66 \\
Fix expert set & 0.57 & 3.23 & 6.08 & 0.08 & 1.93 & 0.21 \\
Fix set and weights & 1.03 & 4.08 & 5.95 & 0.21 & 2.13 & 0.26 \\
\midrule
Attention, $0.5$ & 0.32 & 0.50 & 0.54 & 0.54 & 0.53 & 0.25 \\
Attention, $1.5$ & 0.87 & 1.12 & 0.55 & 0.50 & 0.66 & 0.82 \\
MoE, $0.5$ & 1.23 & 1.80 & 1.59 & 0.83 & 2.02 & 1.19 \\
MoE, $1.5$ & 2.48 & 6.85 & 1.02 & 0.29 & 1.35 & 1.27 \\
Both, $0.5$ & 1.46 & 1.84 & 1.91 & 1.79 & 1.96 & 1.26 \\
Both, $1.5$ & 7.38 & 8.65 & 1.73 & 1.04 & 2.83 & 2.48 \\
\bottomrule
\end{tabular*}
\par\vspace{8pt}
\begin{tabular*}{\textwidth}{@{\extracolsep{\fill}}lcccccc@{}}
\toprule
Edit / layer & 13 & 14 & 15 & 16 & 17 & 18 \\
\midrule
Remove attention & 0.44 & 0.05 & 0.30 & 0.28 & 0.01 & 0.01 \\
Remove MoE & 2.55 & 2.44 & 1.45 & 0.07 & 0.22 & 0.09 \\
Remove both writes & 4.28 & 2.19 & 1.52 & 0.54 & 0.23 & 0.09 \\
Fix expert set & 0.06 & 0.07 & 0.09 & -0.02 & -0.04 & -0.01 \\
Fix set and weights & 0.10 & 0.12 & 0.18 & -0.02 & 0.00 & 0.00 \\
\midrule
Attention, $0.5$ & 0.10 & -0.15 & 0.04 & 0.09 & -0.01 & -0.10 \\
Attention, $1.5$ & 0.11 & 0.42 & 0.12 & -0.03 & 0.00 & 0.04 \\
MoE, $0.5$ & 0.74 & 0.45 & 0.35 & 0.00 & 0.03 & 0.00 \\
MoE, $1.5$ & 0.18 & 0.67 & 0.30 & -0.01 & 0.06 & -0.01 \\
Both, $0.5$ & 1.13 & 0.30 & 0.33 & 0.14 & -0.01 & -0.05 \\
Both, $1.5$ & 0.35 & 1.17 & 0.41 & -0.02 & 0.06 & 0.09 \\
\bottomrule
\end{tabular*}
}

\accMechanismTable{\textbf{Second-pass computation remains useful across training stages.} Baseline NLL is in nats/token; intervention effects and their 95\% document bootstrap intervals are in millinats/token. Each stage uses its own intact baseline on the same raw documents, without chat templates.}{tab:mechanism-stages}{
\begin{tabular*}{\textwidth}{@{\extracolsep{\fill}}lccc@{}}
\toprule
Measurement & Base-32K & Instruct & RL \\
\midrule
Baseline NLL & 1.5674 & 1.6974 & 1.7046 \\
Remove all writes & 13.38 [12.92, 13.87] & 9.66 [8.92, 10.41] & 7.22 [6.44, 7.97] \\
Remove all MoE & 38.23 [37.28, 39.18] & 43.04 [41.76, 44.30] & 47.20 [45.87, 48.53] \\
Remove all attention & 16.94 [16.38, 17.52] & 29.79 [28.82, 30.70] & 29.65 [28.68, 30.56] \\
Fix all expert sets & 12.58 [12.10, 13.06] & 17.98 [17.22, 18.71] & 19.56 [18.78, 20.33] \\
\bottomrule
\end{tabular*}
}

\Needspace*{5\baselineskip}
\subsection{Layer-Group Structure of Nonadditive Effects}
\label{app:mechanism-interactions}

Fifteen group interventions compare contiguous and spaced removals to characterize how layer arrangement shapes joint effects (Table~\ref{tab:mechanism-groups-complete}), including the equal-size triplets in Table~\ref{tab:mechanism-interaction-main}. For a removed set $S$, the additive reference is $\sum_{\ell\in S}\Delta\mathrm{NLL}_{\ell}$; the joint/sum ratio divides the measured joint effect by this reference. Ratios below one indicate subadditivity. All values use the same paired $k=12$ baseline.

\accMechanismTable{\textbf{Joint effects for every tested three-, six- and nine-layer set.} Effects and intervals are in millinats/token. Sums use unrounded single-layer estimates. Ratio intervals resample the joint and single-layer effects on the same documents.}{tab:mechanism-groups-complete}{
\begin{tabular*}{\textwidth}{@{\extracolsep{\fill}}lccccc@{}}
\toprule
Removed layers & 16--18 & 13--15 & 10--12 & 7--9 & 7, 11, 15 \\
\midrule
Joint effect & 0.25 & 1.23 & 3.61 & 8.07 & 16.49 \\
95\% interval & [0.18, 0.32] & [1.09, 1.37] & [3.37, 3.84] & [7.72, 8.46] & [15.98, 17.04] \\
Sum of singles & 0.86 & 8.00 & 19.79 & 24.65 & 16.32 \\
Joint / sum & 0.29 & 0.15 & 0.18 & 0.33 & 1.01 \\
Ratio 95\% interval & [0.23, 0.36] & [0.14, 0.17] & [0.17, 0.19] & [0.32, 0.34] & [1.00, 1.03] \\
\bottomrule
\end{tabular*}
\par\vspace{8pt}
\begin{tabular*}{\textwidth}{@{\extracolsep{\fill}}lccccc@{}}
\toprule
Removed layers & 8, 12, 16 & 9, 13, 17 & 10, 14, 18 & 13--18 & 10--15 \\
\midrule
Joint effect & 15.81 & 12.19 & 9.13 & 1.45 & 4.82 \\
95\% interval & [15.28, 16.37] & [11.78, 12.64] & [8.76, 9.53] & [1.30, 1.59] & [4.56, 5.08] \\
Sum of singles & 15.27 & 12.15 & 9.56 & 8.86 & 27.79 \\
Joint / sum & 1.03 & 1.00 & 0.96 & 0.16 & 0.17 \\
Ratio 95\% interval & [1.02, 1.05] & [0.99, 1.02] & [0.94, 0.97] & [0.15, 0.18] & [0.17, 0.18] \\
\bottomrule
\end{tabular*}
\par\vspace{8pt}
\begin{tabular*}{\textwidth}{@{\extracolsep{\fill}}lccccc@{}}
\toprule
Removed layers & 7--12 & \shortstack{7, 9, 11\\13, 15, 17} & \shortstack{8, 10, 12\\14, 16, 18} & 10--18 & 7--15 \\
\midrule
Joint effect & 11.89 & 23.83 & 23.20 & 5.01 & 13.18 \\
95\% interval & [11.44, 12.36] & [23.19, 24.57] & [22.52, 23.96] & [4.74, 5.28] & [12.73, 13.65] \\
Sum of singles & 44.44 & 28.47 & 24.83 & 28.65 & 52.44 \\
Joint / sum & 0.27 & 0.84 & 0.93 & 0.17 & 0.25 \\
Ratio 95\% interval & [0.26, 0.27] & [0.83, 0.85] & [0.92, 0.95] & [0.17, 0.18] & [0.24, 0.26] \\
\bottomrule
\end{tabular*}
}

All adjacent triplet ratio intervals lie below one; spaced triplet point estimates remain close to one. Contiguous six- and nine-layer sets yield ratios of 0.16--0.27, versus 0.84 and 0.93 for the two spaced six-layer sets. The tested sets include both suffixes and interior segments.

\accMechanismTable{\textbf{Pair effects depend on layer identity.} Seven layer-pair interventions quantify variation in joint effects, with 95\% document bootstrap intervals. Joint $\Delta$NLL is in millinats/token.}{tab:mechanism-pairs-compact}{
\begin{tabular*}{\textwidth}{@{\extracolsep{\fill}}lccccccc@{}}
\toprule
Removed layers & 7, 8 & 8, 9 & 8, 11 & 9, 10 & 9, 11 & 10, 11 & 12, 13 \\
\midrule
Joint effect & 8.78 & 12.91 & 16.60 & 14.45 & 16.56 & 6.70 & 8.61 \\
95\% interval & [8.43, 9.16] & [12.45, 13.43] & [16.07, 17.18] & [13.97, 14.98] & [16.04, 17.13] & [6.40, 7.01] & [8.28, 8.93] \\
Sum of singles & 17.02 & 16.71 & 15.93 & 14.91 & 14.49 & 14.13 & 9.94 \\
Joint / sum & 0.52 & 0.77 & 1.04 & 0.97 & 1.14 & 0.47 & 0.87 \\
Ratio 95\% interval & [0.50, 0.53] & [0.76, 0.79] & [1.03, 1.05] & [0.95, 0.98] & [1.13, 1.16] & [0.46, 0.49] & [0.85, 0.88] \\
\bottomrule
\end{tabular*}
}

Pair interventions refine the group-level finding (Table~\ref{tab:mechanism-pairs-compact}): layers 10+11 have a joint/sum ratio of 0.47, whereas the adjacent pair 9+10 has a ratio of 0.97. Nonadditivity also occurs within a layer: removing layer 8's MoE write costs 10.23 millinats/token, compared with 9.07 for removing both writes. The effect of a write depends on the remaining computation.

These comparisons characterize the tested layer sets: adjacent and spaced groups contain different layers, so they do not isolate layer distance as the cause of the difference. Many internal removals also produce paths outside the five trained prefixes. The results therefore describe sensitivity to these interventions rather than establish why fixed-prefix training outperforms other path designs.

Second-pass writes at layers 7, 8, 10 and 11 oppose the next layer's first-pass write (cosines $-0.45$--$-0.38$, versus $-0.04$--$0.03$ at $k=0$). These alignments characterize update direction along the intact path; the Hydra effect~\citep{mcgrath2023hydra} concerns downstream write changes induced by ablation. Identifying such compensation here would require measuring the downstream response to removal; neither intact-path alignment nor the joint/sum ratio alone establishes it.

\Needspace*{5\baselineskip}
\subsection{Length Dependence and Structured Prediction}
\label{app:mechanism-length}
\label{app:mechanism-dyck}

The following behavioral probes connect additional depth to input structure. They complement the internal measurements with controlled length bins and a structured answer task. The length analysis uses 7,863 held-out documents from the same three domains, divided into nine bins up to 8K, following the length-binned diagnostic in SMELT~\citep{wang2026smelt}. Within each bin, the gain is the token-weighted reduction $\mathrm{NLL}_0-\mathrm{NLL}_k$ on complete documents.

\accMechanismTable{\textbf{Length preference across all nonzero budgets.} Long/short gain ratios for Accordion-Base-32K. Long bins end at 512--4,096 tokens; short bins at 32--256 tokens. The ratio compares unweighted means of the four bin-level gains in each group. Intervals are 95\% document bootstrap intervals.}{tab:mechanism-length-ratio}{
\begin{tabular*}{\textwidth}{@{\extracolsep{\fill}}lcccc@{}}
\toprule
Measurement & $k=3$ & $k=6$ & $k=9$ & $k=12$ \\
\midrule
Long / short gain & 1.15 & 1.29 & 1.44 & 1.44 \\
95\% interval & [0.97, 1.43] & [1.10, 1.57] & [1.23, 1.73] & [1.23, 1.73] \\
\bottomrule
\end{tabular*}
}

Depth gains in the five longer bins remain within 13.2--15.1 millinats/token across checkpoints, while the shortest-bin gain decreases from 20.0 to 4.5. The length and position profiles appear in Figure~\ref{fig:mechanism-length-compact}, and Table~\ref{tab:mechanism-length-ratio} reports the ratios at all four nonzero budgets. At $k=12$, the long/short gain ratio consequently rises from 0.85 at 2.01T tokens to 1.14 at 3T and 1.44 for Accordion-Base-32K: longer-bin gains persist as short-input gains decrease. This within-8K analysis complements the 32K retrieval evaluation in Appendix~\ref{app:pretraining-evidence}.

\accMechanismFigure{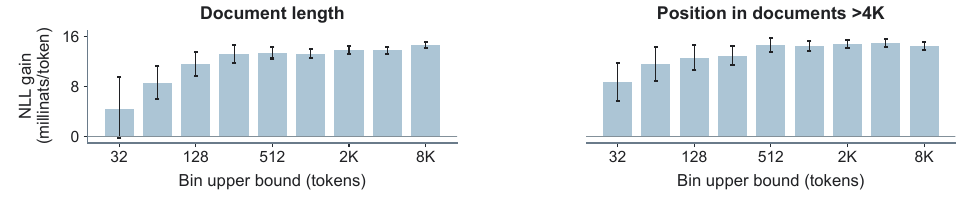}{\textbf{Depth gains across input lengths and positions.} Accordion-Base-32K, $k=12$ versus $k=0$. All nine bins are shown; whiskers give 95\% document bootstrap intervals. The final bin ends at 8,191 tokens.}{fig:mechanism-length-compact}

\paragraph{Bracket completion.}
Additional depth improves structured answer prediction on 400 newly generated prompts in the BIG-bench Dyck Languages format~\citep{srivastava2023beyond}, each with 32 demonstrations. The task predicts the closing sequence for unmatched opening brackets. Teacher-forced exact match requires every target bracket to be the argmax when conditioned on the gold preceding sequence; answer NLL sums the target-bracket losses.

\accMechanismTable{\textbf{Depth improves bracket completion with long demonstrations.} All five budgets on the same 400 prompts with 32 demonstrations. Exact match is teacher-forced, and answer NLL changes are nats per answer relative to $k=0$. At all tested positive demonstration counts below 32, paired exact-match gain intervals include zero.}{tab:mechanism-dyck}{
\begin{tabular*}{\textwidth}{@{\extracolsep{\fill}}lccccc@{}}
\toprule
Metric & $k=0$ & $k=3$ & $k=6$ & $k=9$ & $k=12$ \\
\midrule
Exact match (\%) & 8.50 & 10.00 & 11.75 & 12.00 & 12.00 \\
Answer NLL change & 0.000 & -0.224 & -0.259 & -0.272 & -0.259 \\
\bottomrule
\end{tabular*}
}

With 32 demonstrations, additional depth raises teacher-forced exact match (Table~\ref{tab:mechanism-dyck}) from 8.5\% to 12.0\% (+3.5 points; 95\% question-bootstrap interval [1.25, 5.75]) and reduces answer NLL by 0.259 nats at $k=12$, demonstrating a structured-prediction benefit under long-demonstration inputs.

\paragraph{Depth gains without demonstrations.}
A nine-task sweep (excluding MMLU) measures depth gains at zero and sixteen demonstrations: 1.45 and 0.95 points, respectively. Paired intervals for the change across shot counts include zero. The positive zero-shot point estimate locates a depth benefit without demonstrations.
\label{app:mechanism-end}

\endgroup
\FloatBarrier

\FloatBarrier
\section{Supervised Fine-Tuning Details}
\label{app:sft-details}

Accordion-Instruct is trained for 8,614 steps from Accordion-Base-32K, with all five inference budgets sharing the resulting weights.
The common conversational evaluation protocol supports paired comparisons with Accordion-RL at each budget.

\subsection{Data Sources and Packing}
Training-token counts account for the number of epochs assigned to each of the 12 data components (Table~\ref{tab:sft-data}). Single-pass counts measure packed input tokens, including multiple retained solutions per problem.
The resulting training set contains 71.734B input tokens, of which 68.5B contribute to the loss. The source-level counts in Table~\ref{tab:sft-data} separate mathematics, code, and instruction/other data.

\begin{table*}[!htbp]
  \centering
  \caption{\textbf{SFT training data by source.} All 12 packed training components, with input tokens in billions before and after applying the indicated epochs. Category shading identifies mathematics, code, and instruction/other data. Validation is excluded; the 68.5B loss-bearing tokens are a subset of the 71.734B training input tokens. Totals use unrounded counts.}
  \label{tab:sft-data}
  \accTableFont{10.5}{13}
  \setlength{\tabcolsep}{3.5pt}
  \accTableStretch{1.15}
  \begin{tabular*}{\textwidth}{@{\extracolsep{\fill}}ccccc@{}}
    \toprule
    Category & Dataset / retained subset & Single pass (B) & Epochs & Training (B) \\
    \midrule
\cellcolor[HTML]{F8F3E8} & Nemotron-SFT-Math-v3 & 18.098 & 1 & 18.098 \\
\cellcolor[HTML]{F8F3E8} & OpenMathReasoning (CoT) & 23.592 & 1 & 23.592 \\
\cellcolor[HTML]{F8F3E8}\multirow{-3}{*}{Mathematics} & OpenR1-Math-220k & 1.028 & 1 & 1.028 \\
\addlinespace[3pt]
\cellcolor[HTML]{EDF4F0} & Nemotron-SFT-Comp.-Prog.-v2 (Python) & 1.985 & 1 & 1.985 \\
\cellcolor[HTML]{EDF4F0} & Nemotron-SFT-Comp.-Prog.-v2 (C++) & 2.014 & 1 & 2.014 \\
\cellcolor[HTML]{EDF4F0} & AceReason-1.1-SFT (code) & 9.529 & 1 & 9.529 \\
\cellcolor[HTML]{EDF4F0}\multirow{-4}{*}{Code} & Dolci-Think-SFT-Python (correct solutions) & 3.867 & 1 & 3.867 \\
\addlinespace[3pt]
\cellcolor[HTML]{F2EDF7} & Nemotron instruction/chat-v2 (think) & 2.304 & 2 & 4.608 \\
\cellcolor[HTML]{F2EDF7} & Nemotron instruction/chat-v2 (direct) & 1.320 & 2 & 2.639 \\
\cellcolor[HTML]{F2EDF7} & Nemotron chat-v1 & 1.662 & 2 & 3.323 \\
\cellcolor[HTML]{F2EDF7} & Nemotron Comp.-Prog.-v2 (exercism / SQL) & 0.496 & 2 & 0.993 \\
\cellcolor[HTML]{F2EDF7}\multirow{-5}{*}{Instruction and other} & Nemotron safety-v1 & 0.029 & 2 & 0.058 \\
\midrule
\textbf{Total} & & & & \textbf{71.734} \\
    \bottomrule
  \end{tabular*}
\end{table*}

\subsection{Optimization}
The SFT recipe resets the optimizer state and uses a global batch of 256 packed sequences of length 32,768, corresponding to a nominal 8,388,608 input tokens per update.
The iteration counter also resets; with the budget-sampling seed unchanged, SFT restarts the budget sequence used at the beginning of pretraining.
Micro-batch size is one sequence, with context parallelism of two and data parallelism of 256 across 512 GPUs.
Packed conversations have separate attention boundaries and reset position indices.
Consumed-token counts reflect actual pack occupancy within the nominal batch capacity.

\subsection{Generation and Scoring Protocol}
Evaluation uses vLLM with bfloat16 precision and identical task-specific prompts, response limits and scorers for SFT and RL.
Table~\ref{tab:posttraining-response-limits} groups tasks by their output-token limits.
Sampling and deterministic IFEval scoring follow \S\ref{sec:sft-eval}; the greedy comparison changes only the decoding policy for the eight non-AIME tasks.
Paired percentile intervals resample problems after averaging repeated responses within each problem. Table~\ref{tab:sft-decoding-full} specifies the SFT bootstrap settings; the RL endpoint intervals in Table~\ref{tab:rl-depth-gains} use 2,000 resamples with Python \texttt{random.Random(0)}.
The incomplete-response rate counts answers without a closing reasoning tag; it is distinct from reaching the token limit.

\begin{table}[!htbp]
\centering
\caption{\textbf{Output-token limits for the common post-training evaluation.} The matched SFT and RL comparison uses the same limits. Public references request these limits, capped by their remaining context space as specified in Appendix~\ref{app:posttraining-common}.}
\label{tab:posttraining-response-limits}
\accTableFont{10.5}{13}
\accTableStretch{1.12}
\begin{tabularx}{\textwidth}{@{}>{\raggedright\arraybackslash}m{.60\textwidth}Y@{}}
\toprule
Tasks & Maximum output tokens \\
\midrule
GSM8K, IFEval, IFBench, MMLU-Redux & 8,192 \\
MATH500, HumanEval+, MBPP+ & 16,384 \\
AIME 2024, AIME 2025, LiveCodeBench & 30,000 \\
\bottomrule
\end{tabularx}
\end{table}

\begin{figure}[!htbp]
\centering
\includegraphics[width=4.6in]{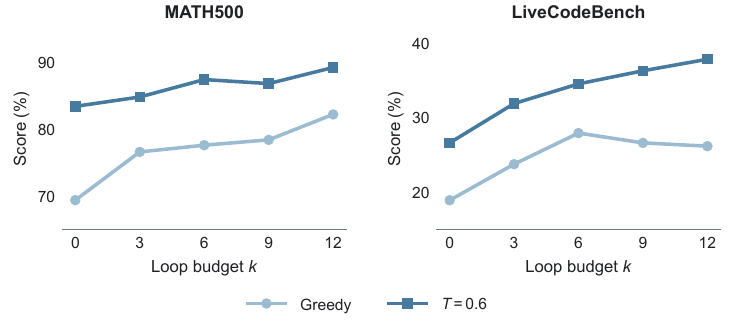}
\caption{\textbf{SFT budget response under greedy and sampled decoding.} Sampling uses $T=0.6$, top-$p=0.95$ and one response per problem. Output limits are held fixed within each task. Complete scores and paired intervals appear in Table~\ref{tab:sft-decoding-full}.}
\label{fig:sft-decoding}
\end{figure}

\begin{table*}[!htbp]
\centering
\accTableStretch{1.12}
\caption{\textbf{SFT decoding policies across all five budgets.} vLLM scores (\%), endpoint differences $\Delta_{12}$ and paired 95\% intervals (percentage points). Differences are calculated before rounding. Intervals use NumPy \texttt{default\_rng(0)}; the resample count is unrecorded in the archived evaluation. Sampled scores serve as the SFT baseline in Table~\ref{tab:rl-final}; the greedy comparison tests depth gains across decoding policies. AIME has one 16-response sampled evaluation. Ranks compare the five budgets within each task and decoding-policy row.}
\label{tab:sft-decoding-full}
\label{tab:sft-vllm-greedy}
\label{tab:sft-vllm-sampled}
\accTableFont{10.5}{13}
\setlength{\tabcolsep}{3pt}
\begin{tabular*}{\textwidth}{@{\hspace{12pt}\extracolsep{\fill}}cccccccc@{\hspace{12pt}}}
\toprule
Task & $k=0$ & $k=3$ & $k=6$ & $k=9$ & $k=12$ & $\Delta_{12}$ & 95\% CI \\
\midrule
\multicolumn{8}{c}{\textit{Greedy (one response per problem)}} \\
GSM8K & 75.82 & 79.76 & 81.20 & \textbf{82.26} & \underline{81.27} & $+5.46$ & $[+3.26,+7.58]$ \\
MATH500 & 69.40 & 76.60 & 77.60 & \underline{78.40} & \textbf{82.20} & $+12.80$ & $[+8.80,+16.60]$ \\
HumanEval+ & 65.24 & 63.41 & \textbf{70.12} & \underline{68.29} & 65.85 & $+0.61$ & $[-6.10,+6.71]$ \\
MBPP+ & 60.05 & 62.17 & \underline{66.40} & 66.14 & \textbf{66.67} & $+6.61$ & $[+2.38,+10.85]$ \\
LiveCodeBench & 18.94 & 23.79 & \textbf{27.97} & \underline{26.65} & 26.21 & $+7.27$ & $[+4.19,+10.35]$ \\
IFEval & 52.31 & 54.34 & \textbf{57.67} & \underline{56.01} & 54.16 & $+1.85$ & $[-2.59,+6.47]$ \\
IFBench & 16.00 & \underline{18.67} & \textbf{20.00} & 18.33 & \underline{18.67} & $+2.67$ & $[-1.67,+7.33]$ \\
MMLU-Redux & 53.26 & 62.27 & \textbf{64.32} & 62.74 & \underline{63.10} & $+9.83$ & $[+8.44,+11.37]$ \\
\midrule
\multicolumn{8}{c}{\textit{Sampling ($T=0.6$, top-$p=0.95$; AIME averages 16 responses)}} \\
GSM8K & 83.40 & 85.67 & \textbf{88.78} & 87.72 & \underline{88.63} & $+5.23$ & $[+3.26,+7.13]$ \\
MATH500 & 83.40 & 84.80 & \underline{87.40} & 86.80 & \textbf{89.20} & $+5.80$ & $[+2.60,+8.80]$ \\
AIME 2024 & 30.00 & 34.79 & 39.58 & \underline{43.54} & \textbf{43.75} & $+13.75$ & $[+8.54,+19.17]$ \\
AIME 2025 & 26.88 & 28.96 & \underline{32.92} & \textbf{36.88} & 31.88 & $+5.00$ & $[+1.46,+8.96]$ \\
HumanEval+ & \underline{79.88} & 78.05 & 79.27 & \underline{79.88} & \textbf{81.71} & $+1.83$ & $[-4.27,+7.32]$ \\
MBPP+ & 69.58 & 67.72 & \textbf{73.02} & 70.90 & \underline{72.75} & $+3.17$ & $[-1.06,+7.14]$ \\
LiveCodeBench & 26.65 & 31.94 & 34.58 & \underline{36.34} & \textbf{37.89} & $+11.23$ & $[+7.49,+14.98]$ \\
IFEval & 62.11 & \underline{64.88} & \textbf{65.99} & 63.40 & 60.07 & $-2.03$ & $[-6.47,+2.40]$ \\
IFBench & 17.33 & \underline{20.33} & \underline{20.33} & 19.33 & \textbf{23.00} & $+5.67$ & $[+1.00,+10.67]$ \\
MMLU-Redux & 60.88 & 63.47 & \underline{65.82} & \textbf{66.17} & 65.16 & $+4.28$ & $[+2.80,+5.67]$ \\
\bottomrule
\end{tabular*}
\end{table*}
The gains from additional depth persist under both decoding policies (Figure~\ref{fig:sft-decoding}). All eight non-AIME tasks improve between $k=0$ and $k=12$ under greedy decoding, and seven improve under sampling. Under each policy, five of these task gains have paired 95\% intervals excluding zero. Both AIME tasks use the same sampled evaluation in the two comparisons and also have positive endpoint intervals.
Sampled LiveCodeBench improves monotonically across the five budgets, whereas sampled IFEval peaks at $k=6$. These differences favor selecting the budget for the target workload; Table~\ref{tab:sft-decoding-full} provides all scores and intervals.

\subsection{Common-Protocol Public Post-Training Comparison}
\label{app:posttraining-common}

\textbf{Reference configurations.} The reference models are Qwen3-0.6B, Qwen3-1.7B, OLMoE-1B-7B-0125-SFT and Ouro-1.4B-Thinking.
Each model uses its native chat template. Qwen3 enables thinking; OLMoE uses its native direct-answer mode. Ouro uses four total recurrent steps with early exit at step three and preserves special tokens.
Generation uses vLLM 0.24 in bfloat16 with no system prompt. Sampling, reasoning-block extraction and deterministic IFEval scoring follow the common protocol in \S\ref{sec:sft-eval} for all references, SFT and RL.
Requested output limits are those in Table~\ref{tab:posttraining-response-limits}. The realized cap is the smaller of that limit and remaining context space. OLMoE's native context is 4,096, giving realized limits of 3,088--4,069; other references use 32,768.

\textbf{Primary scores and diagnostics.} All primary scores use the common task-specific answer extractors. Alternative code-block and MMLU letter-extraction scores characterize extraction sensitivity.
The model names and context limits identify the reference configurations used in Tables~\ref{tab:sft-public} and~\ref{tab:rl-public}.
Budget gains reflect both answer quality and whether the reasoning trace reaches a final answer.
On MATH500, sampling raises $k=0$ accuracy from 69.4\% to 83.4\% and reduces responses without a closing reasoning tag from 24.8\% to 5.4\%. Additional depth raises sampled accuracy to 89.2\%, yielding a 5.80-point endpoint gain alongside the 12.80-point gain under greedy decoding.
On sampled LiveCodeBench, this incomplete-response fraction falls from 36.8\% to 21.1\% as $k$ increases from 0 to 12.
These completion statistics accompany the matched decoding-policy comparisons in Table~\ref{tab:sft-decoding-full}.

\FloatBarrier
\section{Reinforcement-Learning Details}
\label{app:rl-details}
\subsection{Data Sources}
Expert training draws on public collections of instruction-following, mathematics and code problems. Source-level counts before difficulty and code-verifier filtering appear in Table~\ref{tab:rl-data-sources}; the final expert pools and balanced distillation sample appear in Table~\ref{tab:mopd-data}.

\begin{table*}[!htbp]
\centering
\accTableStretch{1.12}
\caption{\textbf{Public sources for RL.} Counts describe the deduplicated and decontaminated source pools before difficulty and code-verifier filtering. The final expert pool sizes are reported in the text. Mathematics counts sum to 207,156, IF to 152,608 and code to 22,016. Dataset names link to their public releases where separately identifiable.}
\label{tab:rl-data-sources}
\accTableFont{10.5}{13}
\setlength{\tabcolsep}{4pt}
\begin{tabularx}{\textwidth}{@{}cYc@{}}
\toprule
Domain & Public dataset / subset & Prompts \\
\midrule
Instruction following & \href{https://huggingface.co/datasets/nvidia/Nemotron-Cascade-RL-Instruction-Following}{Nemotron-Cascade-RL-Instruction-Following} & 98,092 \\
 & \href{https://huggingface.co/datasets/nvidia/Nemotron-RL-instruction_following}{Nemotron-RL-instruction\_following} & 45,079 \\
 & \href{https://huggingface.co/datasets/nvidia/Nemotron-RL-instruction_following-structured_outputs}{Nemotron-RL instruction following (structured outputs)} & 9,437 \\
\midrule
Mathematics & \href{https://huggingface.co/datasets/nvidia/Nemotron-RL-math-OpenMathReasoning}{Nemotron-RL-math-OpenMathReasoning} & 81,277 \\
 & \href{https://huggingface.co/datasets/open-r1/OpenR1-Math-220k}{OpenR1-Math-220k} & 59,703 \\
 & \href{https://huggingface.co/datasets/nvidia/AceReason-Math}{AceReason-Math} & 46,280 \\
 & Nemotron-Math-v2 (AoPS subset) & 17,479 \\
 & \href{https://huggingface.co/datasets/nvidia/Nemotron-Cascade-RL-Math}{Nemotron-Cascade-RL-Math} & 2,417 \\
\midrule
Code & \href{https://huggingface.co/datasets/nvidia/Nemotron-RL-coding-competitive_coding}{Nemotron-RL-coding-competitive\_coding} (competitive problems) & 14,170 \\
 & Same release (filtered OpenCodeReasoning-25k subset) & 7,846 \\
\bottomrule
\end{tabularx}
\end{table*}

\subsection{Expert Training and Model Selection}
Accordion-RL is obtained after ten distillation steps from a weighted merge of the three RL experts. All experts share Accordion-Instruct as their initialization. Their selected training steps and stopping criteria are summarized in Table~\ref{tab:rl-stage-status}.

\begin{table*}[!htbp]
\centering
\accTableStretch{1.12}
\caption{\textbf{Expert selection and entropy-based stopping.} Steps count domain-specific rollout/training iterations, each containing up to five per-budget optimizer updates; hardware allocations are reported below. The learning-rate scheduler advances once per training step. Validation scores average all five budgets. Training stops when entropy exceeds the domain-specific threshold. The selected experts are merged to initialize the student, selected after ten distillation training steps.}
\label{tab:rl-stage-status}
\accTableFont{10.5}{13}
\setlength{\tabcolsep}{3pt}
\begin{tabular*}{\textwidth}{@{\extracolsep{\fill}}lccccc@{}}
\toprule
Expert & Selected step & Validation score (\%) & Stopping step & Entropy at stopping & Entropy threshold \\
\midrule
IF & 130 & 81.11 & 133 & 1.811 & 1.8 \\
Mathematics & 30 & 47.89 & 45 & 1.533 & 1.5 \\
Code & 32 & 70.85 & 35 & 1.527 & 1.5 \\
\bottomrule
\end{tabular*}
\end{table*}

Expert training uses the filtered domain pools in \S\ref{sec:rl-data} and the prompt-group budget assignment and per-budget updates in \S\ref{sec:rl-expert-training}. Response generation and policy evaluation use the same budget.
The soft overlength penalty is disabled for mathematics and code. Prompts contain up to 2,048 tokens for mathematics and code, and 4,096 tokens for instruction following, within the 32,768-token context window.
The mathematics branch uses 24 nodes and code 40 nodes; the IF branch uses 24 nodes at stopping.
Training uses verl~\citep{sheng2024hybridflow}, FSDP2 and vLLM~\citep{kwon2023vllm}.

Validation samples two responses per problem at $T=0.6$ across all five budgets.
The IF and mathematics validation sets each contain 512 problems; the code set contains 387.
AIME's 60 problems are additionally monitored for mathematics but do not enter expert stopping decisions.
A new checkpoint must improve over the current best by more than one paired standard error to replace it; the plateau rule stops after three evaluations without such improvement.
In all three runs, the entropy threshold is reached before the validation plateau rule would stop training (Table~\ref{tab:rl-stage-status}).
The paired one-standard-error selection rule retains the step-30 mathematics expert: the subsequent 0.51-point increase at step 40 is below its 0.61-point paired standard error (Table~\ref{tab:rl-training-dynamics}).

\paragraph{Merge and student selection.}
The grouped merge is selected from the parameter-group alternatives in Table~\ref{tab:rl-merge-variants}. Student selection compares downstream evaluations of MOPD steps 10, 20, 30, 40 and 50 and selects step 10. Candidates after step 10 use an added task-reward term and a lower learning rate. Reported paired intervals quantify evaluation-sample uncertainty conditional on the selected models, without adjustment for checkpoint or merge selection.

\paragraph{Learning signals.}
During the ten-step distillation run, token distillation loss falls from 0.0708 to 0.0074, while monitored task reward rises from 0.466 to 0.649 under a distillation-only objective. Table~\ref{tab:rl-training-dynamics} places these signals alongside expert training gains and the mathematics selection statistics.

\begin{table*}[!htbp]
\centering
\caption{\textbf{Expert learning, distillation progress and selection statistics.} Expert validation averages the five budgets. The ten-step distillation objective uses teacher supervision; task reward is monitored separately. The lower block reports the paired mathematics checkpoint-selection comparison.}
\label{tab:rl-training-dynamics}
\accTableFont{10.5}{13}
\accTableStretch{1.12}
\begin{tabularx}{\textwidth}{@{}>{\raggedright\arraybackslash}m{.60\textwidth}Y@{}}
\toprule
Quantity & Value or change \\
\midrule
IF mean training reward, first 50 recorded steps & $\approx 0.47 \rightarrow 0.72$ \\
Mathematics validation gain over initial evaluation (pp) & $+1.11$ \\
Code validation gain over initial evaluation (pp) & $+6.59$ \\
MOPD task reward, ten-step run & $0.466 \rightarrow 0.649$ \\
MOPD token distillation loss, ten-step run & $0.0708 \rightarrow 0.0074$ \\
\midrule
Mathematics validation at step 40 (\%) & 48.40 \\
Step 40 improvement over selected step 30 (pp) & 0.51 \\
Paired standard error of that improvement (pp) & 0.61 \\
\bottomrule
\end{tabularx}
\end{table*}

\subsection{Full Matched Final-Model Evaluation}
\label{app:rl-final-paired}
The matched comparison evaluates Accordion-RL and Accordion-Instruct on the same problems at each budget. Table~\ref{tab:rl-final} reports all scores and the equal-task mean RL$-$SFT change within each displayed group: knowledge/instruction following (three tasks), mathematics (four) and code (three).
IFEval uses the deterministic scoring procedure in \S\ref{sec:sft-eval}. The matched generation and scoring protocol is specified there and in Appendix~\ref{app:sft-details}.

\begin{table*}[!htbp]
\centering
\accTableStretch{1.12}
\caption{\textbf{Complete matched SFT and RL evaluation.} Benchmark cells give Accordion-Instruct / Accordion-RL scores (\%). Mean $\Delta$ rows average the displayed RL$-$SFT differences equally over tasks within each group, in percentage points. Bold and underlining identify the best and second-best of the ten model--budget scores per benchmark; group changes are not ranked.}
\label{tab:rl-final}
\accTableFont{10.5}{13}
\setlength{\tabcolsep}{3pt}
\begin{tabularx}{\textwidth}{l*{5}{Y}}
\toprule
 & \multicolumn{5}{c}{\textbf{\accTableLogo Accordion-Instruct / Accordion-RL}} \\
\multirow{-2}{*}{Benchmark} & $k=0$ & $k=3$ & $k=6$ & $k=9$ & $k=12$ \\
\midrule
IFEval & 62.11 / 87.99 & 64.88 / \underline{88.91} & 65.99 / \textbf{89.83} & 63.40 / \textbf{89.83} & 60.07 / \textbf{89.83} \\
IFBench & 17.33 / 48.33 & 20.33 / \underline{50.67} & 20.33 / \underline{50.67} & 19.33 / \textbf{52.67} & 23.00 / 46.67 \\
MMLU-Redux & 60.88 / 70.98 & 63.47 / 73.68 & 65.82 / 74.13 & 66.17 / \textbf{75.18} & 65.16 / \underline{74.77} \\
\rowcolor{accTableOurs}
\textbf{Knowledge / IF mean $\Delta$} & $+22.33$ & $+21.53$ & $+20.83$ & $+22.93$ & $+21.01$ \\
\midrule
GSM8K & 83.40 / 84.15 & 85.67 / 87.34 & 88.78 / 88.63 & 87.72 / \textbf{89.54} & 88.63 / \underline{89.16} \\
MATH500 & 83.40 / 83.80 & 84.80 / 84.80 & 87.40 / 88.60 & 86.80 / \underline{89.60} & 89.20 / \textbf{89.80} \\
AIME 2024 & 30.00 / 34.58 & 34.79 / 39.79 & 39.58 / 40.83 & 43.54 / \textbf{45.21} & \underline{43.75} / \textbf{45.21} \\
AIME 2025 & 26.88 / 27.92 & 28.96 / 32.71 & 32.92 / \textbf{36.88} & \textbf{36.88} / 35.83 & 31.88 / \underline{36.67} \\
\rowcolor{accTableOurs}
\textbf{Mathematics mean $\Delta$} & $+1.69$ & $+2.61$ & $+1.57$ & $+1.31$ & $+1.85$ \\
\midrule
HumanEval+ & 79.88 / 79.88 & 78.05 / 78.05 & 79.27 / \textbf{86.59} & 79.88 / 82.93 & 81.71 / \underline{84.15} \\
MBPP+ & 69.58 / 69.58 & 67.72 / 71.43 & \underline{73.02} / 71.43 & 70.90 / 72.49 & 72.75 / \textbf{73.81} \\
LiveCodeBench & 26.65 / 28.63 & 31.94 / 29.74 & 34.58 / \textbf{39.87} & 36.34 / \underline{37.89} & \underline{37.89} / 35.24 \\
\rowcolor{accTableOurs}
\textbf{Code mean $\Delta$} & $+0.66$ & $+0.50$ & $+3.67$ & $+2.06$ & $+0.28$ \\
\bottomrule
\end{tabularx}
\end{table*}

\begin{figure*}[!htbp]
\centering
\includegraphics[width=5.4in]{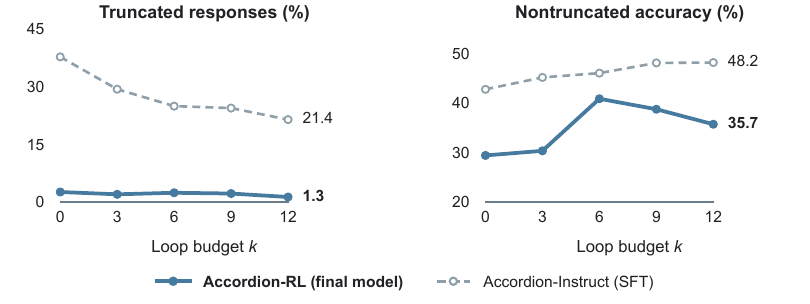}
\caption{\textbf{Response completion and conditional accuracy on LiveCodeBench.} RL and SFT use matched generation settings. Conditional scores characterize each model's own nontruncated subset, with RL$-$SFT differences of $-14.83$ to $-5.18$ points across these different subsets. Overall accuracy in Table~\ref{tab:rl-final} uses the full paired problem set.}
\label{fig:rl-completion}
\end{figure*}

RL reduces the response-level truncation rate on LiveCodeBench from 21.4--37.7\% under SFT to 1.3--2.6\% across budgets. On the full matched problem set, accuracy at $k=6$ rises from 34.58\% to 39.87\%; Table~\ref{tab:rl-final} reports all five budgets. Figure~\ref{fig:rl-completion} separates response completion from accuracy conditional on completion.

\begin{table*}[!htbp]
\centering
\caption{\textbf{Accordion-RL gains from additional depth.} Differences compare $k=12$ with $k=0$, in percentage points, and are calculated before rounding. Paired 95\% percentile intervals use 2,000 problem-level resamples with Python \texttt{random.Random(0)}; a dash indicates that no interval is reported here.}
\label{tab:rl-depth-gains}
\accTableFont{10.5}{13}
\accTableStretch{1.12}
\begin{tabular*}{\textwidth}{@{\extracolsep{\fill}}ccc@{\hspace{18pt}}ccc@{}}
\toprule
Task & $\Delta_{12}$ & 95\% CI & Task & $\Delta_{12}$ & 95\% CI \\
\midrule
GSM8K & $+5.00$ & $[3.18,6.82]$ & HumanEval+ & $+4.27$ & -- \\
MATH500 & $+6.00$ & $[3.00,9.00]$ & MBPP+ & $+4.23$ & $[0.53,7.94]$ \\
AIME 2024 & $+10.63$ & -- & LiveCodeBench & $+6.61$ & $[3.08,10.35]$ \\
AIME 2025 & $+8.75$ & -- & MMLU-Redux & $+3.79$ & $[2.57,4.99]$ \\
\bottomrule
\end{tabular*}
\end{table*}

\subsection{Model Merging}
\label{app:rl-merging}
All expert deltas are defined relative to the common SFT initialization and combined using the merge rule in \S\ref{sec:rl-construction}.
Table~\ref{tab:rl-merge-parameter-groups} specifies the coefficients and tensor counts for each parameter group.
Expert-comparison domain summaries average task differences equally within each budget and then average the five budgets; MMLU-Redux is excluded. Their 95\% intervals use paired problem-bootstrap resampling, averaging repeated AIME responses within each problem, without adjustment across comparisons.
Merging uses single-precision arithmetic on corresponding tensors with matching shapes, preserving the shared architecture and the frozen SFT balancing biases.

\begin{table}[!htbp]
\centering
\caption{\textbf{Parameter groups in the selected RL merge.} Coefficients multiply expert deltas relative to SFT. The first row defines the 289 shared-component tensors referred to in the merge comparisons. Frozen balancing biases are copied, not merged.}
\label{tab:rl-merge-parameter-groups}
\accTableFont{10.5}{13}
\accTableStretch{1.12}
\setlength{\tabcolsep}{5pt}
\begin{tabularx}{\textwidth}{@{}>{\raggedright\arraybackslash}m{.56\textwidth}*{4}{Y}@{}}
\toprule
\multicolumn{1}{c}{Parameter group} & Tensors & IF & Code & Math \\
\midrule
Attention projections and attention normalization; shared MoE experts; loop gates; layer normalization, including the final norm & 289 & 1 & 0.5 & 0.5 \\
\addlinespace[4pt]
Routed expert projections; routers; embeddings; language-model head & 74 & 1 & 1 & 1 \\
\addlinespace[4pt]
\multicolumn{1}{c}{Frozen balancing biases} & 48 & \multicolumn{3}{c}{Copied from SFT} \\
\midrule
\multicolumn{1}{c}{Total} & 411 & -- & -- & -- \\
\bottomrule
\end{tabularx}
\end{table}

Scaling code and mathematics deltas in shared components improves the mathematics-domain difference by 3.99 points relative to unit-weight merging, while retaining positive gains over the instruction-following and code experts (Table~\ref{tab:rl-merge-variants}). These merge controls test the parameter grouping used to initialize three-domain distillation.
After distillation, the student exceeds the corresponding specialist by 1.65 points in instruction following, 1.37 in mathematics and 2.04 in code, with all three paired 95\% intervals above zero (Table~\ref{tab:rl-integration-summary}). These equal-task, equal-budget domain averages establish integration of the three specialists in one checkpoint.

\begin{table*}[!htbp]
\centering
\accTableStretch{1.12}
\caption{\textbf{Merge variants relative to the corresponding domain expert.} Entries are equal-task, equal-budget mean gains [paired 95\% intervals], in percentage points. Omitted domains in pairwise merges have coefficient zero; unspecified included coefficients are one. ``Shared'' denotes the 289 tensors in Table~\ref{tab:rl-merge-parameter-groups}. A mathematics-domain interval is unavailable for the IF--code merge.}
\label{tab:rl-merge-variants}
\accTableFont{10.5}{13}
\setlength{\tabcolsep}{2pt}
\begin{tabular*}{\textwidth}{@{\extracolsep{\fill}}cccc@{}}
\toprule
Variant & Instruction following & Code & Mathematics \\
\midrule
IF + code & +1.76 [+0.35,+3.11] & +0.47 [\textminus0.77,+1.62] & -- \\
IF + math & +1.93 [+0.51,+3.28] & +2.03 [+0.83,+3.21] & \textminus0.52 [\textminus1.34,+0.32] \\
Code + math & \textminus24.67 [\textminus26.44,\textminus22.91] & +1.97 [+0.85,+3.13] & \textminus0.06 [\textminus0.84,+0.69] \\
Unit weights & +1.85 [+0.47,+3.21] & +0.44 [\textminus0.76,+1.66] & \textminus4.92 [\textminus5.95,\textminus3.93] \\
Math 0.5 throughout & +2.69 [+1.31,+4.11] & +1.17 [+0.12,+2.27] & \textminus2.24 [\textminus3.03,\textminus1.44] \\
Shared code/math 0.5 & +2.36 [+0.90,+3.79] & +1.78 [+0.64,+2.98] & \textminus0.93 [\textminus1.75,\textminus0.10] \\
Remaining code/math 0.5 & +1.97 [+0.65,+3.38] & +1.11 [\textminus0.06,+2.29] & \textminus3.18 [\textminus4.14,\textminus2.23] \\
Shared code 0.5 & +2.04 [+0.63,+3.44] & +1.31 [+0.22,+2.40] & \textminus2.00 [\textminus2.84,\textminus1.17] \\
Shared IF 0.5 & \textminus3.41 [\textminus4.78,\textminus1.99] & +1.52 [+0.41,+2.67] & \textminus1.73 [\textminus2.57,\textminus0.92] \\
\bottomrule
\end{tabular*}
\end{table*}

\begin{table*}[!htbp]
\centering
\caption{\textbf{Domain-level gains after merging and distillation.} Differences are percentage points, averaging tasks equally within each budget and then averaging the five budgets. Intervals use paired problem-bootstrap resampling.}
\label{tab:rl-integration-summary}
\accTableFont{10.5}{13}
\accTableStretch{1.12}
\begin{tabularx}{\textwidth}{@{}>{\raggedright\arraybackslash}m{.50\textwidth}YY@{}}
\toprule
Comparison & Mean gain & 95\% CI \\
\midrule
Student versus IF expert, instruction following & $+1.65$ & $[0.30,3.03]$ \\
Student versus mathematics expert, mathematics & $+1.37$ & $[0.65,2.11]$ \\
Student versus code expert, code & $+2.04$ & $[0.82,3.21]$ \\
Student versus grouped merge, mathematics & $+2.30$ & $[1.46,3.14]$ \\
\bottomrule
\end{tabularx}
\end{table*}

\subsection{Multi-Teacher On-Policy Distillation}
\label{app:rl-mopd}

\textbf{Teacher matching.} MOPD initializes the student from the grouped merge and selects one of the three expert teachers according to each prompt's domain.
The teacher uses the student's loop budget and scores the same response generated by the student.
For a student response $y$ to prompt $x$ from domain $j$, let $c_t=(x,y_{<t})$ denote the token context. The token-level training signal is
\[
a_t^{\mathrm{distill}}=-\operatorname{clip}\!\left(
\log\frac{\pi_{\mathrm{student}}(y_t\mid c_t;k)}{\pi_{\mathrm{teacher},j}(y_t\mid c_t;k)},-10,10\right).
\]
Both probabilities therefore score the same token and prefix at the same budget. This signal is used as a token-level advantage with proximal policy optimization (PPO) clipping at 0.2 and token-level importance weights correcting rollout/training discrepancies.
The first ten training steps optimize distillation alone, with coefficient one; task rewards are logged but do not enter this objective.

\textbf{Sampling and updates.} Distillation samples equally from the three domain pools in \S\ref{sec:rl-data}. It uses the same identifier-based budget assignment and separate per-budget optimizer updates as expert training (\S\ref{sec:rl-expert-training}). The learning-rate scheduler advances once per training step, which can contain up to five optimizer updates. Teacher inference instances serve fixed domain--budget combinations, ensuring that supervision matches the student's domain and budget. The generation, optimization and hardware settings are listed in Table~\ref{tab:rl-mopd-settings}.

\begin{table}[!htbp]
\centering
\caption{\textbf{Multi-teacher on-policy distillation settings.} The student generates its own responses, and each domain expert evaluates them at the student's inference budget.}
\label{tab:rl-mopd-settings}
\accTableFont{10.5}{13}
\accTableStretch{1.12}
\setlength{\tabcolsep}{6pt}
\begin{tabular*}{\textwidth}{@{\extracolsep{\fill}}cccc@{}}
\toprule
Setting & Value & Setting & Value \\
\midrule
Training pool (prompts) & 57,600 & Warmup (training steps) & 10 \\
Prompts per rollout batch & 576 & Peak learning rate & $10^{-6}$ \\
Responses per prompt & 8 & Student nodes & 36 \\
Responses per rollout batch & 4,608 & Additional teacher nodes & 4 \\
Max. prompt tokens & 2,048 & GPUs per node & 8 H800 \\
Max. response tokens & 28,672 & Teacher inference instances & 32 \\
Context limit (tokens) & 32,768 &  &  \\
\bottomrule
\end{tabular*}
\end{table}

Domain-matched teachers and budget-matched supervision preserve the five inference paths during expert integration. The resulting domain-average improvements are reported in Table~\ref{tab:rl-integration-summary}.

\FloatBarrier

\Needspace*{6\baselineskip}
\begingroup
\accLocalParskip{9.5pt plus 1pt minus 1pt}
\setlength{\intextsep}{6pt plus 2pt minus 2pt}
\setlength{\textfloatsep}{10pt plus 2pt minus 2pt}
\section{Computational Details}
\label{app:main-results}

\subsection{Computational Cost}
\label{app:cost}
Uniform budget sampling jointly trains five inference paths in one checkpoint, with an average calculated training cost of approximately $1.202\times$ the unlooped path per token at 8K (a 20.2\% increase). Estimates include forward and backward attention and active-parameter computation, excluding communication and activation recomputation. KV storage assumes bf16 values, four 128-dimensional heads and separate caches for repeated layers (Table~\ref{tab:cost}).
Latency grows approximately with execution count, while additional memory closely follows KV storage. Table~\ref{tab:speed} includes the 95th-percentile end-to-end latency across timed iterations.

\phantomsection
\label{app:runtime}
We measure inference cost across all budgets using one 80\,GB H800 with vLLM 0.24.0~\citep{kwon2023vllm} and PyTorch 2.11. Weights and activations use bf16, with FlashAttention, fused MoE kernels, graph compilation and CUDA graphs.
Each layer execution has a separate KV cache. Chunked prefill processes up to 32,768 tokens per step; prefix caching is disabled, and the KV pool is sized for each workload. Prompts are contiguous held-out web spans. Measurements use Accordion-RL, whose architecture is unchanged by post-training. Table~\ref{tab:speed} specifies prompt lengths and batch sizes.

Greedy decoding ignores the end-of-sequence marker.
Let $T_n$ denote the synchronized latency to generate $n$ output tokens in one iteration. We measure $T_1$ and $T_{128}$ and compute
\begin{equation}
  \mathrm{TTFT}=T_1,\qquad
  \mathrm{TPOT}=\frac{T_{128}-T_1}{127},\qquad
  T_{\mathrm{end\mbox{-}to\mbox{-}end}}=T_{128}.
  \label{eq:runtime-latency}
\end{equation}
Five warm-up iterations precede 20 timed iterations.
Reported latencies average two hosts' medians, with budget order reversed between hosts. The largest discrepancy is 7.7\% relative to the smaller measurement. Peak allocated GPU memory includes weights (approximately 11\,GiB), the KV pool and CUDA-graph buffers. At the 32K workload with batch size eight, peak allocated memory ranges from 24.90 to 30.90\,GiB across the five budgets on one 80\,GB H800.

\begin{table}[H]
  \centering
  \caption{\textbf{Calculated model costs and measured inference cost.} The first block is analytical. In the runtime block, each metric cell gives \textbf{batch 1 / batch 8}, measured on one H800 with 128 output tokens per sequence. Throughput counts tokens across the batch; latency covers the batch. Prompt lengths group all five budgets, with 32,640 prompt tokens per sequence for the 32K workload. Measurement settings appear above.}
  \label{tab:cost}
  \label{tab:speed}
  \label{tab:speed-batch8}
  \accTableFont{10.5}{13}
  \accTableStretch{1.0}
  \setlength{\tabcolsep}{3pt}
  \begin{tabular*}{0.80\textwidth}{@{\extracolsep{\fill}}cccccc@{}}
    \toprule
    Calculated model costs at 8K & $k=0$ & $k=3$ & $k=6$ & $k=9$ & $k=12$ \\
    \midrule
    Layer executions & 24 & 27 & 30 & 33 & 36 \\
    Rel.\ train FLOPs & 1.000 & 1.101 & 1.202 & 1.303 & 1.404 \\
    KV/token (KiB) & 48 & 54 & 60 & 66 & 72 \\
    \bottomrule
  \end{tabular*}
  \par\vspace{5pt}
\begin{tabular*}{\textwidth}{@{\extracolsep{\fill}}*{9}{c}@{}}
\toprule
Prompt tokens & $k$ & Prefill (tok/s) & TTFT (ms) & Decode (tok/s) & TPOT (ms) & p95 (ms) & Peak (GiB) & KV (GiB) \\
\midrule
8,192 & 0 & 198.3k / 216.9k & 41.3 / 302.2 & 379 / 1,310 & 2.64 / 6.11 & 391 / 1,080 & 12.17 / 15.94 & 0.38 / 3.00 \\
 & 3 & 176.7k / 192.0k & 46.4 / 341.3 & 342 / 1,191 & 2.93 / 6.72 & 420 / 1,223 & 12.22 / 16.33 & 0.42 / 3.38 \\
 & 6 & 159.6k / 172.7k & 51.3 / 379.4 & 309 / 1,073 & 3.24 / 7.45 & 478 / 1,343 & 12.27 / 16.71 & 0.47 / 3.75 \\
 & 9 & 144.2k / 157.2k & 56.8 / 416.9 & 275 / 944 & 3.64 / 8.48 & 523 / 1,499 & 12.31 / 17.09 & 0.52 / 4.13 \\
 & 12 & 133.3k / 144.0k & 61.4 / 455.1 & 252 / 887 & 3.96 / 9.02 & 567 / 1,640 & 12.36 / 17.47 & 0.56 / 4.50 \\
\midrule
16,384 & 0 & 179.8k / 178.7k & 91.1 / 733.6 & 343 / 1,083 & 2.92 / 7.38 & 463 / 1,685 & 12.91 / 18.94 & 0.75 / 6.00 \\
 & 3 & 158.1k / 158.2k & 103.6 / 828.6 & 330 / 990 & 3.03 / 8.08 & 502 / 1,869 & 13.01 / 19.70 & 0.84 / 6.75 \\
 & 6 & 142.4k / 140.4k & 115.0 / 933.4 & 281 / 895 & 3.56 / 8.94 & 569 / 2,088 & 13.10 / 20.46 & 0.94 / 7.50 \\
 & 9 & 129.7k / 126.7k & 126.3 / 1,034.2 & 257 / 818 & 3.89 / 9.78 & 622 / 2,296 & 13.20 / 21.21 & 1.03 / 8.25 \\
 & 12 & 118.2k / 116.3k & 138.6 / 1,127.0 & 246 / 764 & 4.07 / 10.47 & 674 / 2,480 & 13.29 / 21.97 & 1.13 / 9.00 \\
\midrule
32,640 & 0 & 134.5k / 130.5k & 242.7 / 2,001.0 & 314 / 876 & 3.19 / 9.14 & 648 / 3,167 & 14.39 / 24.90 & 1.49 / 11.95 \\
 & 3 & 118.8k / 115.8k & 274.7 / 2,254.4 & 293 / 781 & 3.42 / 10.24 & 724 / 3,560 & 14.57 / 26.40 & 1.68 / 13.45 \\
 & 6 & 107.3k / 104.4k & 304.2 / 2,501.0 & 256 / 709 & 3.91 / 11.29 & 802 / 3,952 & 14.76 / 27.90 & 1.87 / 14.94 \\
 & 9 & 97.4k / 95.0k & 335.1 / 2,749.1 & 235 / 645 & 4.25 / 12.41 & 877 / 4,336 & 14.95 / 29.40 & 2.05 / 16.44 \\
 & 12 & 89.3k / 86.9k & 365.7 / 3,004.6 & 217 / 590 & 4.61 / 13.57 & 954 / 4,733 & 15.14 / 30.90 & 2.24 / 17.93 \\
\bottomrule
\end{tabular*}
\end{table}

\par\endgroup

\end{document}